\documentclass{article}

\usepackage{microtype}
\usepackage{graphicx}
\usepackage{subcaption}
\usepackage{booktabs} 

\usepackage{hyperref}

\usepackage[accepted]{conference}

\usepackage{amsmath}
\usepackage{amssymb}
\usepackage{mathtools}
\usepackage{amsthm}
\usepackage{siunitx}
\usepackage[T1]{fontenc}
\usepackage{adjustbox}
\usepackage[normalem]{ulem} 
\usepackage{tablefootnote}
\usepackage{multirow}

\usepackage[capitalize,noabbrev]{cleveref}

\usepackage[dvipsnames,table]{xcolor}

\theoremstyle{plain}

\theoremstyle{definition}

\theoremstyle{remark}

\usepackage{framed}

\definecolor{color_finetune}{HTML}{cacaca}
\definecolor{color_optimization}{HTML}{f5b4c7}
\definecolor{color_regularization}{HTML}{e7c3b0}
\definecolor{color_replay}{HTML}{b5d1d6}
\definecolor{color_tailored}{HTML}{bdd3b2}
\definecolor{color_tailored_border}{HTML}{314527}

\definecolor{color_ctta}{HTML}{cc6600}

\definecolor{color_naive}{HTML}{d8c3f2}
\definecolor{color_naive_border}{HTML}{8645d6}

\usepackage[disable,textsize=tiny]{todonotes}

\usepackage{tikz}
\DeclareRobustCommand\circled[1]{\tikz[baseline=(char.base)]{\node[shape=circle,draw=blue,minimum size=0.4cm,inner sep=0pt,fill=lightblue] (char) {\fontfamily{phv}\selectfont \scriptsize \textbf{#1}};}}
\definecolor{lightblue}{HTML}{DAE8FC}

\DeclareRobustCommand\naivecircled[1]{\tikz[baseline=(char.base)]{\node[shape=circle,draw=color_naive_border,minimum size=0.4cm,inner sep=0pt,fill=color_naive] (char) {\fontfamily{phv}\selectfont \scriptsize \textbf{#1}};}}

\DeclareRobustCommand\tailoredcircled[1]{\tikz[baseline=(char.base)]{\node[shape=circle,draw=color_tailored_border,minimum size=0.4cm,inner sep=0pt,fill=color_tailored] (char) {\fontfamily{phv}\selectfont \scriptsize \textbf{#1}};}}

\usepackage{array}
\usepackage{ragged2e}
\newcolumntype{P}[1]{>{\RaggedRight\hspace{0pt}}p{#1}}

\newcommand{\sourceD}{{D_s}}
\newcommand{\sourceDPrime}{{D_{s^{\prime}}}}
\newcommand{\sourceBPrime}{{B_{s^{\prime}}}}

\newcommand{\sourceX}{{\mathbf{x}^{s}_i}}
\newcommand{\sourceY}{{y^{s}_i}}
\newcommand{\priorPrime}{{\Phi_{s^{\prime}}}}

\newcommand{\targetD}{{D_t}}

\newcommand{\targetX}{{\mathbf{x}^{t}_i}}
\newcommand{\targetY}{{y^{t}_i}}

\renewcommand{\algorithmicrequire}{\textbf{Input:}}

\icmltitlerunning{Continual Learning of Domain-Invariant Representations}

\begin{document}

\twocolumn[
  \icmltitle{Continual Learning of Domain-Invariant Representations
  }



  \icmlsetsymbol{equal}{*}

  \begin{icmlauthorlist}
    \icmlauthor{Pascal Janetzky}{LMU,MCML,BCAI}
    \icmlauthor{Tobias Schlagenhauf}{HSALBSIG}
    \icmlauthor{Stefan Feuerriegel}{LMU,MCML}
  \end{icmlauthorlist}

  \icmlaffiliation{LMU}{LMU Munich, Munich, Germany}
  \icmlaffiliation{BCAI}{Bosch Center for Artificial Intelligence, Renningen, Germany}
  \icmlaffiliation{MCML}{Munich Center for Machine Learning, Munich, Germany}
  \icmlaffiliation{HSALBSIG}{Albstadt-Sigmaringen University}

  \icmlcorrespondingauthor{Pascal Janetzky}{pascal.janetzky@bosch.com}

  \icmlkeywords{Machine Learning, Continual Learning, Continual Source Domain Generalization, Domain Invariant Representation}

  \vskip 0.3in
]



\printAffiliationsAndNotice{}  

%

\begin{abstract}
Continual learning (CL) aims to train models sequentially over multiple domains without forgetting previously learned knowledge. However, existing CL methods optimize for in-domain performance and are therefore prone to learning spurious, domain-specific cues (``shortcut learning''), which limits generalization to unseen domains after deployment. In this paper, we address this limitation through \emph{continual learning of domain-invariant representation}. We introduce a broad class of CL methods that sequentially learn representations capturing invariant structures across domains. Our methods are motivated by the observation that such invariant structures often preserve the underlying causal mechanisms, which can reduce the risk of overfitting to domain-specific cues and thus offer better out-of-domain generalization. Our proposed CL methods combine replay-based training with a tailored sequential invariance alignment to learn---and preserve---invariant structures over time. We evaluate our methods under a deployment-oriented protocol that measures performance on unseen target domains. Across six benchmark and real-world datasets spanning vision, medicine, manufacturing, and ecology, our methods consistently outperform existing CL baselines in terms of generalization to unseen target domains. As an ablation, we further show that na\"ive extensions of sequential training with existing domain-invariant representation learning (DIRL) methods provide only limited benefits. To the best of our knowledge, this is the first work to develop domain-invariant representation methods for CL.
\end{abstract}

\section{Introduction}
Continual learning (CL) trains machine learning models sequentially on multiple domains without forgetting previously learned knowledge \citep[see Section~\ref{sec:related_work} for a detailed method overview]{parisi2019continual,wang2024comprehensive}. This setting arises in many real-world applications, including autonomous driving \citep{shaheen2022continual}, robotics \citep{thrun1995lifelong}, medicine \citep{vokinger2021continual,bruno2025continual}, and manufacturing \citep{hurtado2023continual}, where models must be continually updated as new data becomes available. A key challenge is to prevent forgetting, often referred to as ``\emph{catastrophic forgetting}'' \citep{ratcliff1990connectionist}, as new information is learned, which is typically measured by the in-domain performance on held-out splits of the training domains \citep{chaudry2019efficient,caccia2021new}.

However, the primary objective in existing CL methods is to maintain performance on previously seen training domains \citep{kirkpatrick2017overcoming,saha2021gradient,wang2024comprehensive}. This can encourage models to rely on spurious, domain-specific cues that are predictive within individual domains, but that do \emph{not} reflect stable relationships shared across domains. Such behavior is closely related to \emph{``shortcut learning''}, where models exploit superficial regularities that are sufficient for in-domain generalization but fail on new domains \citep{geirhos2020shortcut,hermann2023foundations}. As a result, representations learned by standard CL methods may generalize poorly to new domains encountered after deployment. 

This limitation is especially important in practice because many applications of CL are governed by biological \citep[e.g.][]{lee2020clinical,kiyasseh2021clinical,bruno2025continual} or physical \citep[e.g.][]{hurtado2023continual,wang2023contrastive,tang2024incremental} processes, in which stable causal mechanisms determine the relationship between inputs and labels. For example, in medicine, physiological responses to radiotherapy follow the same underlying biological principles across patients, as radiation damages cancer cells through well-understood mechanisms. Similarly, in manufacturing, failure patterns are often driven by persistent physical processes, such as excessive heat, mechanical stress, or material fatigue, which remain consistent across machines, fabrication sizes, or production batches. Hence, these settings motivate learning \emph{domain-invariant representations}, which are representations that capture structures shared across domains while being insensitive to domain-specific variations, such as the background in images. For physical systems, these representations are known to improve performance \citep{da2020remaining,tanwani2021dirl,hua2025domain}, as they are more likely to preserve stable causal mechanisms and, as a result, to support robust generalization to new domains. \emph{However, to the best of our knowledge, methods for CL with domain-invariant representations are lacking.}  

\begin{figure*}[!htbp]
    \centering
    \includegraphics[width=\linewidth]{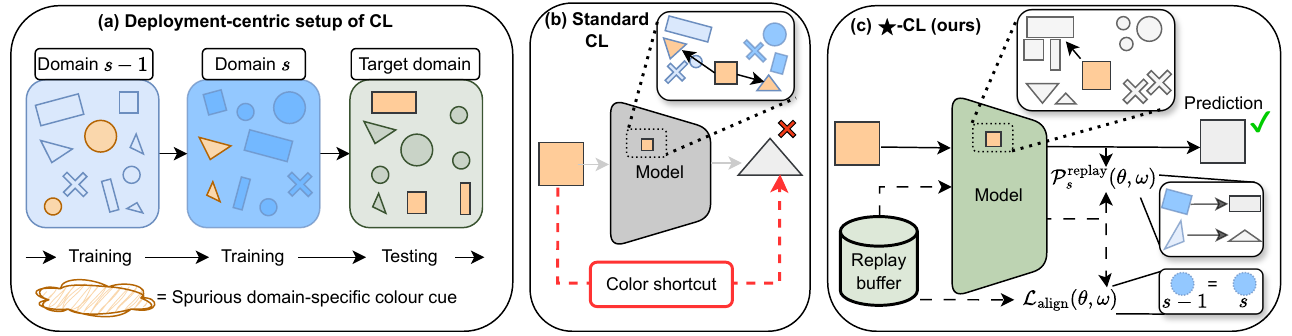}
    \caption{
    \textbf{Deployment-centric setup of CL.} \emph{(a):} Training domains (shown in different colors) arrive sequentially; upon deployment, the model is evaluated on an arbitrary target domain. \emph{(b):} Standard CL methods are prone to learning spurious, domain-specific cues (e.g., the color) and thus fail to classify data from unseen domains at deployment time. \emph{(c):} Our methods learn domain-invariant representation (e.g., the shape) and thus generalize to both seen \emph{and} unseen domains. Hence, our evaluation explicitly tests the out-of-domain generalization to ensure that robust patterns are learned, as in works on domain-invariant learning \citep{krueger2021out,rame2022fishr}.}
    \label{fig:fig1}
\end{figure*}

In this paper, we address the above limitation through \textbf{\emph{continual learning of domain-invariant representations}}. We introduce a broad class of CL methods that sequentially learn representations while capturing invariant structures across domains, with the goal of improving generalization to unseen domains after deployment. We empirically show that learning such invariant representations is crucial for robust post-deployment performance. 

\circled{1} \textbf{\emph{Why continual learning of domain-invariant representation is non-trivial (\cref{sec:naive_methods}})}. Constructing CL methods that learn domain-invariant representations is challenging. Existing domain-invariant representation learning methods \citep[e.g.][]{parascandolo2020learning,krueger2021out,rame2022fishr} are designed for \emph{joint} access to multiple domains and then \emph{simultaneously} optimize invariance constraints across domains; yet, this does \emph{not} directly extend to sequential training. Here, domains arrive sequentially and past data are no longer fully accessible. As a result, na{\"i}ve adaptations, such as replaying static invariance statistics or matching current representations to previously computed summaries, fail to faithfully reproduce multi-domain invariance objectives and offer little improvement over standard continual fine-tuning. We nevertheless include such na\"ive extensions in our experiments. While these adaptations have not been studied before in continual settings and are therefore novel in their own right, we later show that they provide only limited benefits over standard continual fine-tuning.

\circled{2} \textbf{\emph{New CL methods for domain-invariant representations (\cref{sec:our_methods})}}. To overcome the above challenges, we develop a broad class of CL methods that explicitly learn and preserve invariant structure over time. The proposed methods leverage replay-based training together with multi-domain invariance computation and sequential invariance alignment, which thus allows invariant structures to be maintained as new domains arrive. We consider multiple notions of domain invariance, namely, (i)~risk-based, (ii)~gradient-based, and (iii)~feature-based—domain invariances. As a result, we propose five new CL algorithms: $\bigstar$-\textbf{CL-VREX}, $\bigstar$-\textbf{CL-Fishr}, $\bigstar$-\textbf{CL-CORAL}, $\bigstar$-\textbf{CL-MMD}, and $\bigstar$-\textbf{CL-ANDMask}.

\circled{3} \textbf{\emph{Deployment-oriented evaluation protocol (\Cref{sec:problem_statement})}}. We adopt a deployment-oriented evaluation protocol to demonstrate generalization. Specifically, our evaluation mirrors the workflow in practical applications, given by \emph{sequential training} $\rightarrow$ \emph{deployment} $\rightarrow$ \emph{testing on a new target domain} (\Cref{fig:fig1}). This setting reflects real-world scenarios, such as in medicine or manufacturing, where models are trained on a sequence of source domains but where the characteristics of the eventual target domain are unknown at training time. Hence, we carefully measure the \emph{out-of-domain performance} in addition to the in-domain performance that is typically reported in standard CL evaluations. Under this protocol, we conduct extensive experiments on six established benchmark and real-world datasets. Across all settings, our proposed methods consistently outperform existing CL baselines. This highlights that our proposed methods do not overfit to spurious, in-sample cues but actually learn domain-invariant mechanisms that generalize robustly to new domains. 

\textbf{Contributions:} \textbf{(1)}~We propose a set of tailored CL methods that learn domain-invariant representations over a sequence of source domains with the aim of generalizing to a new target domain. \textbf{(2)}~We propose a deployment-oriented evaluation protocol to assess how methods generalize to new, previously unseen domains. \textbf{(3)}~We perform extensive experiments across six datasets, showing that the proposed methods achieve SOTA performance. Therein, we further show that na\"ive extensions of sequential fine-tuning for domain-invariant learning have limited benefits, which motivates our proposed methods.

\section{Related work\protect\footnote{We provide an extended related work in \cref{sec:extended_related_work}.}}\label{sec:related_work}

\textbf{Continual learning (CL):} CL studies how models can be sequentially trained on a stream of data while preventing forgetting \citep{mccloskey1989catastrophic}. Most CL methods are designed around the stability-plasticity tradeoff \citep{lu2025rethinking} and are typically evaluated by how well they perform on held-out splits of previously seen training domains \citep{kirkpatrick2017overcoming,saha2021gradient,wang2024comprehensive}. To achieve this, existing method broadly fall into four categories: \textbf{(1)}~optimization-based methods, which modify gradient updates \citep{chaudry2019efficient,elsayed2024addressing}; \textbf{(2)}~regularization-based methods, which reduce forgetting by regularizing updates to model weights or features \citep{kirkpatrick2017overcoming,zenke2017continual,magistri2024elastic}; \textbf{(3)}~architecture-based approaches, which add a dedicated new network module for each new domain (and, hence, often scale quadratically or even exponentially with the number of domains) \citep{rusu2016progressive,yoon2017lifelong,yoon2019scalable}; and \textbf{(4)}~replay-based approaches, which maintain a buffer to replay past samples \citep{chaudhry2019tiny,churamani2023towards,eskandar2025star}, features \citep{iscen2020memory,toldo2022bring}, or latent information \citep{de2021continual,sarfraz2023sparse,sarfraz025semantic}. \emph{Later, we benchmark our proposed methods against state-of-the-art CL methods, which serve as the main baselines in our CL setting.}

However, the main goal of existing CL methods is to optimize performance on previously seen domains. This objective encourages learning patterns that are sufficient for high in-domain performance, including spurious, domain-specific cues that need not hold beyond the training distributions. As a result, models can achieve strong performance on all observed domains while still failing to generalize to new, unseen target domains after deployment. \emph{In contrast, our work explicitly targets generalization beyond the training domains by learning domain-invariant structures over time.}

\textbf{Domain invariant representation learning (DIRL):} DIRL aims to learn features that transfer to new domains by enforcing invariances shared across multiple source domains \citep{li2018domain,krishnamachari2024uniformly}. As a result, invariant representations seek to encode the underlying causal mechanisms, which are stable across domains and which reduces overfitting to domain-specific cues and thus improves robustness to distribution shifts \citep{arjovsky2019invariant,krueger2021out,mahajan2021domain,rame2022fishr}. Common examples are CORAL \citep{sun2016deep} and MMD \citep{li2018domain} which align features across domains, VREX \citep{krueger2021out}, which aligns risks, and Fishr \citep{rame2022fishr} and ANDMask \citep{parascandolo2020learning}, which align gradients.

However, a key limitation of existing DIRL methods is that these assume \emph{joint} access to multiple source domains during training to \emph{simultaneously} optimize invariance constraints across domains \citep{sun2016deep,li2018domain,rojas2018invariant,arjovsky2019invariant,parascandolo2020learning,shi2021gradient,rame2022fishr,krishnamachari2024uniformly}. This is \emph{fundamentally different} from our CL scenario, where domains arrive \emph{sequentially} and where past domains are no longer fully accessible. \emph{Hence, standard DIRL methods are \underline{not} directly applicable to CL without substantial modification.}

\textbf{Research gap:} To the best of our knowledge, there is currently \textbf{\underline{no}} CL method that explicitly learns domain-invariant structures to achieve out-of-domain generalization. At the same time, DIRL methods are \textbf{\underline{not}} directly applicable to CL settings, and, as we show later, na{\"i}ve extensions provide only limited benefits, which motivate our tailored methods.

\section{Problem setup}\label{sec:problem_statement}
We follow standard CL literature \citep{serra2018overcoming,kang2022forget,sarfraz025semantic} and consider a supervised deployment-oriented setting where a model $h$ is trained on a sequence $\mathcal{S}$ of $k$ labeled \emph{source} domains $\sourceD = \{\sourceX, \sourceY\}_i^{n_s}$ and afterwards evaluated on an unseen \emph{target} domain $\targetD =\{\targetX, \targetY\}_i^{n_t}$. We use the term `domain $s$' or `time $s$' interchangeably when referring to $\sourceD$. Each labeled domain $D_\diamond$, $\diamond \in \{ s, t \}$ has $n_\diamond$ labeled samples $\{(\mathbf{x}_i^\diamond, y_i^\diamond) \}_{i=1}^{n_\diamond}$. All domains share the same label space, i.e., $y_i \in \{1, \ldots, C\}$, where $C$ is the total number of classes. This setting is often referred to as domain-incremental learning \citep{van2022three}; but we evaluate on an unseen target domain.

We consider a deep neural network $h$, which consists of: (i)~a \emph{feature extractor} $f_\theta:\mathcal{X} \xrightarrow{} \mathcal{H}$, which projects input data to latent space $\mathcal{H}$; and (ii)~a \emph{classifier} $g_\omega:\mathcal{H}\xrightarrow{} \mathcal{Y}$, which classifies the latent representations based on the set of predefined classes $C$. The full model is given by $h = g_\omega \circ f_\theta = g_\omega(f_\theta(\cdot))$. 

Following standard practice in CL, training $h$ proceeds sequentially over the domains. For each source domain $\sourceD$, the model parameters $(\theta,\omega)$ are updated by minimizing the empirical risk
\begin{equation}
    \min_{\theta,\omega}\quad\frac{1}{n_s} \sum_{i=1}^{n_s} \mathcal{L} (h(\sourceX;\theta, \omega), \sourceY),\label{eq:objective_cl}
\end{equation}
where $\mathcal{L}: \mathcal{Y} \times \mathcal{Y} \xrightarrow{}[0,\infty)$ is a classification loss, such as the cross-entropy loss. This objective corresponds to standard empirical risk minimization (ERM) applied to the current domain $s$. In the continual learning setting, data from previous domains is typically no longer accessible once training moves on to a new domain. We therefore assume that only data from the current domain $\sourceD$ is available at training time. Here, one typically allows for storing a small subset in a domain-partitioned \emph{memory buffer} $M_s$, with $|M_s| \ll |\sourceD|$.

\section{Na{\"i}ve domain-invariant CL algorithms}\label{sec:naive_methods}
To learn domain-invariant representations along source domain sequence $S$, we first construct a set of na\"ive continual learning extensions of standard domain-invariant learning methods \citep[e.g.][]{li2018domain,krueger2021out}. For this, we decompose the training objective for domain $s$ into an ERM term and an invariance penalty: 
\begin{equation}
    \min_{\theta,\omega}\quad 
    \mathcal{L}_{\mathrm{ERM}}(\sourceD;\theta,\omega) \;+\; \lambda\,\mathcal{P}_s(\theta,\omega),
    \label{eq:na\"ive_template}
\end{equation}
where $\mathcal{L}_{\mathrm{ERM}}(\sourceD;\theta,\omega)=\frac{1}{n_s}\sum_{i=1}^{n_s}\mathcal{L}(h(\sourceX;\theta,\omega),\sourceY)$, where $\mathcal{P}_s$ is the invariance penalty for domain $s$, and where $\lambda>0$ is a scalar weighting factor.

\begin{table*}[!htbp]
\centering
\tiny
\caption{\textbf{Our proposed set of CL methods} for sequentially learning domain-invariant representations. Each method defines a per-domain ${s^{\prime}}$ statistic $\widehat{\phi}_{s^{\prime}}(\theta,\omega;\sourceBPrime)$ computed on minibatch $\sourceBPrime$ (current domain or replay), and a stored domain prior $\priorPrime$ computed at the end of domain ${s^{\prime}}$. The invariance penalty $\mathcal{P}^{\mathrm{replay}}_s$ compares $\{\widehat{\phi}_{s^{\prime}}\}_{e\le s}$ \emph{within the same update}; the alignment loss $\mathcal{L}_{\mathrm{align}}$ keeps replay-domain statistics representative of their historical domain.}
\label{tab:tailored_overview}
\begingroup
\setlength{\tabcolsep}{4pt}

\begin{tabular}{P{1.5cm} P{5.5cm} P{4.6cm} P{4.3cm}}
\toprule
\textbf{Method}
& \textbf{Per-domain stats \& stored prior}
& \textbf{Invariance penalty $\mathcal{P}^{\mathrm{replay}}_s$}
& \textbf{Alignment loss $\mathcal{L}_{\mathrm{align}}$} \\
\midrule

$\bigstar$-\textbf{CL-VREX}
&

$\widehat{\phi}_{s^{\prime}} \equiv \widehat{r}_{s^{\prime}}
= \mathbb{E}_{(\mathbf{x},y)\sim \sourceBPrime}\!\left[\mathcal{L}(h(\mathbf{x};\theta,\omega),y)\right]$
\newline
$\priorPrime \equiv \bar{r}_{s^{\prime}}$ (domain-level average risk)
&
$
\mathcal{P}^{\mathrm{replay}}_s
=
\frac{1}{s}\sum_{{s^{\prime}}\le s}(\widehat{r}_{s^{\prime}}-\bar{r})^2,
$
\newline
with
$\bar{r}=\frac{1}{s}\sum_{{s^{\prime}}\le s}\widehat{r}_{s^{\prime}}$
&
$
\mathcal{L}_{\mathrm{align}}
=
\sum_{e<s}(\widehat{z}_{s^{\prime}}-\bar{z}_{s^{\prime}})^2
$
\newline
with $\widehat{z}_{s^{\prime}}=h(\cdot;\theta_{s^{\prime}},\omega_{s^{\prime}})$ and $\bar{z}_{s^{\prime}}=h(\cdot,\theta_s, \omega_s)$
\\
\midrule

$\bigstar$-\textbf{CL-Fishr}
&
$\widehat{\phi}_{s^{\prime}} \equiv \widehat{\mathbf{v}}_{s^{\prime}}
= \mathrm{Var}_{(\mathbf{x},y)\sim \sourceBPrime}\!\left(\nabla_{\omega}\mathcal{L}(h(\mathbf{x};\theta,\omega),y)\right)$ 
\newline
$\priorPrime \equiv \bar{\mathbf{v}}_{s^{\prime}}$ (domain-level gradient-variance prototype via Welford's method)
&
$
\mathcal{P}^{\mathrm{replay}}_s
=
\frac{1}{s}\sum_{{s^{\prime}}\le s}\|\widehat{\mathbf{v}}_{s^{\prime}}-\bar{\mathbf{v}}\|_2^2,
$
\newline
with
$\bar{\mathbf{v}}=\frac{1}{s}\sum_{{s^{\prime}}\le s}\widehat{\mathbf{v}}_{s^{\prime}}$
&
$
\mathcal{L}_{\mathrm{align}}
=
\sum_{e<s}(1- \cos(\log(\widehat{\mathbf{v}}_{s^{\prime}}),\log(\bar{\mathbf{v}}_{s^{\prime}}))
$

\\
\midrule
$\bigstar$-\textbf{CL-CORAL}
&
$\widehat{\phi}_{s^{\prime}} \equiv (\widehat{\boldsymbol{\mu}}_{s^{\prime}},\widehat{\boldsymbol{\Sigma}}_{s^{\prime}})$ with
$\widehat{\mu}_{s^{\prime}}=\mathbb{E}_{\mathbf{x}\sim \sourceBPrime}[f_\theta(\mathbf{x})]$
and
$\widehat{\Sigma}_{s^{\prime}}=\mathrm{Cov}_{\mathbf{x}\sim \sourceBPrime}(f_\theta(\mathbf{x}))
$
\newline
$\priorPrime \equiv (\bar{\mu}_{s^{\prime}},\bar{\Sigma}_{s^{\prime}})$ (batch-level feature mean and covariances)
&
$
\mathcal{P}^{\mathrm{replay}}_s
=
\frac{1}{s}\sum_{{s^{\prime}}\le s}\Big(
\|\widehat{\mu}_{s^{\prime}}-\bar{\mu}\|_2^2+\|\widehat{\Sigma}_{s^{\prime}}-\bar{\Sigma}\|_F^2
\Big),
$
\newline
with $(\bar{\mu}, \bar{\Sigma})= \frac{1}{s}\sum_{{s^{\prime}}\le s}(\widehat{\mu}_{s^{\prime}},\widehat{\Sigma}_{s^{\prime}})$

&
$
\mathcal{L}_{\mathrm{align}}
=
\sum_{e<s}\Big(
\|\widehat{\mu}_{s^{\prime}}-\bar{\mu}_{s^{\prime}}\|_2^2+\|\widehat{\Sigma}_{s^{\prime}}-\bar{\Sigma}_{s^{\prime}}\|_F^2
\Big)
$
\\
\midrule

$\bigstar$-\textbf{CL-MMD}
&
$\widehat{\phi}_{s^{\prime}} \equiv \widehat{\mu}^z_{s^{\prime}}
= \mathbb{E}_{\mathbf{x}\sim \sourceBPrime}\!\left[z(f_\theta(\mathbf{x}))\right]$, with $z(\cdot)$ as random Fourier features for an RBF kernel
\newline
$\priorPrime \equiv \bar{\mu}^z_{s^{\prime}}$ (domain-level RFF mean embedding)
&
$
\mathcal{P}^{\mathrm{replay}}_s
=
\frac{1}{s}\sum_{{s^{\prime}}\le s}\|\widehat{\mu}^z_{s^{\prime}}-\bar{\mu}^z\|_2^2,
$
\newline
with $\bar{\mu}^z=\frac{1}{s}\sum_{{s^{\prime}}\le s}\widehat{\mu}^z_{s^{\prime}}$
&
$
\mathcal{L}_{\mathrm{align}}
=
\sum_{e<s}\|\widehat{\mu}^z_{s^{\prime}}-\bar{\mu}^z_{s^{\prime}}\|_2^2
$
\\
\midrule

$\bigstar$-\textbf{CL-ANDMask}
&
Per-domain gradient
$
g_{s^{\prime}}=\nabla_{\theta,\omega}\mathcal{L}_{\mathrm{ERM}}(\sourceBPrime;\theta,\omega)
$,
\newline
agreement mask
$
\mathbf{m}=\mathbb{I}\!\left(\left|\frac{1}{s}\sum_{{s^{\prime}}\le s}\mathrm{sign}(g_{s^{\prime}})\right|\ge\tau\right)
$
\newline with threshold $\tau \in [0,1]$
&
\emph{Implicit (via masked update)}:
$
\nabla \leftarrow \mathbf{m}\odot \frac{1}{s}\sum_{{s^{\prime}}\le s} g_{s^{\prime}},
$
\newline
with $g_{s^{\prime}}=\nabla_{\theta,\omega}\mathcal{L}_{\mathrm{ERM}}(\sourceBPrime;\theta,\omega)$
\newline
(i.e., $\mathcal{P}^{\mathrm{replay}}_s$ is realized by enforcing cross-domain sign agreement).
&
$
\mathcal{L}_{\mathrm{align}}
=
\sum_{e<s}\mathbb{E}_{(\mathbf{x},y,\mathbf{z})\sim \sourceBPrime}
\Big[\big\|h(\mathbf{x};\theta,\omega)-\mathbf{z}\big\|_2^2\Big]
$
\\
\bottomrule
\end{tabular}
\endgroup
\end{table*}

\textbf{Invariance penalty:} To compute the invariance penalty, the na\"ive extensions maintain a set of domain-specific invariance priors $\{\priorPrime\}_{{s^{\prime}} < s}$, where each $\priorPrime$ is computed from $\sourceDPrime$ at the end of training on domain ${s^{\prime}}$. $\priorPrime$ thus stores domain-specific summary statistics and takes the form
\begin{equation}
\hspace{-2mm}
        \priorPrime \;=\; \textsc{Aggregate}\big(\phi(\mathbf{x},y;\theta,\omega)\;\; \text{for } (\mathbf{x},y)\sim \sourceDPrime\big)
        \label{eq:prior_def}
\end{equation}
where $\phi(\cdot)$ is a function that extracts method-specific invariance information (e.g., feature moments \citep{sun2016deep}, risk \citep{krueger2021out}, gradient statistics \citep{rame2022fishr}), and \textsc{Aggregate} denotes a running estimator using Welford's method \citep{welford1962note} to efficiently compute the average statistics for a domain.

\textbf{Training:} During training on minibatches from domain $\sourceD$, the na\"ive extensions penalize differences between current invariance statistics and stored priors $\priorPrime$ as
\begin{equation}
        \mathcal{P}_s(\theta,\omega)
        \;=\;
        \frac{1}{s-1}\sum_{{s^{\prime}}=1}^{s-1}
        d\!\left(\widehat{\phi}_s(\theta,\omega),\, \priorPrime\right),
        \label{eq:prior_matching}
\end{equation}
where $\widehat{\phi}_s(\cdot)$ extracts method-specific invariance information from the current domain-$s$ batch, and $d(\cdot,\cdot)$ is a distance function such as $\ell_2$. The na\"ive extensions thus implement $\mathcal{P}_s$ by comparing a domain-$s$ quantity (estimated from the current minibatch from domain $s$) to \emph{stored} quantities from previous domains. Intuitively, Eq.~(\ref{eq:prior_matching}) treats each past domain as an ``anchor'' and encourages the current domain to match previously observed invariance statistics.

\textbf{Na{\"i}ve extensions:} We construct the following five na\"ive extensions, which use different methodology to compute $\widehat{\phi}_s$ and $\priorPrime$: \textbf{(1)}~Na{\"i}ve-CL-VREX, based on \citet{krueger2021out}, which uses training risk as invariance information; \textbf{(2)} Na{\"i}ve-CL-Fishr, based on \citet{rame2022fishr}, which uses gradient statistics; \textbf{(3)} Na{\"i}ve-CL-CORAL, based on \citet{sun2016deep}, which uses second-moment estimates from latent features; \textbf{(4)} Na{\"i}ve-CL-MMD, based on \citet{li2018domain}, which uses Hilbert space kernel transformation on features; and \textbf{(5)} Na{\"i}ve-CL-ANDMask, based on \citet{parascandolo2020learning}, which uses gradient sign agreement information. We provide further details in \cref{sec:naive_details}.

As we empirically show later, these na\"ive algorithms achieve only minor improvements over standard continual fine-tuning. We attribute this to the fact that the stored static summaries can only \emph{approximate} the real domain-invariance methods. The reason is that the original DIRL methods were designed for \textit{joint} access to all source domains, which the summaries fail to faithfully reproduce. These limitations motivate our tailored methods

\section{Continual learning of domain-invariant representations}\label{sec:our_methods}

To overcome the limitations of the na{\"i}ve methods, we propose a set of tailored methods that explicitly learn and preserve domain-invariant structure over time. Novel to our methods is that these (i)~perform multi-domain invariance computation via replay and (ii) prevent the degradation of invariance information through domain-conditioned alignment. Overall, we provide five different methods (i.e.,$\bigstar$-\textbf{CL-VREX}, $\bigstar$-\textbf{CL-Fishr}, $\bigstar$-\textbf{CL-CORAL}, $\bigstar$-\textbf{CL-MMD}, $\bigstar$-\textbf{CL-ANDMask}). An overview of our proposed methods is provided in \Cref{tab:tailored_overview}.

\subsection{Components of our methods}
Our proposed methods utilize the following components to sequentially learn domain-invariant representations: \tailoredcircled{1} {memory buffers}, \tailoredcircled{2} {replay-augmented ERM}, \tailoredcircled{3} {invariance computation via replay}, and \tailoredcircled{4} {domain-conditioned invariance alignment}. In the following, we detail each component, while the specific instantiation for each $\bigstar$-CL method is given in \cref{tab:tailored_overview}.

\tailoredcircled{1} \textbf{Memory buffer.}
We maintain a memory buffer $M$ with domain-conditioned partitions $M_{s^{\prime}}$ for each previously seen source domain ${s^{\prime}} < s$. Each stored element is a triple $(\mathbf{x},y,\mathbf{z})$ where $\mathbf{z}$ are additional, method-specific information produced at insertion time. For example, $\mathbf{z} = h(\mathbf{x};\theta_{s^{\prime}},\omega_{s^{\prime}})$ (logits) or $\mathbf{z} = f_{\theta_{s^{\prime}}}(\mathbf{x})$ (features), where $(\theta_{s^{\prime}},\omega_{s^{\prime}})$ denotes the model parameters at time ${s^{\prime}}$. We later use $\mathbf{z}$ for aligning invariances across time.

\tailoredcircled{2} \textbf{ERM with replay.}
We expand the ERM term from Eq.~(\ref{eq:na\"ive_template}) via
\begin{equation}
    \mathcal{L}_{\mathrm{ERM}}^{\mathrm{replay}}(\theta,\omega)
    =
    \mathbb{E}_{(\mathbf{x},y)\sim B}\!\left[\mathcal{L}(h(\mathbf{x};\theta,\omega),y)\right],
    \label{eq:erm_replay}
\end{equation}
where $B$ is a joint data batch from the current domain $s$, $B_s \sim \sourceD$, and replay domains ${s^{\prime}}$, $\sourceBPrime \sim M_{{s^{\prime}}} \;\;\text{for}\;\; {s^{\prime}}=1,\dots,s-1$, constructed by $B \;=\; \bigcup_{e\le s} B_{e}$.

\tailoredcircled{3} \textbf{Multi-domain invariance computation via replay.}
For each domain, we define the invariance penalty on the corresponding minibatch $\sourceBPrime$. Let $\widehat{\phi}_{s^{\prime}}(\theta,\omega;\sourceBPrime)$ denote the per-domain statistic (e.g., risk \citep{krueger2021out}, moments \citep{sun2016deep}, gradient statistics \citep{rame2022fishr}, etc.) computed on $\sourceBPrime$. We then instantiate the invariance penalty over $\{\sourceBPrime\}_{{s^{\prime}}\le s}$\footnote{This includes the current domain by design, hence ${s^{\prime}}\boldsymbol{\le}s$.} as
\begin{equation}
\hspace{-2mm}
    \mathcal{P}_s^{\mathrm{replay}}(\theta,\omega)
    =
    \textsc{InvPenalty}\Big(\big\{\widehat{\phi}_{s^{\prime}}(\theta,\omega;\sourceBPrime)\big\}_{{s^{\prime}}\le s}\Big)
    \label{eq:replay_penalty_generic}
\end{equation}
which compares domain quantities \emph{within the same update}, rather than via static loss priors.

\tailoredcircled{4} \textbf{Domain-conditioned invariance alignment.}
The previous Eq.~(\ref{eq:replay_penalty_generic}) recovers within-step comparisons, but limited replay can make estimates noisy and can cause domain identity drift of replayed statistics as $h$ evolves through subsequent training. We therefore utilize knowledge distillation \citep{hinton2015distilling} and reuse stored invariance priors $\{\priorPrime\}_{{s^{\prime}}<s}$ from Eq.~(\ref{eq:prior_def}) to align replay-domain statistics to their original domain references:
\begin{equation}
    \mathcal{L}_{\mathrm{align}}(\theta,\omega)
    =
    \sum_{{s^{\prime}}<s}
    d\!\left(\widehat{\phi}_{s^{\prime}}(\theta,\omega;\sourceBPrime),\,\priorPrime\right),
    \label{eq:align_loss}
\end{equation}
where $d(\cdot, \cdot)$ is a distance function as in Eq.~(\ref{eq:prior_matching}). However, in contrast to Eq.~(\ref{eq:prior_matching}), Eq.~(\ref{eq:align_loss}) does \emph{not} anchor domain statistics to the past; instead, it keeps replayed samples representative of their historical domain by sampling batches $\sourceBPrime$ from replay buffer $M_{{s^{\prime}}}$.

\textbf{Overall objective.}
Our overall objective combines replay-augmented ERM, sequential invariance penalties, and invariance alignment to give the generic training objective at domain $s$:
\begin{equation}
    \min_{\theta,\omega}\;
    \mathcal{L}_{\mathrm{ERM}}^{\mathrm{replay}}(\theta,\omega)
    +
    \lambda\,\mathcal{P}_s^{\mathrm{replay}}(\theta,\omega)
    +
    \beta\,\mathcal{L}_{\mathrm{align}}(\theta,\omega),
    \label{eq:tailored_objective}
\end{equation}
where $\lambda,\beta\ge 0$ are scalar weights.

\subsection{Proposed  methods}
In total, we instantiate five different methods which differ in (i) the statistics $\phi(\cdot)$, (ii) the invariance operator \textsc{InvPenalty}$(\cdot)$, and (iii) the alignment distance $d(\cdot,\cdot)$. Our methods are (1) $\bigstar$-\textbf{CL-VREX}, where $\phi$ is the domain-wise empirical risk and \textsc{InvPenalty} is the risk variance across domains; (2) $\bigstar$-\textbf{CL-Fishr}, where $\phi$ is the per-domain gradient variance on $\omega$ and \textsc{InvPenalty} matches these variances across domains in the update; (3) $\bigstar$-\textbf{CL-CORAL}, where $\phi$ uses feature moments $(\mu_{s^{\prime}},\Sigma_{s^{\prime}})$ that \textsc{InvPenalty} matches across domains; (4) $\bigstar$-\textbf{CL-MMD}, where $\phi$ is a kernel mean embedding that \textsc{InvPenalty} matches across domains; and (5) $\bigstar$-\textbf{CL-ANDMask}, where $\phi$ is the set of per-domain gradients and \textsc{InvPenalty} keeps only gradients whose signs agree across $\{\sourceBPrime\}_{{s^{\prime}}\le s}$. 

For all methods, we chose the alignment term in Eq.~(\ref{eq:align_loss}) to match the corresponding invariance statistic, ensuring that replayed samples remain consistent with their original domain invariance signature.
In \cref{tab:tailored_overview}, we provide a detailed overview of the methods and the method-specific equations. We provide minimal corresponding pseudocodes in \cref{alg:ours_condensed}. We offer further details and extended pseudocodes in \cref{sec:ours_details,alg:tailored_seq_dirl}, respectively.

\begin{algorithm}[h]
\scriptsize
\captionsetup{font=scriptsize}
\caption{
{\setlength{\fboxsep}{0pt}
Pseudocode for \colorbox{color_tailored}{our $\bigstar$-CL methods}. See Alg.~(\ref{alg:tailored_seq_dirl}) for extended codes.
}
}
\label{alg:ours_condensed}
\begin{algorithmic}[1]
\REQUIRE Domain sequence $S$, memory buffer $M$
\FOR{each domain $D_s$}
    \FOR{each batch in $D_s \cup M$}
        \STATE Compute $\mathcal{L}_{\mathrm{ERM}}^{\mathrm{replay}}$
        \STATE Compute invariances via $\mathcal{P}_s^{\mathrm{replay}}$
        \STATE Align invariances to original reference via $\mathcal{L}_{\mathrm{align}}$
    \ENDFOR
\ENDFOR
\end{algorithmic}
\end{algorithm}

\section{Experimental setup}\label{sec:experimental_setup}
\textbf{Datasets.} We use six datasets that are commonly used in the literature \citep[e.g.,][]{koh2021wilds,shi2023unified,liang2023loss,sarfraz025semantic}: {(1)~RotatedMNIST} \cite{ghifary2015domain}, which is a variation of MNIST \cite{lecun1998mnist} with multiple domains that correspond to different rotations of the original images. {(2)~CIFAR10C} \citep{hendrycks2019benchmarking} is a variation of CIFAR10 \citep{Krizhevsky09learningmultiple}, but with multiple domains where each is subject to different corruptions of the input data. {(3)~TinyImageNetC} \citep{hendrycks2019benchmarking} is a similar modification of TinyImageNet \citep{le2015tiny}, where domains are created by applying different corruptions to the input pixels. {(4)~Camelyon17}~\citep{bandi2018detection,koh2021wilds} is a real-world clinical image dataset of lymph node sections collected at different hospitals. {(5)~WM811K}~\citep{wafermap,wu2015wafer}, a large-scale dataset from wafer manufacturing. {(6)~Covertype}~\citep{blackard1998covertype}, a dataset of cartographic and environmental features for predicting forest cover type. Details about the dataset, including source and target domain splits, are in \cref{sec:dataset_details}.

\textbf{Network architectures.}
Following existing works in CL \citep{buzzega2020dark,liang2023loss}, we use an ImageNet-pretrained ResNet18 \citep{he2016deep} model for all experiments with large image datasets. For RotatedMNIST, we use a four-layer CNN following \citet{rame2022fishr}. For Covertype, we use a four-layer MLP.

\textbf{Baselines.}
We compare our proposed methods against 13 strong and widely used CL 
and three CDA/CTTA baselines:
$\bullet$\,\emph{optimization-based methods:} \textbf{AGEM}~\citep{chaudry2019efficient} and \textbf{UPGD}~\citep{elsayed2024addressing}; 
$\bullet$\,\emph{regularization-based methods:} \textbf{EWC}~\citep{kirkpatrick2017overcoming}, \textbf{SI}~\citep{zenke2017continual}, and \textbf{SNR}~\citep{farias2024self}; $\bullet$\,\emph{sample-replay-based methods:} \textbf{FDR}~\citep{benjamin2019measuring}, \textbf{ER-ACE}~\citep{caccia2021new}, \textbf{LODE}~\citep{liang2023loss}, and \textbf{STAR}~\citep{eskandar2025star}; 
$\bullet$\,\emph{latent-information replay-based methods:} \textbf{CoPE}~\citep{de2021continual}, \textbf{EFC}~\citep{magistri2024elastic}, and \textbf{SARL}~\citep{sarfraz025semantic};
$\bullet$\,\emph{CDA/CTTA methods:} \textbf{TENT}~\citep{wang2021tent}, \textbf{SHOT++}~\citep{liang2021source}, and \textbf{CoTTA}~\citep{wang2022continual}.
We also benchmark against sequential fine-tuning, referred to as \textbf{Finetune}. As an upper-bound, we use \textbf{URM}~\citep{krishnamachari2024uniformly} in a \textit{offline setting.} URM has joint access to all source domains during training.

\textbf{Experimental settings.}
We report the average classification accuracy (Acc) on the target domain $\pm$standard error across three runs. For WM811K, we report the macro F1 score$\pm$standard error. In comparisons, we report performance differences in \%-points as \emph{pp}. For replay-based methods, we follow prior work \citep{liang2023loss} and set the memory buffer to $1000$ (for RotatedMNIST and Covertype) or $5000$ (remaining datasets) samples. Details about hyperparameter optimization are in \cref{sec:hparam_details}. All experiments are conducted on NVIDIA A100 and H100 GPU slices (MIG mode) with 32GB RAM and four CPU cores.

\section{Results}
\circled{1} We first show the main experimental results for our proposed $\bigstar$-CL methods, followed by several ablation and robustness studies. \naivecircled{2} We then demonstrate the subpar performance of the na\"ive extensions. \tailoredcircled{3} Afterwards, we show the improvements from our methods.

\begin{table*}[!tbp]
\caption{\textbf{Main results:}
{\setlength{\fboxsep}{0pt}
Comparison of \colorbox{color_tailored}{our proposed methods} against state-of-the-art CL baselines (\colorbox{color_finetune}{Finetune}, \colorbox{color_optimization}{optimization}-, \colorbox{color_regularization}{regularization}-, and \colorbox{color_replay}{replay}-based methods) on six benchmark datasets. We report the mean$\pm$standard error \emph{target domain performance} across three independent runs. Results marked as \textbf{best}, \underline{second}, \dotuline{third}. ``{---}'' denotes \textit{not applicable}.
}
}
\label{tab:main_results}
\adjustbox{max width=\textwidth}{%
\begin{tabular}{l|ccccccc|ccc}
\toprule
& \multicolumn{7}{c|}{Accuracy / F1$^\dagger$ ($\uparrow$)} & \multicolumn{3}{c}{Rank ($\downarrow$)} \\
\textbf{Method}                            & \textbf{RotatedMNIST}                      & \textbf{CIFAR10C}                          & \textbf{TinyImageNetC}                                               & \textbf{WM811K}                           & \textbf{Covertype}                      & \textbf{Camelyon17}                     & \textbf{Avg}                               & \textbf{Arith. mean}                       & \textbf{Geom. mean}                        & \textbf{Median}                            \\
\midrule
\rowcolor{color_finetune} Finetune                                        & 39.5 $\pm$ 1.7                                                            & 66.3 $\pm$ 0.4                                                            & 18.5 $\pm$ 2.6                                                            & 82.8 $\pm$ 0.3                                                            & 8.1 $\pm$ 0.0                                                             & 86.9 $\pm$ 2.1                                                            & 50.4                                                                      & 14.3                                                                      & 14.0                                                                      & 15.0                                                                      \\
\midrule
\rowcolor{color_optimization} AGEM                                        & 54.8 $\pm$ 4.7                                                            & \underline{69.2 $\pm$ 0.2}                                                & 22.8 $\pm$ 1.1                                                            & 84.1 $\pm$ 0.2                                                            & 29.0 $\pm$ 8.5                                                            & \dotuline{91.0 $\pm$ 0.1}                                                 & 58.5                                                                      & 7.5                                                                       & 6.2                                                                       & 8.5                                                                       \\
\rowcolor{color_optimization} UPGD                                        & 29.1 $\pm$ 7.7                                                            & 53.9 $\pm$ 4.6                                                            & 19.6 $\pm$ 2.2                                                            & 74.6 $\pm$ 1.1                                                            & 23.4 $\pm$ 9.6                                                            & 88.7 $\pm$ 1.0                                                            & 48.2                                                                      & 16.0                                                                      & 15.8                                                                      & 16.5                                                                      \\
\midrule
\rowcolor{color_regularization} EWC                                       & 42.5 $\pm$ 1.3                                                            & 66.7 $\pm$ 1.3                                                            & 20.1 $\pm$ 0.6                                                            & 84.2 $\pm$ 0.4                                                            & 8.0 $\pm$ 0.2                                                             & 90.7 $\pm$ 0.6                                                            & 52.0                                                                      & 10.8                                                                      & 9.8                                                                       & 10.5                                                                      \\
\rowcolor{color_regularization} SI                                        & 38.8 $\pm$ 0.7                                                            & 64.3 $\pm$ 2.4                                                            & 20.8 $\pm$ 1.2                                                            & 83.3 $\pm$ 0.6                                                            & 24.5 $\pm$ 6.8                                                            & 86.0 $\pm$ 3.6                                                            & 52.9                                                                      & 14.2                                                                      & 14.1                                                                      & 13.5                                                                      \\
\rowcolor{color_regularization} SNR                                       & 40.4 $\pm$ 0.2                                                            & 62.9 $\pm$ 1.6                                                            & 22.0 $\pm$ 1.7                                                            & 76.2 $\pm$ 0.6                                                            & 25.4 $\pm$ 10.4                                                           & 89.7 $\pm$ 0.9                                                            & 52.8                                                                      & 13.3                                                                      & 13.0                                                                      & 14.0                                                                      \\
\midrule
\rowcolor{color_replay} FDR                                               & 68.6 $\pm$ 1.3                                                            & 65.0 $\pm$ 0.9                                                            & 23.0 $\pm$ 1.0                                                            & 84.0 $\pm$ 0.4                                                            & 38.3 $\pm$ 0.7                                                            & 88.5 $\pm$ 1.0                                                            & 61.2                                                                      & 8.5                                                                       & 8.1                                                                       & 8.5                                                                       \\
\rowcolor{color_replay} ER-ACE                                            & 67.5 $\pm$ 2.0                                                            & 66.9 $\pm$ 1.1                                                            & \textbf{29.0 $\pm$ 1.2}                                                   & \underline{85.4 $\pm$ 0.4}                                                & 37.7 $\pm$ 0.6                                                            & 90.3 $\pm$ 1.5                                                            & 62.8                                                                      & 5.2                                                                       & 4.1                                                                       & 6.5                                                                       \\
\rowcolor{color_replay} LODE                                              & 60.0 $\pm$ 2.1                                                            & 64.7 $\pm$ 0.6                                                            & 21.6 $\pm$ 2.3                                                            & 80.1 $\pm$ 1.7                                                            & 37.5 $\pm$ 1.1                                                            & 65.2 $\pm$ 5.1                                                            & 54.9                                                                      & 12.5                                                                      & 12.1                                                                      & 11.5                                                                      \\
\rowcolor{color_replay} STAR                                              & 67.4 $\pm$ 1.4                                                            & \textbf{69.5 $\pm$ 0.3}                                                   & 25.6 $\pm$ 1.5                                                            & \dotuline{85.1 $\pm$ 0.7}                                                 & 38.3 $\pm$ 1.6                                                            & 86.9 $\pm$ 2.7                                                            & 62.1                                                                      & 6.5                                                                       & 4.8                                                                       & 5.5                                                                       \\
\rowcolor{color_replay} COPE                                              & 68.7 $\pm$ 3.2                                                            & 64.2 $\pm$ 0.7                                                            & 20.1 $\pm$ 4.9                                                            & 84.0 $\pm$ 0.7                                                            & 9.4 $\pm$ 1.4                                                             & 90.6 $\pm$ 0.7                                                            & 56.2                                                                      & 11.0                                                                      & 10.0                                                                      & 12.0                                                                      \\
\rowcolor{color_replay} EFC                                               & 65.8 $\pm$ 3.1                                                            & 66.7 $\pm$ 0.8                                                            & 21.3 $\pm$ 2.7                                                            & 84.7 $\pm$ 1.3                                                            & 25.8 $\pm$ 9.1                                                            & 90.9 $\pm$ 0.8                                                            & 59.2                                                                      & 8.3                                                                       & 7.7                                                                       & 9.0                                                                       \\
\rowcolor{color_replay} SARL                                              & 63.9 $\pm$ 1.8                                                            & 62.9 $\pm$ 0.8                                                            & 0.7 $\pm$ 0.1                                                             & 75.5 $\pm$ 1.4                                                            & \dotuline{41.2 $\pm$ 5.6}                                                 & 79.6 $\pm$ 4.6                                                            & 54.0                                                                      & 13.7                                                                      & 11.8                                                                      & 16.5                                                                      \\
\midrule
\rowcolor{color_tailored} $\bigstar$-CL-VREX                              & 70.2 $\pm$ 5.8                                                            & 67.2 $\pm$ 1.4                                                            & \dotuline{26.3 $\pm$ 1.2}                                                 & 84.1 $\pm$ 0.3                                                            & 40.8 $\pm$ 1.5                                                            & \underline{91.5 $\pm$ 0.6}                                                & \underline{63.4}                                                          & \underline{4.3}                                                           & \dotuline{4.0}                                                            & \dotuline{4.0}                                                            \\
\rowcolor{color_tailored} $\bigstar$-CL-Fishr                             & 68.6 $\pm$ 0.4                                                            & 64.2 $\pm$ 1.2                                                            & \underline{29.0 $\pm$ 2.2}                                                & 83.6 $\pm$ 0.5                                                            & 27.0 $\pm$ 7.7                                                            & 89.7 $\pm$ 2.0                                                            & 60.3                                                                      & 9.2                                                                       & 7.8                                                                       & 10.0                                                                      \\
\rowcolor{color_tailored} $\bigstar$-CL-CORAL                             & \textbf{72.8 $\pm$ 2.6}                                                   & 68.5 $\pm$ 1.7                                                            & 25.0 $\pm$ 2.8                                                            & 84.8 $\pm$ 0.6                                                            & \textbf{45.2 $\pm$ 3.9}                                                   & \textbf{91.7 $\pm$ 0.4}                                                   & \textbf{64.7}                                                             & \textbf{2.8}                                                              & \textbf{2.1}                                                              & \textbf{2.5}                                                              \\
\rowcolor{color_tailored} $\bigstar$-CL-MMD                               & \dotuline{70.7 $\pm$ 0.5}                                                 & \dotuline{69.0 $\pm$ 0.5}                                                 & 25.8 $\pm$ 0.8                                                            & \textbf{85.5 $\pm$ 0.4}                                                   & 37.6 $\pm$ 0.9                                                            & 90.1 $\pm$ 0.5                                                            & \dotuline{63.1}                                                           & \dotuline{4.5}                                                            & \underline{3.6}                                                           & \underline{3.5}                                                           \\
\rowcolor{color_tailored} $\bigstar$-CL-ANDMask                           & \underline{71.4 $\pm$ 2.6}                                                & 64.7 $\pm$ 0.7                                                            & 11.8 $\pm$ 3.6                                                            & 84.6 $\pm$ 0.6                                                            & \underline{43.7 $\pm$ 2.9}                                                & 89.1 $\pm$ 1.5                                                            & 60.9                                                                      & 8.3                                                                       & 6.1                                                                       & 8.5                                                                       \\
\midrule
URM (Upper Bound)                                                                      & 81.3 $\pm$ 1.4                                                            & {70.6 $\pm$ 0.7}                                                   & {31.8 $\pm$ 0.3}                                                   & {91.3 $\pm$ 0.4}                                                   & {---}                                                                         & {93.1 $\pm$ 0.8}                                                   & {73.3}                                                                         & {---}                                                                          & {---}                                                                          & {---}                                                                                                                                                                                                                         \\
\bottomrule
\end{tabular}}
\tiny{$^\dagger$ macro F1 for WM811K}
\end{table*}

\circled{1} \textbf{Main results.} We now perform the main experiment and benchmark our proposed methods against strong CL baselines. The results are in \cref{tab:main_results}. We find: \textbf{(1)}~Our methods achieve the first, second-, and third-best overall performance with \SI{64.7}{\percent}, \SI{63.4}{\percent}, and \SI{63.1}{\percent} accuracy / F1, respectively. \textbf{(2)}~Our $\bigstar$-CL methods outperform optimization-, regularization-, and replay-based baselines by \num{6} pp, \num{10} pp, and \num{2} pp, respectively. \textbf{(3)}~Our methods perform best in terms of rank: overall, the best method by rank are $\bigstar$-CL-CORAL, $\bigstar$-CL-MMD, $\bigstar$-CL-VREX. $\Rightarrow$ \textbf{Takeaway}: \emph{Our $\bigstar$-CL methods reach SOTA in terms of \underline{both average performance and average rank}.}

\begin{figure}
    \centering
    \includegraphics[width=0.75\linewidth]{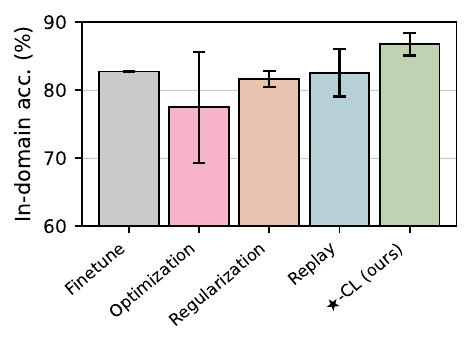}
    \vspace{-5pt}
    \caption{\setlength{\fboxsep}{0pt}{
       \textbf{In-domain performance.} Average across six datasets, per method group. The large std. err. (whisker) for \colorbox{color_optimization}{optimization} is caused by UPGD, which reaches a low in-domain performance. Our proposed methods outperform \colorbox{color_finetune}{Finetune} and strong \colorbox{color_regularization}{regularization} and \colorbox{color_replay}{replay CL} baselines.}}
    \label{fig:source_performance}
    \vspace{-10pt}
\end{figure}

\circled{$\bullet$}\,\textbf{Comparison against CDA/CTTA baselines.} We compare against CDA and CTTA methods in \cref{tab:cda_ctta_results}. For these baselines, we allow \textit{unsupervised updates on the target domain}. Still, our proposed methods outperform the baselines by a large margin (up to 10pp). $\Rightarrow$ \textbf{Takeaway}: \emph{Our $\bigstar$-CL methods outperform CDA and CTTA baselines.}

\circled{$\bullet$}\,\textbf{In-domain performance.}
We report the in-domain performance in \cref{fig:source_performance}, with details in \Cref{sec:source_performance}. We find: \textbf{(1)}~Our proposed methods outperform strong CL baselines. \textbf{(2)}~Nearly all methods attain sufficient in-domain performance, i.e., they have similar ``starting conditions''. The better \emph{out-of-domain} generalization from our methods must thus come from their ability to better learn the underlying causal mechanisms. \textbf{Takeaway}: \emph{The use of DIRL benefits in-domain performance.}

\circled{$\bullet$}\,\textbf{Forgetting.}
We report forgetting, as measured through the common backwards transfer (BWT) metric \citep{lopez2017gradient}, in \Cref{tab:bwt_results}. Negative numbers imply forgetting; positive numbers indicate retrospective improvement (which is desired). The results show that our proposed methods retain or retrospectively improve the performance on previous domains. \textbf{Takeaway}: \emph{Our methods retrospectively improve the performance.}

\circled{$\bullet$}\,\textbf{Runtime analysis.}
We provide runtimes in \Cref{tab:runtime_results}. The runtimes for our methods are in line with existing CL methods. \textbf{Takeaway}: \emph{Runtimes are similar to the baselines, yet our methods generalize better to unseen domains.}

\circled{$\bullet$}\,\textbf{Ablation: invariance alignment}. We here study the effect of $\mathcal{L}_{\mathrm{align}}$ via (1) disabling alignment by setting $\beta = 0$, and (2) dynamically recomputing $\priorPrime$ at the end of each domain. The results are in \Cref{tab:abl_no_inv_align} and \Cref{tab:abl_dyn_anchors}. We find that aligning the invariance statistics over time is important for the final performance. This is an interesting finding, as aligning statistics to old domains is often used to \emph{prevent forgetting}. However, our results indicate that aligning invariance statistics is equally important for \emph{generalization} to unseen domains. $\Rightarrow$ \textbf{Takeaway}: \emph{Aligning invariance statistics over time improves generalization to unseen domains.}

\begin{figure}[!tbp]
\centering
    \begin{subfigure}{.45\linewidth}
        \centering
        \includegraphics[width=0.9\linewidth]{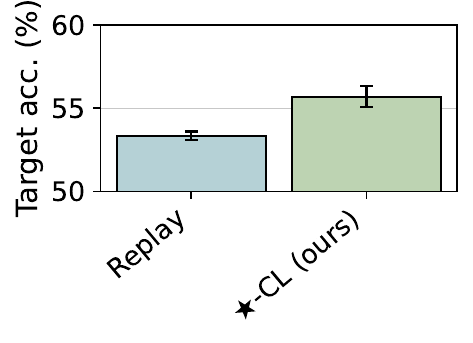}
    \end{subfigure}%
    \begin{subfigure}{.45\linewidth}
        \centering
        \includegraphics[width=0.9\linewidth]{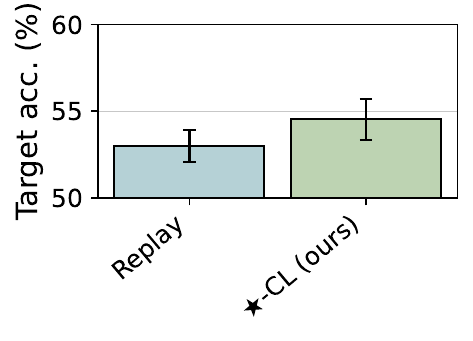}
    \end{subfigure}
    \vspace{-5pt}
    \caption{
    {
    \setlength{\fboxsep}{0pt}
    \textbf{Reduced buffer sizes.}
    Average results for \SI{50}{\percent} (left) and \SI{25}{\percent} (right) memory capacity, providing less domain information. \colorbox{color_tailored}{Our methods} can nonetheless learn domain-invariant representations and outperform strong \colorbox{color_replay}{replay-baselines.
    }
    }}
    \label{fig:robustness_plots}
    \vspace{-7pt}
\end{figure}

\circled{$\bullet$}\,\textbf{Sensitivity: reduced buffer sizes.}
Prior work in CL has shown that, for replay-based methods, the buffer size strongly influences the final performance \citep[e.g.,][]{aljundi2019online,aljundi2019gradient}, where larger buffers lead to better performance. We thus now make the problem \emph{more challenging} and reduce the buffer capacity to \SI{50}{\percent} and \SI{25}{\percent}, respectively. We benchmark our proposed methods against best-performing replay-methods (FDR, ER-ACE, and STAR). To ensure fair comparison, we perform new hyperparameter searches. The results are in \cref{fig:robustness_plots,tab:robustness_50p_buffer,tab:robustness_25p_buffer}. We observe: \textbf{(1)} Our methods again reach the best performance, both in terms of average performance and position in the ranking. \textbf{(2)}~Our proposed methods outperform strong replay baselines by $\sim$\num{4} pp. \textbf{(3)}~Our findings are consistent across the different buffer sizes. $\Rightarrow$ \textbf{Takeaway}: \emph{Our methods are robust to the memory buffer size and consistently outperform strong CL baselines.}

\circled{$\bullet$}\,\textbf{Robustness: different target domain.}
We now exchange the dataset-specific target domain. We benchmark our proposed methods against baselines AGEM, UPGD, EWC, SI, FDR, and STAR. Note that, for fairness, exchanging the target domain necessitates that we again perform a new hyperparameter search. The results are in \Cref{tab:robustness_different_target_domain}, details in \Cref{sec:robustness_different_target_domain}. Our methods again outperform optimization-based (UPGD, AGEM) and regularization-based (EWC and SI) methods and reach the best overall average performance. $\Rightarrow$ \textbf{Takeaway}: \emph{Our methods outperform existing CL baselines on different target domains.}

\circled{$\bullet$}\,\textbf{Robustness: different compute budget.}
We now make learning robust representations even more challenging by reducing the number of training steps per domain by 50\%, which also limits the number of samples per domain. We benchmark against AGEM, UPGD, EWC, SI, FDR, and STAR. As before, we perform a new hyperparameter study. The results are in \Cref{tab:different_compute_budget}. Our proposed methods outperform Finetune, optimization- and regularization-based methods by a clear margin and reach the best overall performance. The results show that, under reduced compute budgets, our methods can sequentially learn representations that better generalize to unseen target domains. $\Rightarrow$ \textbf{Takeaway}: \emph{Our results are robust to different compute budgets.}

\circled{$\bullet$}\,\textbf{Visualizations}
We visualize the $\lambda, \beta$-hyperparameter sensitivity in \cref{fig:sensitivity_1,fig:sensitivity_2}. We observe that our proposed methods are relatively robust to changes. Further, in \cref{fig:feature_space_tripple,fig:feature_space_double}, we plot the feature space learned by our methods. We observe a \textbf{high representational overlap between the source and (unseen) target domains}, indicating that our methods successfully learn domain-invariant representations.

\begin{table}
\centering
\caption{
{\setlength{\fboxsep}{0pt}
Results for \colorbox{color_naive}{na\"ive extensions}, which highlights the limited improvement over the \colorbox{color_finetune}{default sequential training} baseline. Further, na{\"i}ve methods underperform baselines \colorbox{color_optimization}{AGEM}, \colorbox{color_replay}{ER-ACE}, and \colorbox{color_regularization}{SI}. Shown: mean$\pm$standard error target domain accuracy across three independent runs. Marked: \textbf{best}, \underline{second}, \dotuline{third}.
}
}
\label{tab:naive_results_cropped}
\adjustbox{max width=\linewidth}{%
\begin{tabular}{lcccc|c}
\toprule
& \multicolumn{4}{c|}{Accuracy ($\uparrow$)} & $\Delta$ ($\uparrow$)\\
\midrule
\textbf{Method}                            & \textbf{RotatedMNIST}                      & \textbf{CIFAR10C}                          & \textbf{Covertype}                         & \textbf{Avg}                               & \textbf{to Finetune}                      \\
\midrule
\rowcolor{color_finetune} Finetune                                        & 39.5 $\pm$ 1.7                                                            & 66.3 $\pm$ 0.4                                                            & 8.1 $\pm$ 0.0                                                             & 38.0                                                                      & +0.0                                                                      \\
\midrule
\rowcolor{color_optimization} AGEM                                        & \underline{54.8 $\pm$ 4.7}                                                & \textbf{69.2 $\pm$ 0.2}                                                   & \underline{29.0 $\pm$ 8.5}                                                & \underline{51.0}                                                          & \underline{+13.0}                                                         \\
\midrule
\rowcolor{color_regularization} SI                                        & 38.8 $\pm$ 0.7                                                            & 64.3 $\pm$ 2.4                                                            & \dotuline{24.5 $\pm$ 6.8}                                                 & \dotuline{42.5}                                                           & \dotuline{+4.5}                                                           \\
\midrule
\rowcolor{color_replay} ER-ACE                                            & \textbf{67.5 $\pm$ 2.0}                                                   & 66.9 $\pm$ 1.1                                                            & \textbf{37.7 $\pm$ 0.6}                                                   & \textbf{57.4}                                                             & \textbf{+19.4}                                                            \\
\midrule
\rowcolor{color_naive} Na{\"i}ve-CL-VREX                                & 45.8 $\pm$ 6.8                                                            & \dotuline{67.4 $\pm$ 0.3}                                                 & 8.1 $\pm$ 0.1                                                             & 40.4                                                                      & +2.4                                                                      \\
\rowcolor{color_naive} Na{\"i}ve-CL-Fishr                               & 41.4 $\pm$ 1.5                                                            & \underline{67.7 $\pm$ 0.3}                                                & 8.2 $\pm$ 0.1                                                             & 39.1                                                                      & +1.1                                                                      \\
\rowcolor{color_naive} Na{\"i}ve-CL-CORAL                               & \dotuline{50.2 $\pm$ 5.3}                                                 & 66.3 $\pm$ 0.8                                                            & 8.0 $\pm$ 0.1                                                             & 41.5                                                                      & +3.5                                                                      \\
\rowcolor{color_naive} Na{\"i}ve-CL-MMD                                 & 39.9 $\pm$ 1.0                                                            & 66.3 $\pm$ 0.8                                                            & 7.8 $\pm$ 0.3                                                             & 38.0                                                                      & +0.0                                                                      \\
\rowcolor{color_naive} Na{\"i}ve-CL-ANDMask                             & 21.5 $\pm$ 4.5                                                            & 26.9 $\pm$ 1.4                                                            & 6.6 $\pm$ 1.2                                                             & 18.3                                                                      & -19.7                                                                     \\
\bottomrule
\end{tabular}}
\vspace{-7pt}
\end{table}

\naivecircled{2} \textcolor{color_naive_border}{\textbf{{Na{\"i}ve extensions fail to meaningfully outperform Finetune baseline.}}} We benchmark the na{\"i}ve extensions (labeled by Na{\"i}ve-CL) against Finetune and a subset of the baselines (i.e., best-performing ones per each category; full results are in \cref{tab:naive_results_full}), and evaluate on an unseen target domain. The results are in \Cref{tab:naive_results_cropped}. We observe: (1)~The na{\"i}ve extensions improve only slightly over the Finetune baseline by $\sim$3 pp, and (2) are considerably outperformed by the baselines SI and ER-ACE. These findings highlight the need for more tailored methods that learn domain-invariant representations in CL. $\Rightarrow$ \textbf{Takeaway}: \emph{Na{\"i}ve extensions for domain-invariant learning in CL offer only minimal benefits over standard sequential training.}

\begin{figure}[t]
    \centering
    \includegraphics[width=0.8\linewidth]{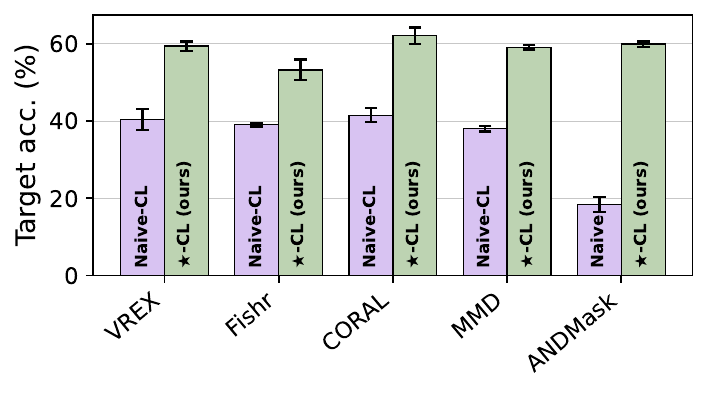}
    \vspace{-5pt}
    \caption{
    {\setlength{\fboxsep}{0pt}
    \textbf{Improvement from $\bigstar$-CL over Na\"ive-CL.} \colorbox{color_tailored}{Our proposed methods} outperform the \colorbox{color_naive}{na\"ive extensions} across all underlying methodologies for computing domain-invariant representations. Same datasets as in \cref{tab:naive_results_cropped}.
    }
    }
    \vspace{-8pt}
    \label{fig:naive_vs_tailored_barplot}
\end{figure}

\tailoredcircled{3} \textcolor{color_tailored_border}{\textbf{Our tailored methods outperform the na\"ive extensions.}} We now evaluate our proposed set of methods against the na{\"i}ve extensions. For each pair of methods (e.g., Na{\"i}ve-CL-VREX versus $\bigstar$-CL-VREX), we compute the performance improvement from our methods. The results are in \Cref{fig:naive_vs_tailored_barplot,tab:naive_vs_tailored_full}. We observe the following patterns: \textbf{(1)} Our proposed methods consistently improve over the na\"ive extensions. \textbf{(2)} Across all datasets, the improvement is $\sim10$pp or higher. \textbf{(3)} The performance gains are strongest for RotatedMNIST and real-world datasets such as Covertype. Our explanation is that, on RotatedMNIST, the na\"ive methods focus on the relative (but spurious) position of the pixels, which breaks in the target domain. Covertype introduces a severe domain shift, as an entirely new wilderness area with distinct vegetation patterns and elevation is used as the target domain. Here, our methods can better learn the factors underlying the presence of vegetation and is thus robust wrt. domain shifts. $\Rightarrow$ \textbf{Takeaway}: \emph{{Our proposed methods consistently and considerably improve over the na\"ive extensions.}}

\textbf{Discussion:} We highlight three main strengths of this paper. \textbf{(1)}~We introduce a \emph{deployment-centric} view on CL, where sequentially trained models are evaluated on an unseen target domain to explicitly assess generalization beyond the training distributions. We encourage future work to adopt this evaluation protocol, as generalization after deployment is a critical yet underexplored aspect of continual learning. \textbf{(2)}~We perform a rigorous empirical evaluation, which spans diverse benchmark and real-world datasets and includes extensive hyperparameter tuning, amounting to more than 10'000 individual runs (with full carbon offsetting). \textbf{(3)}~We derive a new class of methods for continually learning domain-invariant representations, achieving SOTA performance in generalization to unseen target domains. Overall, our results suggest that $\bigstar$-CL methods successfully preserve domain-invariant structure and help avoid shortcut learning in sequential settings.

\section{Impact statement}
We expect our findings to be of practical relevance for domains where generalization post-deployment is essential, but where source data cannot be freely shared (as in medicine) or fully stored (as in manufacturing), and where models must thus be trained sequentially. In such scenarios, standard continual learning methods may rely on spurious, domain-specific cues that are predictive during training but fail after deployment---thus suffering from shortcut learning \cite{geirhos2020shortcut}. Our results suggest that continually learning domain-invariant representations can mitigate this risk by discouraging reliance on such shortcuts and instead capturing more stable, generalizable structure. This makes our approach broadly applicable to domains such as computer vision, robotics, manufacturing, and medicine. In medical settings in particular, improved robustness to distribution shifts could benefit patients, although we emphasize the need for cautious validation and responsible deployment.

\textbf{Limitations:}
We discuss solutions to two limitations of our framework. First, we utilize replay buffers, a common technique in continual learning \citep[e.g.,][]{aljundi2019gradient,buzzega2020dark,sarfraz025semantic}, for storing domain-specific data. In the domain-incremental source setting that we studied, the domain labels are always considered to be available to maintain the domain partitioning. These labels might not be available under all scenarios. For these settings, we suggest using artificially constructed domain boundaries: while \Cref{eq:erm_replay,eq:align_loss,eq:replay_penalty_generic} require domain labels, these labels can also be constructed. 
Second, we use invariance notions that can be computed on a per-domain batch basis. While this covers a vast majority of the DIRL principles used in the literature, \citep[e.g.,][]{sun2016deep,li2018domain,arjovsky2019invariant,parascandolo2020learning,krueger2021out,rame2022fishr,krishnamachari2024uniformly}, some invariance notions might require a different representation. In such cases, \cref{eq:replay_penalty_generic} needs to be adapted. \cref{eq:align_loss} needs to be updated to use domain-wise (and not batch-wise) invariance statistics.

\bibliography{bibliography}

@article{arjovsky2019invariant,
  title={Invariant risk minimization},
  author={Arjovsky, Martin and Bottou, L{\'e}on and Gulrajani, Ishaan and Lopez-Paz, David},
  journal={arXiv preprint arXiv:1907.02893},
  year={2019}
}

@incollection{mccloskey1989catastrophic,
  title={Catastrophic interference in connectionist networks: The sequential learning problem},
  author={McCloskey, Michael and Cohen, Neal J},
  booktitle={Psychology of learning and motivation},
  volume={24},
  pages={109--165},
  year={1989},
  publisher={Elsevier}
}

@article{parascandolo2020learning,
  title={Learning explanations that are hard to vary},
  author={Parascandolo, Giambattista and Neitz, Alexander and Orvieto, Antonio and Gresele, Luigi and Sch{\"o}lkopf, Bernhard},
  journal={arXiv preprint arXiv:2009.00329},
  year={2020}
}

@article{wang2024comprehensive,
  title={A comprehensive survey of continual learning: Theory, method and application},
  author={Wang, Liyuan and Zhang, Xingxing and Su, Hang and Zhu, Jun},
  journal={IEEE Transactions on Pattern Analysis and Machine Intelligence},
  volume={46},
  number={8},
  pages={5362--5383},
  year={2024},
  publisher={IEEE}
}

@inproceedings{he2016deep,
  title={Deep residual learning for image recognition},
  author={He, Kaiming and Zhang, Xiangyu and Ren, Shaoqing and Sun, Jian},
  booktitle={Proceedings of the IEEE Conference on Computer Vision and Pattern recognition},
  pages={770--778},
  year={2016}
}

@article{buzzega2020dark,
  title={Dark experience for general continual learning: a strong, simple baseline},
  author={Buzzega, Pietro and Boschini, Matteo and Porrello, Angelo and Abati, Davide and Calderara, Simone},
  journal={Advances in Neural Information Processing Systems},
  volume={33},
  pages={15920--15930},
  year={2020}
}

@article{liang2023loss,
  title={Loss decoupling for task-agnostic continual learning},
  author={Liang, Yan-Shuo and Li, Wu-Jun},
  journal={Advances in Neural Information Processing Systems},
  volume={36},
  pages={11151--11167},
  year={2023}
}

@article{wang2023contrastive,
  title={Contrastive generative replay method of remaining useful life prediction for rolling bearings},
  author={Wang, Tiancheng and Guo, Di and Sun, Xi-Ming},
  journal={IEEE Sensors Journal},
  volume={23},
  number={19},
  pages={23893--23902},
  year={2023},
  publisher={IEEE}
}

@inproceedings{tang2024incremental,
  title={An incremental unified framework for small defect inspection},
  author={Tang, Jiaqi and Lu, Hao and Xu, Xiaogang and Wu, Ruizheng and Hu, Sixing and Zhang, Tong and Cheng, Tsz Wa and Ge, Ming and Chen, Ying-Cong and Tsung, Fugee},
  booktitle={European Conference on Computer Vision},
  pages={307--324},
  year={2024},
}

@article{kiyasseh2021clinical,
  title={A clinical deep learning framework for continually learning from cardiac signals across diseases, time, modalities, and institutions},
  author={Kiyasseh, Dani and Zhu, Tingting and Clifton, David},
  journal={Nature Communications},
  volume={12},
  number={1},
  pages={4221},
  year={2021},
  publisher={Nature Publishing Group UK London}
}

@article{hinton2015distilling,
  title={Distilling the knowledge in a neural network},
  author={Hinton, Geoffrey and Vinyals, Oriol and Dean, Jeff},
  journal={arXiv preprint arXiv:1503.02531},
  year={2015}
}

@article{caccia2021new,
  title={New insights on reducing abrupt representation change in online continual learning},
  author={Caccia, Lucas and Aljundi, Rahaf and Asadi, Nader and Tuytelaars, Tinne and Pineau, Joelle and Belilovsky, Eugene},
  journal={arXiv preprint arXiv:2104.05025},
  year={2021}
}

@inproceedings{sarfraz025semantic,
  title={Semantic Aware Representation Learning for Lifelong Learning},
  author={Sarfraz, Fahad and Arani, Elahe and Zonooz, Bahram},
  booktitle={The Thirteenth International Conference on Learning Representations (ICLR)},
  year={2025}
}

@inproceedings{sarfraz2023sparse,
  title={Sparse coding in a dual memory system for lifelong learning},
  author={Sarfraz, Fahad and Arani, Elahe and Zonooz, Bahram},
  booktitle={Proceedings of the AAAI Conference on Artificial Intelligence},
  volume={37},
  pages={9714--9722},
  year={2023}
}

@inproceedings{de2021continual,
  title={Continual prototype evolution: Learning online from non-stationary data streams},
  author={De Lange, Matthias and Tuytelaars, Tinne},
  booktitle={{Proceedings of the IEEE/CVF International Conference on Computer Vision}},
  pages={8250--8259},
  year={2021}
}

@inproceedings{toldo2022bring,
  title={Bring evanescent representations to life in lifelong class incremental learning},
  author={Toldo, Marco and Ozay, Mete},
  booktitle={Proceedings of the IEEE/CVF Conference on Computer Vision and Pattern Recognition},
  pages={16732--16741},
  year={2022}
}

@inproceedings{iscen2020memory,
  title={Memory-efficient incremental learning through feature adaptation},
  author={Iscen, Ahmet and Zhang, Jeffrey and Lazebnik, Svetlana and Schmid, Cordelia},
  booktitle={European Conference on Computer Vision},
  pages={699--715},
  year={2020}
}

@inproceedings{churamani2023towards,
  title={Towards causal replay for knowledge rehearsal in continual learning},
  author={Churamani, Nikhil and Cheong, Jiaee and Kalkan, Sinan and Gunes, Hatice},
  booktitle={AAAI Bridge Program on Continual Causality},
  pages={63--70},
  year={2023},
  organization={{PMLR}}
}

@article{chaudhry2019tiny,
  title={On tiny episodic memories in continual learning},
  author={Chaudhry, Arslan and Rohrbach, Marcus and Elhoseiny, Mohamed and Ajanthan, Thalaiyasingam and Dokania, Puneet K and Torr, Philip HS and Ranzato, Marc'Aurelio},
  journal={arXiv preprint arXiv:1902.10486},
  year={2019}
}

@article{magistri2024elastic,
  title={Elastic feature consolidation for cold start exemplar-free incremental learning},
  author={Magistri, Simone and Trinci, Tomaso and Soutif-Cormerais, Albin and van de Weijer, Joost and Bagdanov, Andrew D},
  journal={arXiv preprint arXiv:2402.03917},
  year={2024}
}

@article{yoon2017lifelong,
  title={Lifelong learning with dynamically expandable networks},
  author={Yoon, Jaehong and Yang, Eunho and Lee, Jeongtae and Hwang, Sung Ju},
  journal={arXiv preprint arXiv:1708.01547},
  year={2017}
}

@inproceedings{shankar2021image,
  title={Do image classifiers generalize across time?},
  author={Shankar, Vaishaal and Dave, Achal and Roelofs, Rebecca and Ramanan, Deva and Recht, Benjamin and Schmidt, Ludwig},
  booktitle={{Proceedings of the IEEE/CVF International Conference on Computer Vision}},
  pages={9661--9669},
  year={2021}
}

@article{geirhos2020shortcut,
  title={Shortcut learning in deep neural networks},
  author={Geirhos, Robert and Jacobsen, J{\"o}rn-Henrik and Michaelis, Claudio and Zemel, Richard and Brendel, Wieland and Bethge, Matthias and Wichmann, Felix A},
  journal={Nature Machine Intelligence},
  volume={2},
  number={11},
  pages={665--673},
  year={2020},
  publisher={Nature Publishing Group UK London}
}

@article{hermann2023foundations,
  title={On the foundations of shortcut learning},
  author={Hermann, Katherine L and Mobahi, Hossein and Fel, Thomas and Mozer, Michael C},
  journal={arXiv preprint arXiv:2310.16228},
  year={2023}
}

@article{hua2025domain,
  title={Domain-invariant feature exploration for intelligent fault diagnosis under unseen and time-varying working conditions},
  author={Hua, Zehui and Shi, Juanjuan and Dumond, Patrick},
  journal={Mechanical Systems and Signal Processing},
  volume={224},
  pages={112193},
  year={2025},
  publisher={Elsevier}
}

@article{da2020remaining,
  title={Remaining useful lifetime prediction via deep domain adaptation},
  author={da Costa, Paulo Roberto de Oliveira and Ak{\c{c}}ay, Alp and Zhang, Yingqian and Kaymak, Uzay},
  journal={Reliability Engineering \& System Safety},
  volume={195},
  pages={106682},
  year={2020},
  publisher={Elsevier}
}

@inproceedings{tanwani2021dirl,
  title={DIRL: Domain-invariant representation learning for sim-to-real transfer},
  author={Tanwani, Ajay},
  booktitle={Conference on Robot Learning},
  pages={1558--1571},
  year={2021},
  organization={{PMLR}}
}

@article{
doi:10.1126/science.aaa9375,
author = {Cynthia Dwork  and Vitaly Feldman  and Moritz Hardt  and Toniann Pitassi  and Omer Reingold  and Aaron Roth },
title = {The reusable holdout: Preserving validity in adaptive data analysis},
journal = {Science},
volume = {349},
number = {6248},
pages = {636-638},
year = {2015},
doi = {10.1126/science.aaa9375},
URL = {https://www.science.org/doi/abs/10.1126/science.aaa9375},
eprint = {https://www.science.org/doi/pdf/10.1126/science.aaa9375}}

@inproceedings{chaudry2019efficient,
  title={Efficient Lifelong Learning with A-GEM},
  author={Chaudhry, Arslan and Ranzato, Marc’Aurelio and Rohrbach, Marcus and Elhoseiny, Mohamed},
  booktitle={ICLR},
  year={2019}
}

@article{lopez2017gradient,
  title={Gradient episodic memory for continual learning},
  author={Lopez-Paz, David and Ranzato, Marc'Aurelio},
  journal={Advances in Neural Information Processing Systems},
  volume={30},
  year={2017}
}

@inproceedings{hardt2016train,
  title={Train faster, generalize better: Stability of stochastic gradient descent},
  author={Hardt, Moritz and Recht, Ben and Singer, Yoram},
  booktitle={{International Conference on Machine Learning (ICML)}},
  pages={1225--1234},
  year={2016},
  organization={{PMLR}}
}

@inproceedings{elsayed2024addressing,
    title={Addressing Loss of Plasticity and Catastrophic Forgetting in Continual Learning},
    author={Mohamed Elsayed and A. Rupam Mahmood},
    booktitle={The Twelfth International Conference on Learning Representations (ICLR)},
    year={2024}
}

@article{aljundi2019online,
  title={Online continual learning with maximal interfered retrieval},
  author={Aljundi, Rahaf and Belilovsky, Eugene and Tuytelaars, Tinne and Charlin, Laurent and Caccia, Massimo and Lin, Min and Page-Caccia, Lucas},
  journal={Advances in Neural Information Processing Systems},
  volume={32},
  year={2019}
}

@article{mcinnes2018umap,
  title={Umap: Uniform manifold approximation and projection for dimension reduction},
  author={McInnes, Leland and Healy, John and Melville, James},
  journal={arXiv preprint arXiv:1802.03426},
  year={2018}
}

@article{aljundi2019gradient,
  title={Gradient based sample selection for online continual learning},
  author={Aljundi, Rahaf and Lin, Min and Goujaud, Baptiste and Bengio, Yoshua},
  journal={Advances in Neural Information Processing Systems},
  volume={32},
  year={2019}
}

@article{wu2015wafer,
  author={Wu, Ming-Ju and Jang, Jyh-Shing R. and Chen, Jui-Long},
  journal={IEEE Transactions on Semiconductor Manufacturing}, 
  title={Wafer Map Failure Pattern Recognition and Similarity Ranking for Large-Scale Data Sets}, 
  year={2015},
  volume={28},
  number={1},
  pages={1-12},
  doi={10.1109/TSM.2014.2364237}
}

@dataset{wafermap,
  author = {Jang, Jyh-Shing R.},
  title = {MIR-WM811K: Dataset for wafer map failure pattern recognition},
  year = {2015},
howpublished = "\url{http://mirlab.org/dataset/public/}",
  publisher = {Dryad}}

@article{yoon2019scalable,
  title={Scalable and order-robust continual learning with additive parameter decomposition},
  author={Yoon, Jaehong and Kim, Saehoon and Yang, Eunho and Hwang, Sung Ju},
  journal={arXiv preprint arXiv:1902.09432},
  year={2019}
}

@inproceedings{benjamin2019measuring,
  title     = {Measuring and Regularizing Networks in Function Space},
  author    = {Benjamin, Ari S. and Rolnick, David and K{\"o}rding, Konrad P.},
  booktitle = {International Conference on Learning Representations (ICLR)},
  year      = {2019},
  url       = {https://openreview.net/forum?id=SkMwpiR9Y7}
}

@misc{blackard1998covertype,
  author       = {Blackard, Jock},
  title        = {{Covertype}},
  year         = {1998},
  howpublished = {UCI Machine Learning Repository},
  note         = {{DOI}: https://doi.org/10.24432/C50K5N}
}

@inproceedings{zenke2017continual,
  title={Continual learning through synaptic intelligence},
  author={Zenke, Friedemann and Poole, Ben and Ganguli, Surya},
  booktitle={{International Conference on Machine Learning (ICML)}},
  pages={3987--3995},
  year={2017},
  organization={{PMLR}}
}

@inproceedings{koh2021wilds,
  title={Wilds: A benchmark of in-the-wild distribution shifts},
  author={Koh, Pang Wei and Sagawa, Shiori and Marklund, Henrik and Xie, Sang Michael and Zhang, Marvin and Balsubramani, Akshay and Hu, Weihua and Yasunaga, Michihiro and Phillips, Richard Lanas and Gao, Irena and others},
  booktitle={{{International Conference on Machine Learning (ICML)}}},
  pages={5637--5664},
  year={2021},
  organization={{PMLR}}
}

@article{bandi2018detection,
  title={From detection of individual metastases to classification of lymph node status at the patient level: the {C}amelyon17 challenge},
  author={Bandi, Peter and Geessink, Oscar and Manson, Quirine and Van Dijk, Marcory and Balkenhol, Maschenka and Hermsen, Meyke and Bejnordi, Babak Ehteshami and Lee, Byungjae and Paeng, Kyunghyun and Zhong, Aoxiao and others},
  journal={{T}ransactions on {M}edical {I}maging},
  volume={38},
  number={2},
  pages={550--560},
  year={2019},
  publisher={{IEEE}}
}

@article{lee2020clinical,
  title={Clinical applications of continual learning machine learning},
  author={Lee, Cecilia S and Lee, Aaron Y},
  journal={{The Lancet Digital Health}},
  volume={2},
  number={6},
  pages={e279--e281},
  year={2020},
  publisher={Elsevier}
}

@article{van2022three,
  title={Three types of incremental learning},
  author={Van de Ven, Gido M and Tuytelaars, Tinne and Tolias, Andreas S},
  journal={Nature Machine Intelligence},
  volume={4},
  number={12},
  pages={1185--1197},
  year={2022},
  publisher={Nature Publishing Group UK London}
}

@inproceedings{kang2022forget,
  title={Forget-free continual learning with winning subnetworks},
  author={Kang, Haeyong and Mina, Rusty John Lloyd and Madjid, Sultan Rizky Hikmawan and Yoon, Jaehong and Hasegawa-Johnson, Mark and Hwang, Sung Ju and Yoo, Chang D},
  booktitle={{International Conference on Machine Learning (ICML)}},
  pages={10734--10750},
  year={2022},
  organization={{PMLR}}
}

@inproceedings{mallya2018packnet,
  title={Packnet: Adding multiple tasks to a single network by iterative pruning},
  author={Mallya, Arun and Lazebnik, Svetlana},
  booktitle={Proceedings of the IEEE Conference on Computer Vision and Pattern recognition},
  pages={7765--7773},
  year={2018}
}

@inproceedings{gu2022not,
  title={Not just selection, but exploration: Online class-incremental continual learning via dual view consistency},
  author={Gu, Yanan and Yang, Xu and Wei, Kun and Deng, Cheng},
  booktitle={Proceedings of the IEEE/CVF Conference on Computer Vision and Pattern Recognition},
  pages={7442--7451},
  year={2022}
}

@article{boschini2022class,
  title={Class-incremental continual learning into the extended der-verse},
  author={Boschini, Matteo and Bonicelli, Lorenzo and Buzzega, Pietro and Porrello, Angelo and Calderara, Simone},
  journal={{T}ransactions on {P}attern {A}nalysis and {M}achine {I}ntelligence},
  volume={45},
  number={5},
  pages={5497--5512},
  year={2022},
  publisher={IEEE}
}

@inproceedings{blackard1999comparative,
  author = {Jock A. Blackard and Denis J. Dean},
  booktitle = {Computers and Electronics in Agriculture},
  title = {Comparative accuracies of artificial neural networks and discriminant analysis in predicting forest cover types from cartographic variables},
  year = {1999},
}

@inproceedings{gulrajani2021search,
  title={In Search of Lost Domain Generalization},
  author={Gulrajani, Ishaan and Lopez-Paz, David},
  booktitle={International Conference on Learning Representations (ICLR)},
  year={2021}
}

@article{krishnamachari2024uniformly,
  title={Uniformly distributed feature representations for fair and robust learning},
  author={Krishnamachari, Kiran and Ng, See-Kiong and Foo, Chuan-Sheng},
  journal={Transactions on Machine Learning Research},
  year={2024}
}

@inproceedings{li2018domain,
  title={Domain generalization with adversarial feature learning},
  author={Li, Haoliang and Pan, Sinno Jialin and Wang, Shiqi and Kot, Alex C},
  booktitle={Proceedings of the IEEE Conference on Computer Vision and Pattern recognition},
  pages={5400--5409},
  year={2018}
}

@inproceedings{farias2024self,
  title={Self-Normalized Resets for Plasticity in Continual Learning},
  author={Farias, Vivek and Jozefiak, Adam Daniel},
  booktitle={The Thirteenth International Conference on Learning Representations (ICLR)},
  year={2025}
}

@article{li2016revisiting,
  title={Revisiting batch normalization for practical domain adaptation},
  author={Li, Yanghao and Wang, Naiyan and Shi, Jianping and Liu, Jiaying and Hou, Xiaodi},
  journal={arXiv preprint arXiv:1603.04779},
  year={2016}
}

@inproceedings{sun2020test,
  title={Test-time training with self-supervision for generalization under distribution shifts},
  author={Sun, Yu and Wang, Xiaolong and Liu, Zhuang and Miller, John and Efros, Alexei and Hardt, Moritz},
  booktitle={{International Conference on Machine Learning (ICML)}},
  pages={9229--9248},
  year={2020},
  organization={{PMLR}}
}

@inproceedings{wang2021tent,
  title={Tent: Fully Test-Time Adaptation by Entropy Minimization},
  author={Wang, Dequan and Shelhamer, Evan and Liu, Shaoteng and Olshausen, Bruno and Darrell, Trevor},
  booktitle={International Conference on Learning Representations (ICLR)},
  year={2021},
  url={https://openreview.net/forum?id=uXl3bZLkr3c}
}

@inproceedings{sojka2023ar,
    author    = {S\'ojka, Damian and Cygert, Sebastian and Twardowski, Bart{\l}omiej and Trzci\'nski, Tomasz},
    title     = {AR-TTA: A Simple Method for Real-World Continual Test-Time Adaptation},
    booktitle = {{Proceedings of the IEEE/CVF International Conference on Computer Vision} (ICCV) Workshops},
    month     = {October},
    year      = {2023},
    pages     = {3491-3495}
}

@inproceedings{wang2022continual,
  title={Continual test-time domain adaptation},
  author={Wang, Qin and Fink, Olga and Van Gool, Luc and Dai, Dengxin},
  booktitle={Proceedings of the IEEE/CVF Conference on Computer Vision and Pattern Recognition},
  pages={7201--7211},
  year={2022}
}

@article{le2015tiny,
  title={Tiny imagenet visual recognition challenge},
  author={Le, Yann and Yang, Xuan},
  journal={CS 231N},
  volume={7},
  number={7},
  pages={3},
  year={2015}
}

@inproceedings{ghifary2015domain,
  title={Domain generalization for object recognition with multi-task autoencoders},
  author={Ghifary, Muhammad and Kleijn, W Bastiaan and Zhang, Mengjie and Balduzzi, David},
  booktitle={Proceedings of the IEEE International Conference on Computer Vision},
  pages={2551--2559},
  year={2015}
}

@TECHREPORT{Krizhevsky09learningmultiple,
    author = {Alex Krizhevsky},
    title = {Learning multiple layers of features from tiny images},
    institution = {},
    year = {2009}
}

@article{hendrycks2019benchmarking,
  title={Benchmarking neural network robustness to common corruptions and perturbations},
  author={Hendrycks, Dan and Dietterich, Thomas},
  journal={arXiv preprint arXiv:1903.12261},
  year={2019}
}

@inproceedings{eskandar2025star,
 author = {Eskandar, Masih and Imtiaz, Tooba and Hill, Davin and Wang, Zifeng and Dy, Jennifer},
 booktitle = {International Conference on Learning Representations (ICLR)},
 editor = {Y. Yue and A. Garg and N. Peng and F. Sha and R. Yu},
 pages = {49102--49117},
 title = {{STAR}: Stability-Inducing Weight Perturbation for Continual Learning},
 volume = {2025},
 year = {2025}
}

@article{shi2023unified,
  title={A unified approach to domain incremental learning with memory: Theory and algorithm},
  author={Shi, Haizhou and Wang, Hao},
  journal={Advances in Neural Information Processing Systems},
  volume={36},
  pages={15027--15059},
  year={2023}
}

@article{lecun1998mnist,
  title={The MNIST database of handwritten digits},
  author={LeCun, Yann},
  journal={http://yann. lecun. com/exdb/mnist/},
  year={1998}
}

@article{gretton2012kernel,
  title={A kernel two-sample test},
  author={Gretton, Arthur and Borgwardt, Karsten M and Rasch, Malte J and Sch{\"o}lkopf, Bernhard and Smola, Alexander},
  journal={The Journal of Machine Learning Research},
  volume={13},
  number={1},
  pages={723--773},
  year={2012},
  publisher={JMLR. org}
}

@article{lu2025rethinking,
  title={Rethinking the stability-plasticity trade-off in continual learning from an architectural perspective},
  author={Lu, Aojun and Yuan, Hangjie and Feng, Tao and Sun, Yanan},
  journal={arXiv preprint arXiv:2506.03951},
  year={2025}
}

@article{rusu2016progressive,
  title={Progressive neural networks},
  author={Rusu, Andrei A and Rabinowitz, Neil C and Desjardins, Guillaume and Soyer, Hubert and Kirkpatrick, James and Kavukcuoglu, Koray and Pascanu, Razvan and Hadsell, Raia},
  journal={arXiv preprint arXiv:1606.04671},
  year={2016}
}

@article{kirkpatrick2017overcoming,
  title={Overcoming catastrophic forgetting in neural networks},
  author={Kirkpatrick, James and Pascanu, Razvan and Rabinowitz, Neil and Veness, Joel and Desjardins, Guillaume and Rusu, Andrei A and Milan, Kieran and Quan, John and Ramalho, Tiago and Grabska-Barwinska, Agnieszka and others},
  journal={Proceedings of the National Academy of Sciences},
  volume={114},
  number={13},
  pages={3521--3526},
  year={2017},
  publisher={National Acad Sciences}
}

@article{sugiyama2007direct,
  title={Direct importance estimation with model selection and its application to covariate shift adaptation},
  author={Sugiyama, Masashi and Nakajima, Shinichi and Kashima, Hisashi and Buenau, Paul and Kawanabe, Motoaki},
  journal={Advances in Neural Information Processing Systems},
  volume={20},
  year={2007}
}

@article{wang2018low,
  title={Low-rank transfer human motion segmentation},
  author={Wang, Lichen and Ding, Zhengming and Fu, Yun},
  journal={{Transactions on Image Processing}},
  volume={28},
  number={2},
  pages={1023--1034},
  year={2019},
  publisher={IEEE}
}

@inproceedings{ganin2015unsupervised,
  title={Unsupervised domain adaptation by backpropagation},
  author={Ganin, Yaroslav and Lempitsky, Victor},
  booktitle={{International Conference on Machine Learning (ICML)}},
  pages={1180--1189},
  year={2015},
  organization={{PMLR}}
}

@article{qin2019pointdan,
  title={Pointdan: A multi-scale 3d domain adaption network for point cloud representation},
  author={Qin, Can and You, Haoxuan and Wang, Lichen and Kuo, C-C Jay and Fu, Yun},
  journal={Advances in Neural Information Processing Systems},
  volume={32},
  year={2019}
}

@inproceedings{zhang2013domain,
  title={Domain adaptation under target and conditional shift},
  author={Zhang, Kun and Sch{\"o}lkopf, Bernhard and Muandet, Krikamol and Wang, Zhikun},
  booktitle={{International Conference on Machine Learning (ICML)}},
  pages={819--827},
  year={2013},
  organization={{PMLR}}
}

@article{courty2017joint,
  title={Joint distribution optimal transportation for domain adaptation},
  author={Courty, Nicolas and Flamary, R{\'e}mi and Habrard, Amaury and Rakotomamonjy, Alain},
  journal={Advances in Neural Information Processing Systems},
  volume={30},
  year={2017}
}

@inproceedings{mahajan2021domain,
  title={Domain generalization using causal matching},
  author={Mahajan, Divyat and Tople, Shruti and Sharma, Amit},
  booktitle={{International Conference on Machine Learning (ICML)}},
  pages={7313--7324},
  year={2021},
  organization={{PMLR}}
}

@article{rojas2018invariant,
  title={Invariant models for causal transfer learning},
  author={Rojas-Carulla, Mateo and Sch{\"o}lkopf, Bernhard and Turner, Richard and Peters, Jonas},
  journal={Journal of Machine Learning Research},
  volume={19},
  number={36},
  pages={1--34},
  year={2018}
}

@article{hwang2024sf,
  title={SF(DA)2: Source-free domain adaptation through the lens of data augmentation},
  author={Hwang, Uiwon and Lee, Jonghyun and Shin, Juhyeon and Yoon, Sungroh},
  journal={arXiv preprint arXiv:2403.10834},
  year={2024}
}

@inproceedings{liang2020we, 
 title={Do We Really Need to Access the Source Data? Source Hypothesis Transfer for Unsupervised Domain Adaptation}, 
 author={Liang, Jian and Hu, Dapeng and Feng, Jiashi}, 
 booktitle={{International Conference on Machine Learning (ICML)} (ICML)},  
 pages={6028--6039},
 year={2020}
}

@article{liang2021source,  
 title={Source Data-absent Unsupervised Domain Adaptation through Hypothesis Transfer and Labeling Transfer}, 
 author={Liang, Jian and Hu, Dapeng and Wang, Yunbo and He, Ran and Feng, Jiashi},   
 journal={{Transactions on Pattern Analysis and Machine Intelligence (TPAMI)}},
 year={2021}, 
}

@article{feng2023cosda,
  title={Cosda: Continual source-free domain adaptation},
  author={Feng, Haozhe and Yang, Zhaorui and Chen, Hesun and Pang, Tianyu and Du, Chao and Zhu, Minfeng and Chen, Wei and Yan, Shuicheng},
  journal={arXiv preprint arXiv:2304.06627},
  year={2023}
}

@article{saha2021gradient,
  title={Gradient projection memory for continual learning},
  author={Saha, Gobinda and Garg, Isha and Roy, Kaushik},
  journal={arXiv preprint arXiv:2103.09762},
  year={2021}
}

@article{ratcliff1990connectionist,
  title={Connectionist models of recognition memory: constraints imposed by learning and forgetting functions.},
  author={Ratcliff, Roger},
  journal={Psychological Review},
  volume={97},
  number={2},
  pages={285},
  year={1990},
  publisher={American Psychological Association}
}

@inproceedings{serra2018overcoming,
  title={Overcoming catastrophic forgetting with hard attention to the task},
  author={Serra, Joan and Suris, Didac and Miron, Marius and Karatzoglou, Alexandros},
  booktitle={{International Conference on Machine Learning (ICML)}},
  pages={4548--4557},
  year={2018},
  organization={{PMLR}}
}

@article{hurtado2023continual,
  title={Continual learning for predictive maintenance: Overview and challenges},
  author={Hurtado, Julio and Salvati, Dario and Semola, Rudy and Bosio, Mattia and Lomonaco, Vincenzo},
  journal={Intelligent Systems with Applications},
  volume={19},
  pages={200251},
  year={2023},
  publisher={Elsevier}
}

@article{bruno2025continual,
  title={Continual learning in medicine: A systematic literature review},
  author={Bruno, Pierangela and Quarta, Alessandro and Calimeri, Francesco},
  journal={Neural Processing Letters},
  volume={57},
  number={1},
  pages={2},
  year={2025},
  publisher={Springer}
}

@article{vokinger2021continual,
  title={Continual learning in medical devices: FDA's action plan and beyond},
  author={Vokinger, Kerstin N and Feuerriegel, Stefan and Kesselheim, Aaron S},
  journal={The Lancet Digital Health},
  volume={3},
  number={6},
  pages={e337--e338},
  year={2021},
  publisher={Elsevier}
}

@article{parisi2019continual,
  title={Continual lifelong learning with neural networks: A review},
  author={Parisi, German I and Kemker, Ronald and Part, Jose L and Kanan, Christopher and Wermter, Stefan},
  journal={Neural networks},
  volume={113},
  pages={54--71},
  year={2019},
  publisher={Elsevier}
}

@article{thrun1995lifelong,
  title={Lifelong robot learning},
  author={Thrun, Sebastian and Mitchell, Tom M},
  journal={Robotics and autonomous systems},
  volume={15},
  number={1-2},
  pages={25--46},
  year={1995},
  publisher={Elsevier}
}

@article{shaheen2022continual,
  title={Continual learning for real-world autonomous systems: Algorithms, challenges and frameworks},
  author={Shaheen, Khadija and Hanif, Muhammad Abdullah and Hasan, Osman and Shafique, Muhammad},
  journal={Journal of Intelligent \& Robotic Systems},
  volume={105},
  number={1},
  pages={9},
  year={2022},
  publisher={Springer}
}

@article{welford1962note,
  title={Note on a method for calculating corrected sums of squares and products},
  author={Welford, Barry Payne},
  journal={Technometrics},
  volume={4},
  number={3},
  pages={419--420},
  year={1962},
  publisher={Taylor \& Francis}
}

@article{shi2021gradient,
  title={Gradient matching for domain generalization},
  author={Shi, Yuge and Seely, Jeffrey and Torr, Philip HS and Hannun, Awni and Usunier, Nicolas and Synnaeve, Gabriel},
  journal={arXiv preprint arXiv:2104.09937},
  year={2021}
}

@inproceedings{sun2016deep,
  title={Deep coral: Correlation alignment for deep domain adaptation},
  author={Sun, Baochen and Saenko, Kate},
  booktitle={European Conference on Computer Vision},
  pages={443--450},
  year={2016},
  organization={Springer}
}

@article{zhou2022domain,
  title={Domain generalization: A survey},
  author={Zhou, Kaiyang and Liu, Ziwei and Qiao, Yu and Xiang, Tao and Loy, Chen Change},
  journal={IEEE Transactions on Pattern Analysis and Machine Intelligence},
  volume={45},
  number={4},
  pages={4396--4415},
  year={2022},
  publisher={IEEE}
}

@inproceedings{rame2022fishr,
  title={Fishr: Invariant gradient variances for out-of-distribution generalization},
  author={Rame, Alexandre and Dancette, Corentin and Cord, Matthieu},
  booktitle={{International Conference on Machine Learning (ICML)}},
  pages={18347--18377},
  year={2022},
  organization={{PMLR}}
}

@inproceedings{krueger2021out,
  title={Out-of-distribution generalization via risk extrapolation (rex)},
  author={Krueger, David and Caballero, Ethan and Jacobsen, Joern-Henrik and Zhang, Amy and Binas, Jonathan and Zhang, Dinghuai and Le Priol, Remi and Courville, Aaron},
  booktitle={{International Conference on Machine Learning (ICML)}},
  pages={5815--5826},
  year={2021},
  organization={{PMLR}}
}
\bibliographystyle{conference}

\newpage
\appendix
\onecolumn

\section{Extended Related Works}\label{sec:extended_related_work}

\textbf{Continual learning} (CL) focuses on sequentially training on a stream of data distributions while preventing forgetting \citep{mccloskey1989catastrophic,lu2025rethinking}. In the literature, \textbf{three scenarios of CL} exist \citep{van2022three}: \textbf{(1)}~Task-incremental learning, where each domain introduces a new task that needs to be learned additionally to the previous tasks. Due to the usage of task-identities during inference, which are not available for \emph{novel domains}, this setting is essentially solved \citep{mallya2018packnet,kang2022forget}. \textbf{(2)}~Class-incremental learning, where each domain introduces new classes that need to be learned in addition to (now unavailable) existing classes \citep{boschini2022class,gu2022not}. \textbf{(3)}~Domain-incremental learning, where the number of classes is fixed and shared across all domains, but each domain contains data from a different distribution \citep{shi2023unified,wang2024comprehensive}. Our \emph{training} is similar to domain-incremental learning, but then, we evaluate on an unseen target domain.

CL methods are commonly grouped into four major categories:
\textbf{(1)}~Optimization-based methods, which modify gradients to prevent interference with previously learned tasks \citep{saha2021gradient} or prevent weight updates \citep{elsayed2024addressing}.
\textbf{(2)}~Regularization based methods, which reduce forgetting by regularizing updates to model weights or features \citep{kirkpatrick2017overcoming,zenke2017continual,magistri2024elastic}. \textbf{(3)}~Architecture-based methods, which add new network modules for new domains (and hence often scale quadratically or even exponentially with the number of domains) \citep{rusu2016progressive,yoon2017lifelong,yoon2019scalable}. \textbf{(4)}~Replay-based methods, which maintain a buffer to replay (i) samples \citep{chaudhry2019tiny,churamani2023towards,eskandar2025star}, (ii) features \citep{iscen2020memory,toldo2022bring}, or (iii) latent class prototypes \citep{de2021continual,sarfraz2023sparse,sarfraz025semantic}.

\textbf{Domain invariant representation learning} (DIRL) methods aim to learn features that are invariant to the domain and thus generalize better to novel domains \citep{li2018domain,krishnamachari2024uniformly}. Such methods are often motivated by the idea that than e underlying causal mechanism exists across all domains and that causal mechanism which can be uncovered through joint training \citep{arjovsky2019invariant,krueger2021out,mahajan2021domain,rame2022fishr}. DIRL methods can be grouped based on the invariance property they target during training: \textbf{(1)~Features}, where features are aligned across different feature distributions (ie., domains) \citep{sun2016deep,li2018domain,krishnamachari2024uniformly}. \textbf{(2)~Gradients}, where gradient-statistics are aligned across domains \citep{parascandolo2020learning,shi2021gradient,rame2022fishr}. \textbf{(3)~Weights}, where a fixed set of weights, such as a classification head, is aligned across domain-specific representations \citep{rojas2018invariant,arjovsky2019invariant}. These methods are predominantly used in a multi-source setting, where the training data is divided into a set of multiple distinct source domains \citep{gulrajani2021search,zhou2022domain}. After \emph{jointly} training on the set of source domains, the model is evaluated on a unseen target domain. Our work is orthogonal to existing DIRL works in that we focus on sequentially learning invariant representation across a \emph{sequence} of source domains, with the goal of improved generalization to a new target domain.

\textbf{Domain adaptation} (DA) methods adapt a model from a source to a target domain and require access to data of both domains \citep{hwang2024sf}. Commonly, DA is studied under the \textbf{unsupervised domain adaptation (UDA)} setting, where access to unlabeled target data is available \citep{ganin2015unsupervised,qin2019pointdan}. To bridge the differences between source and target distributions, UDA methods try to match the (i)  marginal \citep{sugiyama2007direct,wang2018low} or (ii) conditional distributions \citep{zhang2013domain,courty2017joint}. Some of the methods also lend themselves naturally to \textbf{Continual domain adaptation}, where a pre-trained model is continually adapted to a sequence of changing target distributions \citep{liang2020we,liang2021source,feng2023cosda}. Both DA and CDA differ from the setting studied in this paper. In our deployment-oriented evaluation protocol, we perform training on a \textbf{supervised} sequence of source domains, and afterwards test on a single unseen target domain. Nonetheless, we later also compare against widely-used DA baseline SHOT++ \citep{liang2021source}.

\textbf{Continual test-time adaptation} (CTTA) studies on-the-fly model adaptations at test time to a non-stationary unlabeled (target) stream  \citep{wang2022continual}. For adaptation, methods use test-time training \citep{sun2020test}, entropy-minimization \citep{wang2021tent}, or recomputing normalization statistics \citep{li2016revisiting}. Representative methods include extending test-time augmentations with stochastic weight resetting and EMA teacher-student frameworks \citep{wang2022continual}, self-training with a small replay-buffer and dynamic normalization \citep{sojka2023ar}. CTTA is different from our setting. We have a sequence of \textbf{source} domains and evaluate on a single, fixed target domain. Nonetheless, we later include two widely-used CTTA baselines, namely, TENT \citep{wang2021tent} and CoTTA \citep{wang2022continual}.

\newpage
\section{Details for the na\"ive domain-invariant CL algorithms}\label{sec:naive_details}

We instantiate Eq.~(\ref{eq:na\"ive_template}) by specifying a method-dependent statistic
\(\widehat{\phi}_s(\theta,\omega;B_s)\) computed on a minibatch \(B_s \sim \sourceD\) from the current domain,
and a domain prior \(\Phi_{s'}\) computed at the end of training on each past domain \(s' < s\).
All priors are computed via Welford aggregation (Eq.~\ref{eq:prior_def}) and are fixed once stored.
During training on domain \(s\), the na\"ive penalty matches the current statistic to past priors
(Eq.~\ref{eq:prior_matching}).

\paragraph{NaiveVREX.}
VREx \citep{krueger2021out} encourages invariance by matching risks across environments. In the na\"ive sequential variant, we match the current-domain empirical risk to stored domain-level risk summaries.

\begin{itemize}
    \item \textbf{Per-batch statistic (current domain):}
    \begin{equation}
        \widehat{\phi}_s \equiv \widehat{r}_s
        \;=\;
        \mathbb{E}_{(\mathbf{x},y)\sim B_s}\!\left[\mathcal{L}(h(\mathbf{x};\theta,\omega),y)\right].
        \label{eq:naive_vrex_stat}
    \end{equation}
    \item \textbf{Stored prior (past domains):}
    \begin{equation}
        \Phi_{s'} \equiv \bar{r}_{s'}
        \;=\;
        \textsc{Aggregate}_{(\mathbf{x},y)\sim \sourceDPrime}
        \left(\mathcal{L}(h(\mathbf{x};\theta,\omega),y)\right),
        \qquad s'<s.
        \label{eq:naive_vrex_prior}
    \end{equation}
    \item \textbf{Na\"ive penalty:}
    \begin{equation}
        \mathcal{P}_s^{\textsc{NaiveVREX}}
        \;=\;
        \frac{1}{s-1}\sum_{s'=1}^{s-1}
        \left(\widehat{r}_s - \bar{r}_{s'}\right)^2.
        \label{eq:naive_vrex_penalty}
    \end{equation}
\end{itemize}

\paragraph{Na\"iveFishr.}
Fishr \citep{rame2022fishr} matches gradient-variance statistics across domains to promote invariance. In the na\"ive sequential variant, we compute a gradient-variance vector on the current minibatch and match it to stored per-domain prototypes.

Let \(\nabla_{\omega}\mathcal{L}(h(\mathbf{x};\theta,\omega),y)\) denote the gradient of the loss w.r.t.\ classifier
parameters \(\omega\).\footnote{ (1) As in Fishr, the statistic is typically computed layer-wise and concatenated; for brevity we write a single vector. (2) In our work, for computational reasons, we only align gradients of $\omega$. (\citet{rame2022fishr} also found that the difference between aligning the full weights versus just $\omega$ is minor.)}

\begin{itemize}
    \item \textbf{Per-batch statistic (current domain):}
    \begin{equation}
        \widehat{\phi}_s \equiv \widehat{\mathbf{v}}_s
        \;=\;
        \mathrm{Var}_{(\mathbf{x},y)\sim B_s}
        \!\left(
        \nabla_{\omega}\mathcal{L}(h(\mathbf{x};\theta,\omega),y)
        \right).
        \label{eq:naive_fishr_stat}
    \end{equation}
    \item \textbf{Stored prior (past domains):}
    \begin{equation}
        \Phi_{s'} \equiv \bar{\mathbf{v}}_{s'}
        \;=\;
        \textsc{Aggregate}_{(\mathbf{x},y)\sim \sourceDPrime}
        \left(
        \mathrm{Var}\!\left(\nabla_{\omega}\mathcal{L}(h(\mathbf{x};\theta,\omega),y)\right)
        \right),
        \qquad s'<s.
        \label{eq:naive_fishr_prior}
    \end{equation}
    \item \textbf{Na\"ive penalty:}
    \begin{equation}
        \mathcal{P}_s^{\textsc{Na\"ive-Fishr}}
        \;=\;
        \frac{1}{s-1}\sum_{s'=1}^{s-1}
        \left\|\widehat{\mathbf{v}}_s - \bar{\mathbf{v}}_{s'}\right\|_2^2.
        \label{eq:naive_fishr_penalty}
    \end{equation}
\end{itemize}

\paragraph{Na\"ive-CORAL.}
Deep CORAL \citep{sun2016deep} aligns second-order feature statistics across domains.
In the na\"ive sequential variant, we compute the current feature mean/covariance on a minibatch and match to stored domain-level feature moments.

Let \(\mathbf{z}=f_\theta(\mathbf{x}) \in \mathcal{H}\) be the latent representation of $\mathbf{x}$.

\begin{itemize}
    \item \textbf{Per-batch statistic (current domain):}
    \begin{align}
        \widehat{\phi}_s &\equiv (\widehat{\boldsymbol{\mu}}_s,\widehat{\boldsymbol{\Sigma}}_s), \\
        \widehat{\boldsymbol{\mu}}_s &= \mathbb{E}_{\mathbf{x}\sim B_s}\!\left[f_\theta(\mathbf{x})\right], \\
        \widehat{\boldsymbol{\Sigma}}_s &= \mathrm{Cov}_{\mathbf{x}\sim B_s}\!\left(f_\theta(\mathbf{x})\right).
        \label{eq:naive_coral_stat}
    \end{align}
    \item \textbf{Stored prior (past domains):}
    \begin{equation}
        \Phi_{s'} \equiv (\bar{\boldsymbol{\mu}}_{s'},\bar{\boldsymbol{\Sigma}}_{s'}),
        \qquad
        (\bar{\boldsymbol{\mu}}_{s'},\bar{\boldsymbol{\Sigma}}_{s'})
        =
        \textsc{Aggregate}_{\mathbf{x}\sim \sourceDPrime}
        \left(
        f_\theta(\mathbf{x}),\;
        \mathrm{Cov}(f_\theta(\mathbf{x}))
        \right).
        \label{eq:naive_coral_prior}
    \end{equation}
    \item \textbf{Na\"ive penalty:}
    \begin{equation}
        \mathcal{P}_s^{\textsc{Na\"ive-CORAL}}
        \;=\;
        \frac{1}{s-1}\sum_{s'=1}^{s-1}
        \left(
        \left\|\widehat{\boldsymbol{\mu}}_s-\bar{\boldsymbol{\mu}}_{s'}\right\|_2^2
        +
        \left\|\widehat{\boldsymbol{\Sigma}}_s-\bar{\boldsymbol{\Sigma}}_{s'}\right\|_F^2
        \right).
        \label{eq:naive_coral_penalty}
    \end{equation}
\end{itemize}

\paragraph{Na\"ive-MMD.}
MMD-based alignment \citep{li2018domain} measures discrepancies between feature distributions via a kernel. To avoid quadratic-time estimators, we follow common practice and use random Fourier features (RFF) to obtain an explicit feature map \(z(\cdot)\) approximating an RBF kernel. We then match the (approximate) kernel mean embedding of the current domain to stored embeddings.

\begin{itemize}
    \item \textbf{Per-batch statistic (current domain):}
    \begin{equation}
        \widehat{\phi}_s \equiv \widehat{\boldsymbol{\mu}}^z_s
        \;=\;
        \mathbb{E}_{\mathbf{x}\sim B_s}\!\left[z\!\big(f_\theta(\mathbf{x})\big)\right].
        \label{eq:naive_mmd_stat}
    \end{equation}
    \item \textbf{Stored prior (past domains):}
    \begin{equation}
        \Phi_{s'} \equiv \bar{\boldsymbol{\mu}}^z_{s'}
        \;=\;
        \textsc{Aggregate}_{\mathbf{x}\sim \sourceDPrime}
        \left(
        z\!\big(f_\theta(\mathbf{x})\big)
        \right),
        \qquad s'<s.
        \label{eq:naive_mmd_prior}
    \end{equation}
    \item \textbf{Na\"ive penalty:}
    \begin{equation}
        \mathcal{P}_s^{\textsc{Na\"ive-MMD}}
        \;=\;
        \frac{1}{s-1}\sum_{s'=1}^{s-1}
        \left\|\widehat{\boldsymbol{\mu}}^z_s-\bar{\boldsymbol{\mu}}^z_{s'}\right\|_2^2.
        \label{eq:naive_mmd_penalty}
    \end{equation}
\end{itemize}
\paragraph{Na\"ive-ANDMask (cross-domain gradient sign agreement via stored mean gradients).} ANDMask \citep{parascandolo2020learning} updates parameters only where gradients agree in sign across sources. In our na\"ive sequential variant, we approximate multi-domain sign agreement by storing, for each past domain \(s' < s\), a \emph{domain-level mean gradient} (estimated online after training on \(D_{s'}\)) and combining the stored gradients with the current minibatch gradient.

Let \(B_s \sim \sourceD\) be the current minibatch, and define the current-domain minibatch gradient
\begin{equation}
    g_s \;=\; \nabla_{\theta,\omega}\,\mathcal{L}_{\mathrm{ERM}}(B_s;\theta,\omega).
    \label{eq:naive_andmask_gs}
\end{equation}

\begin{itemize}
    \item \textbf{Stored prior (past domains):} for each past domain \(s' < s\) and each parameter tensor index \(j\),
    we maintain a Welford estimate of the \emph{mean gradient} over minibatches from \(D_{s'}\):
    \begin{equation}
        \Phi_{s'}^{(j)} \;\equiv\; \bar{g}_{s'}^{(j)}
        \;=\;
        \textsc{Aggregate}_{B\sim \sourceDPrime}\!\left(
        \nabla_{\theta,\omega}^{(j)} \mathcal{L}_{\mathrm{ERM}}(B;\theta,\omega)
        \right).
        \label{eq:naive_andmask_prior}
    \end{equation}

    \item \textbf{Per-batch statistic (current domain):} the current minibatch gradient tensor \(g_s^{(j)}\)
    (Eq.~\ref{eq:naive_andmask_gs}) for each parameter index \(j\).

    \item \textbf{Mask construction (hard agreement):} for each parameter index \(j\), form the set of gradients
    \(\{\bar{g}_{s'}^{(j)}\}_{s'<s}\cup\{g_s^{(j)}\}\), compute per-entry sign agreement, and threshold by \(\tau\in[0,1]\):
    \begin{equation}
        \mathbf{m}_s^{(j)}
        \;=\;
        \mathbb{I}\!\left(
        \left|
        \frac{1}{s}\left(
        \mathrm{sign}\!\left(g_s^{(j)}\right)
        + \sum_{s'=1}^{s-1}\mathrm{sign}\!\left(\bar{g}_{s'}^{(j)}\right)
        \right)
        \right|
        \ge \tau
        \right).
        \label{eq:naive_andmask_mask}
    \end{equation}

    \item \textbf{Masked update rule:} let
    \(\tilde{g}_s^{(j)}=\frac{1}{s}\left(g_s^{(j)}+\sum_{s'=1}^{s-1}\bar{g}_{s'}^{(j)}\right)\) be the elementwise average
    gradient across current and stored past domains. The applied gradient is
    \begin{equation}
        \nabla_{\theta,\omega}^{(j)} \;\leftarrow\;
        \frac{\mathbf{m}_s^{(j)}\odot \tilde{g}_s^{(j)}}{\frac{1}{|\theta,\omega|^{(j)}}\sum \mathbf{m}_s^{(j)} + \varepsilon},
        \qquad \varepsilon>0,
        \label{eq:naive_andmask_update}
    \end{equation}
    where the denominator matches the implementation’s normalization by the mask density (to stabilize update magnitude).
\end{itemize}

Unlike the other na\"ive methods, Na\"iveANDMask is realized as a masked gradient update rather than an explicit additive penalty \(\mathcal{P}_s\). Conceptually, the procedure enforces invariance by \emph{restricting updates} to coordinates that exhibit consistent gradient signs across domains, approximated here using stored mean gradients.

\newpage
\section{Details for our CL methods for domain-invariant representations}\label{sec:ours_details}
The na\"ive template in \cref{eq:na\"ive_template} enforces invariance through \emph{static} priors $\{\priorPrime\}_{e<s}$. It can thus not faithfully emulate multi-domain objectives which require \emph{simultaneous} access to multiple domains. To address this, we propose a set of methods which (i) reintroduce multi-domain computation through a memory buffer and (ii) stabilize invariance learning through sequential invariance alignment. The main details are in \Cref{sec:our_methods}; we here provide more details about the individual methods.

\textbf{CL-VREX.}
The underlying VREX aims to match risks across domains by penalizing the variance of domain-specific risks \citep{krueger2021out}. Using the replay buffer, we compute the per-domain minibatch risks $\widehat{r}_{s^{\prime}}(\sourceBPrime)$ for all ${s^{\prime}}\le s$ and define:
\begin{equation}
    \mathcal{P}_s^{\mathrm{replay}}
    \;=\;
    \mathrm{Var}\Big(\big\{\widehat{r}_{s^{\prime}}(\sourceBPrime)\big\}_{e\le s}\Big) .
\end{equation}

\textbf{CL-Fishr.}
Fishr matches gradient-variance statistics across domains \citep{rame2022fishr}. For our sequential setting, we compute per-domain gradient-variance summaries on classifier parameters, denoted by $\widehat{\mathbf{v}}_{s^{\prime}}(\sourceBPrime)$, and enforce invariance through variance matching:
\begin{equation}
    \mathcal{P}_s^{\mathrm{replay}}
    \;=\;
    \frac{1}{s}\sum_{{s^{\prime}}\le s}
    \left\|
    \widehat{\mathbf{v}}_{s^{\prime}}(\sourceBPrime)
    -
    \frac{1}{s}\sum_{j\le s}\widehat{\mathbf{v}}_j(B_j)
    \right\|_2^2 .
\end{equation}
Further, we align replay-domain gradient-variance statistics to stored references $\priorPrime$ using a scale-invariant representation (log-variance) with
\begin{equation}
    \priorPrime = \mathrm{Normalize}\big(\log(\mathbf{v}_{s^{\prime}} + \varepsilon)\big),
    \qquad
    \mathcal{L}_{\mathrm{align}}
    \;=\;
    \frac{1}{s-1}\sum_{e<s}
    \left(
    1 -
    \cos\!\Big(
    \mathrm{Normalize}\big(\log(\widehat{\mathbf{v}}_{s^{\prime}}(\sourceBPrime)+\varepsilon)\big),
    \priorPrime
    \Big)
    \right) .
\end{equation}
where the cosine alignment emphasizes \emph{directional} agreement of invariance statistics while being robust to scale changes over training.

\textbf{CL-CORAL.}
The CORAL methods matches second-order moments of feature, obtained via the feature extractor $f_\theta$,
across domains \citep{sun2016deep}. Using the replay buffer, we compute per-domain feature mean and covariance on minibatches:
\begin{equation}
    \widehat{\mu}_{s^{\prime}}(\sourceBPrime)=\mathbb{E}_{\mathbf{x}\sim \sourceBPrime}[f_\theta(\mathbf{x})],
    \qquad
    \widehat{\Sigma}_{s^{\prime}}(\sourceBPrime)=\mathrm{Cov}_{\mathbf{x}\sim \sourceBPrime}[f_\theta(\mathbf{x})] .
\end{equation}
We then enforce invariance by matching these moments across the set of domains in the current training step:
\begin{equation}
    \mathcal{P}_s^{\mathrm{replay}}
    \;=\;
    \frac{1}{s}\sum_{{s^{\prime}}\le s}
    \Big(
    \big\|\widehat{\mu}_{s^{\prime}}(\sourceBPrime)-\bar{\mu}\big\|_2^2
    +
    \big\|\widehat{\Sigma}_{s^{\prime}}(\sourceBPrime)-\bar{\Sigma}\big\|_F^2
    \Big) .
\end{equation}
where $(\bar{\mu},\bar{\Sigma})$ denote averages across $\{e\le s\}$ For CL-CORAL, the sequential invariance alignment stores domain-specific references $\priorPrime=(\mu_{s^{\prime}},\Sigma_{s^{\prime}})$ at the end of each domain and constrains replay-domain statistics as
\begin{equation}
    \mathcal{L}_{\mathrm{align}}
    \;=\;
    \frac{1}{s-1}\sum_{e<s}
    \left(
    \big\|\widehat{\mu}_{s^{\prime}}(\sourceBPrime)-\mu_{s^{\prime}}\big\|_2^2
    +
    \big\|\widehat{\Sigma}_{s^{\prime}}(\sourceBPrime)-\Sigma_{s^{\prime}}\big\|_F^2
    \right) .
\end{equation}

\textbf{CL-MMD.} 
MMD matches feature distributions by minimizing a kernel distance \citep{gretton2012kernel}. We approximate this using a random Fourier feature (RFF) to obtain a differentiable, minibatch-estimable mean embedding as 
\begin{equation}
    \mathbf{z}(\mathbf{x})
    \;=\;
    \mathrm{RFF}\big(f_\theta(\mathbf{x})\big),
    \qquad
    \widehat{\mu}^{z}_{s^{\prime}}(\sourceBPrime)=\mathbb{E}_{\mathbf{x}\sim \sourceBPrime}[\mathbf{z}(\mathbf{x})] .
\end{equation}
Using the replay buffer, we then enforce invariance by matching the mean embeddings across domains in the update:
\begin{equation}
    \mathcal{P}_s^{\mathrm{replay}}
    \;=\;
    \frac{1}{s}\sum_{{s^{\prime}}\le s}
    \left\|
    \widehat{\mu}^{z}_{s^{\prime}}(\sourceBPrime)
    -
    \frac{1}{s}\sum_{j\le s}\widehat{\mu}^{z}_j(B_j)
    \right\|_2^2 .
\end{equation}
For CL-MMD, the sequential invariance alignment stores domain-level reference embeddings $\priorPrime=\mu^{z}_{s^{\prime}}$ and regularizes replay drift:
\begin{equation}
    \mathcal{L}_{\mathrm{align}}
    \;=\;
    \frac{1}{s-1}\sum_{e<s}
    \big\|
    \widehat{\mu}^{z}_{s^{\prime}}(\sourceBPrime)-\mu^z_{s^{\prime}}
    \big\|_2^2 .
\end{equation}

\textbf{Seq-ANDMask.} 
The ANDMask method aims to update parameters using gradient components that agree in sign across domains \citep{parascandolo2020learning}. Using the replay buffer, we treat the current domain minibatch and replayed domain minibatches as a set of environments $\{\sourceBPrime\}_{e\le s}$. For each network parameter, we compute gradients $\{\nabla_{\theta,\omega}\mathcal{L}_{s^{\prime}}\}_{e\le s}$ and form a (soft) agreement mask $\mathbf{m}$:
\begin{equation}
    \mathbf{m}
    \;=\;
    \sigma\!\left(
    \frac{\big|\frac{1}{s}\sum_{{s^{\prime}}\le s}\mathrm{sign}(\mathbf{g}_{s^{\prime}})\big|-\tau}{T}
    \right),
    \qquad
    \mathbf{g}_{s^{\prime}}=\nabla_{\theta,\omega}\mathcal{L}_{\mathrm{ERM}}(\sourceBPrime),
\end{equation}
where $\tau$ is a sparsity threshold, $T$ is a temperature, and $\sigma(\cdot)$ is the sigmoid function. The masked update uses the averaged gradient restricted to agreement directions:
\begin{equation}
    \widetilde{\mathbf{g}}
    \;=\;
    \frac{\mathbf{m}\odot \big(\frac{1}{s}\sum_{{s^{\prime}}\le s}\mathbf{g}_{s^{\prime}}\big)}{\mathrm{mean}(\mathbf{m})+\varepsilon} .
\end{equation}
To align across domains, we combine ANDMask with logit anchoring and treat the KD gradient as an additional signal whose \emph{conflicting} component is removed before aggregation (projecting onto the non-conflicting subspace). We further adapt $\tau$ online to avoid degenerate masks (all-0 or all-1) under limited replay diversity.

\newpage
\section{Pseudocode}
We provide pseudocode for our experimental setup below.

\begin{algorithm}[h]
\caption{Our proposed methods with replay and invariance alignment.}
\label{alg:tailored_seq_dirl}
\begin{algorithmic}[1]
\REQUIRE Sequential source domains $\{D_1,\dots,D_k\}$; target domain $D_t$ (evaluation only); model $h=g_\omega\circ f_\theta$; classification loss $\mathcal{L}$; steps per domain $T$; batch size $B$.
\REQUIRE Memory buffer $M=\bigcup_{{s^{\prime}}<s} M_{{s^{\prime}}}$ storing tuples $(\mathbf{x},y,\tilde{\mathbf{z}},e)$ with domain tag ${s^{\prime}}$.
\REQUIRE Method-specific statistic $\phi(\cdot)$; invariance penalty operator $\textsc{InvPenalty}(\cdot)$.
\REQUIRE Domain priors $\{\priorPrime\}$, running estimator \textsc{Aggregate}$(\cdot)$ (e.g., Welford), and alignment distance $d_{\textsc{align}}(\cdot,\cdot)$.
\REQUIRE Weights $\lambda\ge 0$ (invariance), $\beta\ge 0$ (alignment); optimizer \textsc{OPT}.

\STATE Initialize $(\theta,\omega)$; $M\gets\emptyset$; $\Phi\gets\emptyset$.
\FOR{$s=1$ to $k$} 
    \FOR{$t=1$ to $T$}
        \STATE Sample current minibatch $B_s \sim \sourceD$ with $|B_s|=B$.
        \STATE Initialize multi-domain batch set $\mathcal{B}\gets\{(s,B_s)\}$.
        \IF{$M\neq\emptyset$}
            \STATE Sample replay minibatches $\{\sourceBPrime\}_{e\in\mathcal{E}}$ from buffer partitions $\{M_{e}\}$ (e.g., balanced across domains), where $\mathcal{E}\subseteq\{1,\dots,s-1\}$.
            \STATE $\mathcal{B}\gets\mathcal{B}\cup\{(e,\sourceBPrime)\}_{e\in\mathcal{E}}$.
        \ENDIF

        \STATE \textbf{(1) Replay-augmented ERM:}
        \STATE $\mathcal{L}_{\mathrm{ERM}} \gets \frac{1}{\sum_{(e,\sourceBPrime)\in\mathcal{B}} |\sourceBPrime|}\sum_{(e,\sourceBPrime)\in\mathcal{B}} \sum_{(\mathbf{x},y,\cdot)\in \sourceBPrime}\mathcal{L}\big(h(\mathbf{x};\theta,\omega),y\big)$.

        \STATE \textbf{(2) Replay-enabled invariance penalty:}
        \STATE For each $(e,\sourceBPrime)\in\mathcal{B}$ compute $\widehat{\phi}_{s^{\prime}} \gets \phi\!\left(\sourceBPrime;\theta,\omega\right)$.
        \STATE $\mathcal{P}^{\mathrm{replay}}_s \gets \textsc{InvPenalty}\!\left(\{\widehat{\phi}_{s^{\prime}}\}_{(e,\sourceBPrime)\in\mathcal{B}}\right)$.

        \STATE \textbf{(3) Sequential invariance alignment to domain priors:}
        \IF{$\Phi\neq\emptyset$}
            \STATE $\mathcal{L}_{\mathrm{align}} \gets \frac{1}{|\mathcal{E}|}\sum_{e\in\mathcal{E}} d_{\textsc{align}}\!\left(\widehat{\phi}_{s^{\prime}},\priorPrime\right)$.
        \ELSE
            \STATE $\mathcal{L}_{\mathrm{align}} \gets 0$.
        \ENDIF

        \STATE Total loss and update:
        \STATE $\mathcal{L}_{\mathrm{total}} \gets \mathcal{L}_{\mathrm{ERM}} + \lambda\,\mathcal{P}^{\mathrm{replay}}_s + \beta\,\mathcal{L}_{\mathrm{align}}$.
        \STATE $(\theta,\omega)\gets \textsc{OPT}\big((\theta,\omega), \nabla_{\theta,\omega}\mathcal{L}_{\mathrm{total}}\big)$.
    \ENDFOR

    \STATE Compute and store domain prior for $\sourceD$:
    \STATE $\Phi_s \gets \textsc{Aggregate}\big(\phi(\mathbf{x},y;\theta,\omega)\ \text{for}\ (\mathbf{x},y)\sim \sourceD\big)$.
\ENDFOR
\STATE \textbf{Evaluate} final model $h$ on target domain $D_t$.
\end{algorithmic}
\end{algorithm}

\newpage
\section{Datasets}\label{sec:dataset_details}
We used the following classification datasets. For each datasets, we construct a deployment-centric version by selecting a source-domain sequence and a held-out target domain, on which performance is evaluated.
\begin{enumerate}
    \item \textbf{RotatedMNIST} \citep{ghifary2015domain} applies counter-clockwise rotations to the original MNIST dataset. 
    We use source domains $D^{0},D^{15},D^{30},D^{45}$ and target domain $D^{75}$, where the superscript indicates the rotation in degree applied to the digits.
    The dataset has \num{70000} black-and-white images in resolution $(1\times28\times28)$ which are evenly split into the aforementioned source and target domains. Raw images (i.e., before applying the rotation) are assigned to one domain only, ensuring no (raw-data) overlap across domains. The dataset has \num{10} classes.
    \item \textbf{CIFAR10C} \citep{hendrycks2019benchmarking} applies corruptions to CIFAR10 images.
    To create source and target domains, we joined separate domains from the corruptions described in \citep{hendrycks2019benchmarking}.
    The source domains are $D^{\mathrm{clean}},D^{\mathrm{gaussian+shot}},D^{\mathrm{impulse+defocus}},D^{\mathrm{glass-blurr+motion-blurr}},D^{\mathrm{zoom-blur+snow}},D^{\mathrm{frost+fog}}$ and $D^{\mathrm{brightness+contrast}}$, where the superscript denotes the corruptions applied to the images (except $D^{\mathrm{clean}}$ where no corruptions are applied).
    As target domain, we selected $D^{\mathrm{elastic-transform+pixelate+jpeg-compression+splatter+saturate}}$ because it consists of the most challenging corruptions as mentioned in \citep{hendrycks2019benchmarking}.
    Additionally, the target-domain corruptions present a different type, namely ``digital'' corruptions (plus spatter and saturate, which have minimal overlap to any of the source corruptions).
    We ensured that no (corrupted) image from the target domain is contained in any of the source domains.
    Each domain, except $D^{\mathrm{clean}}$ and the target domain, has \num{16000} color images in $(3\times32\times32)$ resolution.
    $D^{\mathrm{clean}}$ contains the \num{50000} unmodified images from the CIFAR10 training set \citep{Krizhevsky09learningmultiple}.
    The target domain is joined from the five domains indicated in its superscript and consists of \num{10000} images.
    Each domain has ten classes.
    \item \textbf{TinyImageNetC} \citep{hendrycks2019benchmarking} is similar to CIFAR10C except that it applies corruptions to the TinyImageNet dataset \citep{le2015tiny}. 
    The dataset contains $(3\times64\times64)$-sized images from 200 object categories (i.e., classes).
    As before, $D^{\mathrm{clean}}$ contains unmodified images from the training dataset (\num{100000} samples), and the other domains are joint across two (for the source domains) or three (for the target domain, namely $\mathrm{elastic-transform+pixelate+jpeg-compression}$) separate corruption types.
    The source domains contain \num{16000} images each, except $D^{\mathrm{clean}}$ which contains \num{100000} images.
    The target domain contains \num{6000} unseen images; as with CIFAR10C, we ensured that no image from the target domain is contained in any of the source domains. This way, we avoided any information leakage -- essentially leading to a ``double distribution shift'' (unseen images and unseen corruption types).
    \item \textbf{WM811K} \citep{wafermap,wu2015wafer} contains \num{811457} real-world wafer maps and annotations of common failure types. The image data has pixel values $0,1,2$, where $0$ is background, $1$ are OK regions, and $2$ are faulty regions on the wafer. Depending on the clustering of the failures, one of eight class labels is assigned. Images without any failures are labelled with an additional class label. We resize all wafer maps to a common ($48 \times48$) resolution, rescale the data to $(0,1)$ by dividing by $2$, and then normalize using ImageNet statistics.
    We build independent source and train domains by using the provided production lot information to group wafer maps into domain: four non-overlapping source domains $D^{(1)},D^{(2)},D^{(3)},D^{(4)}$ and a fifth disjoint target domain $D^{5}$.
    \item \textbf{Covertype} \citep{blackard1999comparative} contains cartographic and environmental measurements for predicting forest cover type.
    After preprocessing, the dataset consists of \num{65386} instances with \num{50} features, including continuous attributes (e.g., elevation, slope, distances to hydrology, roads, and fire points) and binary indicators encoding, e.g., soil type.
    Each sample is labeled with one of six forest cover types.
    To construct source and target domains, we partitioned the data by wilderness area, which induces natural domain shifts corresponding to distinct geographic regions with different environmental characteristics.
    The source domains are $D^{\mathrm{Rawah}}, D^{\mathrm{Neota}}, D^{\mathrm{Comanche}}$, while the target domain is $D^{\mathrm{CacheLaPoudre}}$, where the superscript denotes the wilderness area from which the samples originate.
    This split ensures that the target domain corresponds to a geographically distinct region not observed during training.
    The source domains contain approximately \num{45000} instances in total, while the target domain consists of approximately \num{9100} instances.
    All domains share the same feature space and label set.
    Within each domain, we balanced the classes by subsampling all classes to the same number, given by the minority class. 
    \item \textbf{Camelyon17} \citep{bandi2018detection,koh2021wilds} contains histopathology image patches extracted from whole-slide lymph node sections collected at five different hospitals.
    Each sample is a $96\times96$ color image and the binary label indicates whether the central $32\times32$ region contains tumor tissue; the domain corresponds to the hospital of origin. 
    We adapt the split proposed by \citep{koh2021wilds}: we use a sequence of four hospitals with approximately \num{340000} class-balanced samples in total as source domain sequence, and use the fifth hospital as the target domain (approximately \num{85000} patches).
\end{enumerate}

\newpage
\section{Hyperparameter details}\label{sec:hparam_details}
We here give details about the hyperparameters and the corresponding search ranges. Several of the hyperparameters are shared across all methods, notably the learning rate, weight decay, and the batch size. The ranges of the hyperparameters of the CL baseline methods are taken from their respective papers. To tune the hyperparameters, we followed common practice in CL and split the data from each \emph{source} domain into a \SI{80}{\percent} training set and \SI{20}{\percent} validation set. The hyperparameters were then tuned using randomized search with 20 trials across the joint distributions described in \cref{tab:hyperparameter_details}. The best hyperparameters were then selected as those that maximize the performance on the validation sets. This way, no information about the unseen target domain is leaked.

This process was repeated three times, yielding 60 trials in total. The numbers reported in this paper are the mean and standard error across the best runs, with one best run per seed (i.e., across three runs/seeds). 

The number of training steps was \emph{not} optimized. All methods thus have the same computational budget, and the number of training steps depends on the dataset: $1000$ for RotatedMNIST, $1200$ for CIFAR10C, $2500$ for TinyImageNetC, $500$ for WM811K and UCICovertype, and $1000$ for Camelyon17.

\begin{table*}[tbph]
\centering
\scriptsize
\setlength{\tabcolsep}{5pt}
\renewcommand{\arraystretch}{1.08}
\captionsetup{font=scriptsize}
\caption{Hyperparameters, default values, and random-search distributions. \textbf{Uniform}$(a,b)$ denotes continuous sampling; \textbf{UniformInt}$(a,b)$ denotes integer sampling; \textbf{Choice}$\{\cdot\}$ denotes discrete sampling. Expressions such as $10^{\mathrm{Uniform}(a,b)}$ and $2^{\mathrm{Uniform}(a,b)}$ follow the code. Parameters marked \emph{Fixed} are not tuned (random value equals default).}
\label{tab:hyperparameter_details}
\begin{adjustbox}{max width=\textwidth}
\begin{tabular}{p{4.2cm} p{5.2cm} p{2.2cm} p{5.2cm}}
\toprule
\textbf{Condition / Method} & \textbf{Hyperparameter (\texttt{name in code})} & \textbf{Default} & \textbf{Random distribution} \\
\midrule

\multicolumn{4}{l}{\textbf{Dataset-specific (shared across algorithms)}}\\
\midrule
RotatedMNIST / Covertype
& Replay buffer capacity
& $1000$
& Fixed$(1000)$ \\
All other datasets
& Replay buffer capacity
& $5000$
& Fixed$(5000)$ \\

\midrule
\multicolumn{4}{l}{\textbf{Learning rate / regularization / batch size (dataset-dependent)}}\\
\midrule
\texttt{SMALL\_DATASETS}$^\dagger$
& Learning rate
& $10^{-3}$
& $10^{\mathrm{Uniform}(-4.5,\,-0.5)}$ \\
Other datasets
& Learning rate
& $5\cdot 10^{-5}$
& $10^{\mathrm{Uniform}(-5,\,-0.5)}$ \\

\texttt{SMALL\_DATASETS}$^\dagger$
& Weight decay
& $0$
& Fixed$(0)$ \\
Other datasets
& Weight decay
& $0$
& $10^{\mathrm{Uniform}(-6,\,-2)}$ \\

\texttt{SMALL\_DATASETS}$^\dagger$
& Batch size
& $64$
& $2^{\mathrm{Uniform}(3,\,8)} $ \\
Other datasets
& Batch size
& $32$
& $ 2^{\mathrm{Uniform}(3,\,5)} $ \\

\midrule
\multicolumn{4}{l}{\textbf{Algorithm-specific hyperparameters}}\\
\midrule

Na{\"i}ve-CL-CORAL
& CORAL penalty weight $\lambda$
& $1.0$
& $10^{\mathrm{Uniform}(-1,\,1)}$ \\

\midrule
\multirow{3}{*}{$\bigstar$-CL-CORAL}
& CORAL penalty weight $\lambda$
& $1.0$
& $10^{\mathrm{Uniform}(-1,\,1)}$ \\
& Alignment-to-prior weight 
& $1.0$
& $10^{\mathrm{Uniform}(-1,\,1)}$ \\

\midrule
\multirow{2}{*}{Na{\"i}ve-CL-MMD}
& MMD penalty weight $\lambda$
& $1.0$
& $10^{\mathrm{Uniform}(-1,\,1)}$ \\
& \#RFF features (\texttt{mmd\_gamma})
& $128$
& Fixed$(128)$ \\

\midrule
\multirow{3}{*}{$\bigstar$-CL-MMD}
& MMD penalty weight $\lambda$
& $1.0$
& $10^{\mathrm{Uniform}(-1,\,1)}$ \\
& Alignment-to-prior weight
& $1.0$
& $10^{\mathrm{Uniform}(-1,\,1)}$ \\
& \#RFF features
& $128$
& Fixed$(128)$ \\

\midrule
\multirow{2}{*}{Na{\"i}ve-CL-VREX}
& V-REx penalty weight $\lambda$
& $10^{1}$
& $10^{\mathrm{Uniform}(-1,\,5)}$ \\
& Penalty anneal iters
& $100$ / $500^{\ddagger}$
& UniformInt$(0,\,1000)$ / UniformInt$(0,\,2500)^{\ddagger}$ \\

\midrule
\multirow{3}{*}{$\bigstar$-CL-VREX}
& V-REx penalty weight $\lambda$
& $10^{1}$
& $10^{\mathrm{Uniform}(-1,\,5)}$ \\
& Penalty anneal iters
& $100$ / $500^{\ddagger}$
& UniformInt$(0,\,1000)$ / UniformInt$(0,\,2500)^{\ddagger}$ \\
& Alignment weight
& $1.0$
& Uniform$(1.0,\,1000.0)$ \\

\midrule
\multirow{2}{*}{Na{\"i}ve-CL-Fishr}
& Fishr penalty weight $\lambda$
& $1000$
& $10^{\mathrm{Uniform}(1,\,4)}$ \\
& Penalty anneal iters
& $100$ / $500^{\ddagger}$
& UniformInt$(0,\,1000)$ / UniformInt$(0,\,2500)^{\ddagger}$ \\

\midrule
\multirow{3}{*}{$\bigstar$-CL-Fishr}
& Fishr penalty weight $\lambda$
& $1000$
& $10^{\mathrm{Uniform}(1,\,4)}$ \\
& Penalty anneal iters
& $100$ / $500^{\ddagger}$
& UniformInt$(0,\,1000)$ / UniformInt$(0,\,2500)^{\ddagger}$ \\
& Alignment weight $\beta$
& $1.0$
& Uniform$(1.0,\,1000.0)$ \\

\midrule
Na{\"i}ve-CL-ANDMASK
& $\tau$
& $1$
& Uniform$(0.5,\,1.0)$ \\

\midrule
\multirow{2}{*}{$\bigstar$-CL-ANDMask}
& Agreement threshold $\tau$
& $1$
& Uniform$(0.5,\,1.0)$ \\
& Alignment weight $\alpha$
& $0.1$
& Uniform$(0.1,\,0.5)$ \\

\midrule
\multirow{2}{*}{EWC \citep{kirkpatrick2017overcoming}}
& Regularization $\lambda$
& $10.0$
& Choice$\{0.1,\,1.0,\,5.0,\,10.0,\,30.0,\,90.0,\,100.0\}$ \\
& Fisher decay $\gamma$
& $0.9$
& Choice$\{0.8,\,0.9,\,1.0\}$ \\

\midrule
\multirow{2}{*}{SI \citep{zenke2017continual}}
& $\xi$
& $0.1$
& Uniform$(0.001,\,0.1)$ \\
& $c$
& $0.1$
& Uniform$(0.001,\,0.1)$ \\

\midrule
\multirow{2}{*}{STAR \citep{eskandar2025star}}
& $\gamma$
& $0.01$
& Uniform$(0.01,\,0.05)$ \\
& $\lambda$
& $0.01$
& Uniform$(0.01,\,0.05)$ \\

\midrule
LODE \citep{liang2023loss}
& $\rho$
& $0.1$
& Choice$\{0.01,\,0.05,\,0.1,\,0.2\}$ \\

\midrule
\multirow{3}{*}{COPE \citep{de2021continual}}
& $\tau$
& $1$
& Uniform$(0.5,\,1.0)$ \\
& $\gamma$
& $0.9$
& Choice$\{0.8,\,0.9,\,1.0\}$ \\
& Inner repeats
& $3$
& Choice$\{1,\,2,\,3,\,4,\,5\}$ \\

\midrule
\multirow{6}{*}{SARL \citep{sarfraz025semantic}}
& Replay logit anchoring
& $1.0$
& Choice$\{0.2,\,0.5,\,1.0\}$ \\
& Stable-model anchoring
& $1.0$
& Fixed$(1.0)$ \\
& Regularization
& $1.0$
& Choice$\{0.2,\,0.5,\,1.0\}$ \\
& Semantics $\tau$
& $0.8$
& Fixed$(0.8)$ \\
& Semantics weight
& $0.01$
& Fixed$(0.01)$ \\
& Warmup steps 
& $100$ / $500^{\ddagger}$
& UniformInt$(0,\,1000)$ / UniformInt$(0,\,2500)^{\ddagger}$ \\

\midrule
\multirow{3}{*}{EFC}
& $\lambda$
& $0.4$
& Uniform$(0.01,\,0.5)$ \\
& $\alpha$
& $0.5$
& Uniform$(0.01,\,0.99)$ \\
& $\delta$
& $0.9$
& Uniform$(0.001,\,0.1)$ \\

\bottomrule
\end{tabular}
\end{adjustbox}

\vspace{0.6em}
\scriptsize{$^\dagger$\texttt{SMALL\_DATASETS} = \{RotatedMNIST, CIFAR10C, Covertype\}. \\
$^\ddagger$For datasets in \{RotatedMNIST, Covertype, Camelyon17\}, the default range is $100$ and UniformInt$(0,1000)$; otherwise $500$ and UniformInt$(0,2500)$}
\end{table*}

\newpage
\section{Extended results}

\subsection{Comparison against CDA/CTTA baselines}
In \cref{tab:cda_ctta_results}, we provide experimental results for CDA/CTTA baselines TENT~\citep{wang2021tent}, SHOT++~\citep{liang2021source}, and CoTTA~\citep{wang2022continual}. We observe that our proposed $\bigstar$-CL methods outperform the baselines by a substantial margin.

\begin{table*}[!htbp]
\caption{\textbf{Extended main results:}
{\setlength{\fboxsep}{0pt}
Comparison of \colorbox{color_tailored}{our proposed methods} against \colorbox{color_ctta}{CDA and CTTA baselines} on six benchmark datasets. For these baselines, we allow unsupervised updates on the target domain. We report the mean$\pm$standard error \emph{target domain performance} across three independent runs. Results marked as \textbf{best}, \underline{second}, \dotuline{third}.
}
}
\label{tab:cda_ctta_results}
\adjustbox{max width=\textwidth}{%
\begin{tabular}{l|ccccccc|ccc}
\toprule
\textbf{Method}                                                           & \textbf{RotatedMNIST}                                                     & \textbf{CIFAR10C}                                                         & \textbf{TinyImageNetC}                                                    & \textbf{WM811K}                                                           & \textbf{Covertype}                                                        & \textbf{Camelyon17}                                                       & \textbf{Avg}                                                              & \textbf{Arith. mean}                                                      & \textbf{Geom. mean}                                                       & \textbf{Median}                                                           \\
\midrule
\rowcolor{color_finetune} Finetune                                        & 39.5 $\pm$ 1.7                                                            & 66.3 $\pm$ 0.4                                                            & 18.5 $\pm$ 2.6                                                            & 82.8 $\pm$ 0.3                                                            & 8.1 $\pm$ 0.0                                                             & 86.9 $\pm$ 2.1                                                            & 50.4                                                                      & 6.7                                                                       & 6.6                                                                       & 6.5                                                                       \\
\midrule
\rowcolor{color_tailored} $\bigstar$-CL-VREX                              & 70.2 $\pm$ 5.8                                                            & 67.2 $\pm$ 1.4                                                            & \underline{26.3 $\pm$ 1.2}                                                & 84.1 $\pm$ 0.3                                                            & \dotuline{40.8 $\pm$ 1.5}                                                 & \underline{91.5 $\pm$ 0.6}                                                & \underline{63.4}                                                          & \dotuline{3.2}                                                            & \dotuline{3.0}                                                            & \dotuline{3.5}                                                            \\
\rowcolor{color_tailored} $\bigstar$-CL-Fishr                             & 68.6 $\pm$ 0.4                                                            & 64.2 $\pm$ 1.2                                                            & \textbf{29.0 $\pm$ 2.2}                                                   & 83.6 $\pm$ 0.5                                                            & 27.0 $\pm$ 7.7                                                            & 89.7 $\pm$ 2.0                                                            & 60.3                                                                      & 5.0                                                                       & 4.3                                                                       & 5.0                                                                       \\
\rowcolor{color_tailored} $\bigstar$-CL-CORAL                             & \textbf{72.8 $\pm$ 2.6}                                                   & \underline{68.5 $\pm$ 1.7}                                                & 25.0 $\pm$ 2.8                                                            & \underline{84.8 $\pm$ 0.6}                                                & \textbf{45.2 $\pm$ 3.9}                                                   & \textbf{91.7 $\pm$ 0.4}                                                   & \textbf{64.7}                                                             & \textbf{1.8}                                                              & \textbf{1.6}                                                              & \textbf{1.5}                                                              \\
\rowcolor{color_tailored} $\bigstar$-CL-MMD                               & \dotuline{70.7 $\pm$ 0.5}                                                 & \textbf{69.0 $\pm$ 0.5}                                                   & \dotuline{25.8 $\pm$ 0.8}                                                 & \textbf{85.5 $\pm$ 0.4}                                                   & 37.6 $\pm$ 0.9                                                            & 90.1 $\pm$ 0.5                                                            & \dotuline{63.1}                                                           & \underline{2.7}                                                           & \underline{2.3}                                                           & \underline{3.0}                                                           \\
\rowcolor{color_tailored} $\bigstar$-CL-ANDMask                           & \underline{71.4 $\pm$ 2.6}                                                & 64.7 $\pm$ 0.7                                                            & 11.8 $\pm$ 3.6                                                            & \dotuline{84.6 $\pm$ 0.6}                                                 & \underline{43.7 $\pm$ 2.9}                                                & 89.1 $\pm$ 1.5                                                            & 60.9                                                                      & 4.3                                                                       & 3.8                                                                       & 4.5                                                                       \\
\midrule
\rowcolor{color_ctta} TENT                                                & 28.0 $\pm$ 0.6                                                            & \dotuline{67.9 $\pm$ 1.0}                                                 & 20.2 $\pm$ 0.2                                                            & 77.1 $\pm$ 0.4                                                            & 36.8 $\pm$ 1.8                                                            & \dotuline{91.4 $\pm$ 2.1}                                                 & 53.6                                                                      & 5.7                                                                       & 5.2                                                                       & 5.5                                                                       \\
\rowcolor{color_ctta} CoTTA                                               & 29.8 $\pm$ 9.5                                                            & 62.6 $\pm$ 4.3                                                            & 7.3 $\pm$ 5.6                                                             & 78.2 $\pm$ 0.3                                                            & 36.9 $\pm$ 0.9                                                            & 82.5 $\pm$ 2.4                                                            & 49.5                                                                      & 7.3                                                                       & 7.2                                                                       & 8.0                                                                       \\
\rowcolor{color_ctta} SHOTPP                                                                    & 49.8 $\pm$ 10.1                                                           & 19.1 $\pm$ 1.8                                                            & 1.0 $\pm$ 0.3                                                             & 57.5 $\pm$ 0.6                                                            & 22.3 $\pm$ 9.1                                                            & 67.8 $\pm$ 6.6                                                            & 36.2                                                                      & 8.3                                                                       & 8.2                                                                       & 9.0                                                                       \\
\bottomrule
\end{tabular}}
\tiny{Macro F1 for WM811K, accuracy for other datasets}
\end{table*}

\subsection{Na{\"i}ve results}
We here give the full tabular results for the na\"ive extensions.

\begin{table}[htbp]
\centering
\caption{Na\"ive extensions versus CL baselines. Shown: mean$\pm$standard error target domain accuracy across three independent runs. Marked: \textbf{best}, \underline{second}, \dotuline{third}.}
\label{tab:naive_results_full}
\adjustbox{max width=\textwidth}{%

\begin{tabular}{lcccccccc}
\toprule
\textbf{Method}                                                           & \textbf{RotatedMNIST}                                                     & \textbf{CIFAR10C}                                                         & \textbf{TinyImageNetC}                                                    & \textbf{WM811K}                                                           & \textbf{Covertype}                                                        & \textbf{Camelyon17}                                                       & \textbf{Avg}                                                              & \textbf{$\Delta$ to ERM}                                                     \\
\midrule
\rowcolor{color_finetune} Finetune                                        & 39.5 $\pm$ 1.7                                                            & 66.3 $\pm$ 0.4                                                            & 18.5 $\pm$ 2.6                                                            & 82.8 $\pm$ 0.3                                                            & 8.1 $\pm$ 0.0                                                             & 86.9 $\pm$ 2.1                                                            & 50.4                                                                      & +0.0                                                                      \\
\midrule
\rowcolor{color_optimization} AGEM                                        & \underline{54.8 $\pm$ 4.7}                                                & \textbf{69.2 $\pm$ 0.2}                                                   & \dotuline{22.8 $\pm$ 1.1}                                                 & \dotuline{84.1 $\pm$ 0.2}                                                 & \underline{29.0 $\pm$ 8.5}                                                & \textbf{91.0 $\pm$ 0.1}                                                   & \underline{58.5}                                                          & \underline{+8.1}                                                          \\
\midrule
\rowcolor{color_regularization} SI                                        & 38.8 $\pm$ 0.7                                                            & 64.3 $\pm$ 2.4                                                            & 20.8 $\pm$ 1.2                                                            & 83.3 $\pm$ 0.6                                                            & \dotuline{24.5 $\pm$ 6.8}                                                 & 86.0 $\pm$ 3.6                                                            & 52.9                                                                      & +2.6                                                                      \\
\midrule
\rowcolor{color_replay} ER-ACE                                            & \textbf{67.5 $\pm$ 2.0}                                                   & 66.9 $\pm$ 1.1                                                            & \textbf{29.0 $\pm$ 1.2}                                                   & \textbf{85.4 $\pm$ 0.4}                                                   & \textbf{37.7 $\pm$ 0.6}                                                   & \dotuline{90.3 $\pm$ 1.5}                                                 & \textbf{62.8}                                                             & \textbf{+12.4}                                                            \\
\midrule
\rowcolor{color_naive} Na{\"i}ve-CL-VREX                                & 45.8 $\pm$ 6.8                                                            & \dotuline{67.4 $\pm$ 0.3}                                                 & 20.0 $\pm$ 1.3                                                            & 84.0 $\pm$ 0.1                                                            & 8.1 $\pm$ 0.1                                                             & 88.7 $\pm$ 1.5                                                            & 52.3                                                                      & +1.9                                                                      \\
\rowcolor{color_naive} Na{\"i}ve-CL-Fishr                               & 41.4 $\pm$ 1.5                                                            & \underline{67.7 $\pm$ 0.3}                                                & 22.2 $\pm$ 2.2                                                            & 83.8 $\pm$ 0.5                                                            & 8.2 $\pm$ 0.1                                                             & 84.6 $\pm$ 5.5                                                            & 51.3                                                                      & +0.9                                                                      \\
\rowcolor{color_naive} Na{\"i}ve-CL-CORAL                               & \dotuline{50.2 $\pm$ 5.3}                                                 & 66.3 $\pm$ 0.8                                                            & \underline{23.6 $\pm$ 0.6}                                                & 83.6 $\pm$ 0.1                                                            & 8.0 $\pm$ 0.1                                                             & 90.1 $\pm$ 0.7                                                            & \dotuline{53.6}                                                           & \dotuline{+3.3}                                                           \\
\rowcolor{color_naive} Na{\"i}ve-CL-MMD                                 & 39.9 $\pm$ 1.0                                                            & 66.3 $\pm$ 0.8                                                            & 19.8 $\pm$ 1.0                                                            & \underline{84.3 $\pm$ 0.3}                                                & 7.8 $\pm$ 0.3                                                             & \underline{90.7 $\pm$ 1.9}                                                & 51.5                                                                      & +1.1                                                                      \\
\rowcolor{color_naive} Na{\"i}ve-CL-ANDMask                             & 21.5 $\pm$ 4.5                                                            & 26.9 $\pm$ 1.4                                                            & 0.5 $\pm$ 0.1                                                             & 8.7 $\pm$ 0.9                                                             & 6.6 $\pm$ 1.2                                                             & 49.5 $\pm$ 1.0                                                            & 19.0                                                                      & -31.4                                                                     \\
\bottomrule
\end{tabular}}
\tiny{Macro F1 for WM811K, accuracy for other datasets}
\end{table}

\subsection{Na{\"i}ve versus tailored methods}
We here give the full tabular results for benchmarking our proposed methods against the na\"ive counterparts.

\begin{table}[!htbp]
    \centering
    \caption{Our methods against the na{\"i}ve baselines. Our methods considerably outperform their na\"ive counterparts. Shown: mean$\pm$standard error target domain performance across three independent runs, using macro F1 for WM811K and accuracy for the other datasets. Marked: \textbf{best}, \underline{second}, \dotuline{third}.}
    \label{tab:naive_vs_tailored_full}
    \adjustbox{max width=\linewidth}{%
\begin{tabular}{lccccccc}
\toprule
\textbf{Method}                            & \textbf{RotatedMNIST}                      & \textbf{CIFAR10C}                          & \textbf{TinyImageNetC}                     & \textbf{WM811K}                            & \textbf{Covertype}                         & \textbf{Camelyon17}                        & \textbf{Avg}                               \\
\midrule
\rowcolor{color_finetune} Finetune                                        & 39.5 $\pm$ 1.7                                                            & 66.3 $\pm$ 0.4                                                            & 18.5 $\pm$ 2.6                                                            & 82.8 $\pm$ 0.3                                                            & 8.1 $\pm$ 0.0                                                             & 86.9 $\pm$ 2.1                                                            & 50.4                                                                      \\
\midrule
\rowcolor{color_naive} Na{\"i}ve-CL-VREX                                & 45.8 $\pm$ 6.8                                                            & 67.4 $\pm$ 0.3                                                            & 20.0 $\pm$ 1.3                                                            & 84.0 $\pm$ 0.1                                                            & 8.1 $\pm$ 0.1                                                             & 88.7 $\pm$ 1.5                                                            & 52.3                                                                      \\
\rowcolor{color_naive} Na{\"i}ve-CL-Fishr                               & 41.4 $\pm$ 1.5                                                            & \dotuline{67.7 $\pm$ 0.3}                                                 & 22.2 $\pm$ 2.2                                                            & 83.8 $\pm$ 0.5                                                            & 8.2 $\pm$ 0.1                                                             & 84.6 $\pm$ 5.5                                                            & 51.3                                                                      \\
\rowcolor{color_naive} Na{\"i}ve-CL-CORAL                               & 50.2 $\pm$ 5.3                                                            & 66.3 $\pm$ 0.8                                                            & 23.6 $\pm$ 0.6                                                            & 83.6 $\pm$ 0.1                                                            & 8.0 $\pm$ 0.1                                                             & 90.1 $\pm$ 0.7                                                            & 53.6                                                                      \\
\rowcolor{color_naive} Na{\"i}ve-CL-MMD                                 & 39.9 $\pm$ 1.0                                                            & 66.3 $\pm$ 0.8                                                            & 19.8 $\pm$ 1.0                                                            & 84.3 $\pm$ 0.3                                                            & 7.8 $\pm$ 0.3                                                             & \dotuline{90.7 $\pm$ 1.9}                                                 & 51.5                                                                      \\
\rowcolor{color_naive} Na{\"i}ve-CL-ANDMask                             & 21.5 $\pm$ 4.5                                                            & 26.9 $\pm$ 1.4                                                            & 0.5 $\pm$ 0.1                                                             & 8.7 $\pm$ 0.9                                                             & 6.6 $\pm$ 1.2                                                             & 49.5 $\pm$ 1.0                                                            & 19.0                                                                      \\
\midrule
\rowcolor{color_tailored} $\bigstar$-CL-VREX                              & 70.2 $\pm$ 5.8                                                            & 67.2 $\pm$ 1.4                                                            & \underline{26.3 $\pm$ 1.2}                                                & 84.1 $\pm$ 0.3                                                            & \dotuline{40.8 $\pm$ 1.5}                                                 & \underline{91.5 $\pm$ 0.6}                                                & \underline{63.4}                                                          \\
\rowcolor{color_tailored} $\bigstar$-CL-Fishr                             & 68.6 $\pm$ 0.4                                                            & 64.2 $\pm$ 1.2                                                            & \textbf{29.0 $\pm$ 2.2}                                                   & 83.6 $\pm$ 0.5                                                            & 27.0 $\pm$ 7.7                                                            & 89.7 $\pm$ 2.0                                                            & 60.3                                                                      \\
\rowcolor{color_tailored} $\bigstar$-CL-CORAL                             & \textbf{72.8 $\pm$ 2.6}                                                   & \underline{68.5 $\pm$ 1.7}                                                & 25.0 $\pm$ 2.8                                                            & \underline{84.8 $\pm$ 0.6}                                                & \textbf{45.2 $\pm$ 3.9}                                                   & \textbf{91.7 $\pm$ 0.4}                                                   & \textbf{64.7}                                                             \\
\rowcolor{color_tailored} $\bigstar$-CL-MMD                               & \dotuline{70.7 $\pm$ 0.5}                                                 & \textbf{69.0 $\pm$ 0.5}                                                   & \dotuline{25.8 $\pm$ 0.8}                                                 & \textbf{85.5 $\pm$ 0.4}                                                   & 37.6 $\pm$ 0.9                                                            & 90.1 $\pm$ 0.5                                                            & \dotuline{63.1}                                                           \\
\rowcolor{color_tailored} $\bigstar$-CL-ANDMask                           & \underline{71.4 $\pm$ 2.6}                                                & 64.7 $\pm$ 0.7                                                            & 11.8 $\pm$ 3.6                                                            & \dotuline{84.6 $\pm$ 0.6}                                                 & \underline{43.7 $\pm$ 2.9}                                                & 89.1 $\pm$ 1.5                                                            & 60.9                                                                      \\
\bottomrule
\end{tabular}}
\end{table}

\subsection{Backwards transfer}
We here give results for computing the backwards transfer metric (BWT) \citep{lopez2017gradient}. BWT measures the performance loss by contrasting the original performance on a source domain with the performance after training. For BWT, negative numbers imply forgetting, while positive numbers indicate a retrospective accuracy improvement (which is desired). The BWT results are in \cref{tab:bwt_results}.

\begin{table*}[htbp]
\centering
\caption{Backwards transfer for the main results. Shown: mean$\pm$standard error source domain backwards transfer in percentage points across three trials.}
\label{tab:bwt_results}
\adjustbox{max width=\textwidth}{%
\begin{tabular}{lccccccc}
\toprule
\textbf{Method}                            & \textbf{RotatedMNIST}                      & \textbf{CIFAR10C}                          & \textbf{TinyImageNetC}                     & \textbf{WM811K}                            & \textbf{Covertype}                         & \textbf{Camelyon17}                        & \textbf{Avg}                               \\
\midrule
\rowcolor{color_finetune} Finetune                                        & 7.5 $\pm$ 4.7                                                             & \dotuline{11.4 $\pm$ 3.0}                                                 & \underline{25.5 $\pm$ 1.8}                                                & 5.1 $\pm$ 0.8                                                             & 11.8 $\pm$ 3.1                                                            & 6.5 $\pm$ 2.7                                                             & \dotuline{11.3}                                                           \\
\midrule
\rowcolor{color_optimization} AGEM                                              & 4.5 $\pm$ 3.5                                                             & 8.2 $\pm$ 1.9                                                             & 9.9 $\pm$ 8.1                                                             & 5.8 $\pm$ 1.1                                                             & 11.7 $\pm$ 2.3                                                            & 5.3 $\pm$ 1.5                                                             & 7.6                                                                       \\
\rowcolor{color_optimization} UPGD                                        & \underline{13.9 $\pm$ 2.2}                                                & 9.0 $\pm$ 1.6                                                             & 17.7 $\pm$ 6.1                                                            & \textbf{13.1 $\pm$ 1.0}                                                   & -3.7 $\pm$ 2.4                                                            & 3.8 $\pm$ 1.4                                                             & 9.0                                                                       \\
\midrule
\rowcolor{color_regularization} EWC                                       & 7.3 $\pm$ 4.1                                                             & \underline{12.2 $\pm$ 2.5}                                                & 13.3 $\pm$ 11.8                                                           & 4.3 $\pm$ 1.0                                                             & \textbf{21.8 $\pm$ 1.1}                                                   & \textbf{9.9 $\pm$ 0.7}                                                    & \textbf{11.5}                                                             \\
\rowcolor{color_regularization} SI                                        & 7.5 $\pm$ 4.2                                                             & 6.6 $\pm$ 1.1                                                             & 6.6 $\pm$ 9.6                                                             & 5.2 $\pm$ 0.7                                                             & 12.2 $\pm$ 5.4                                                            & \underline{9.5 $\pm$ 4.0}                                                 & 7.9                                                                       \\
\rowcolor{color_regularization} SNR                                       & 7.9 $\pm$ 4.0                                                             & \textbf{12.3 $\pm$ 2.4}                                                   & 19.2 $\pm$ 13.7                                                           & \underline{8.6 $\pm$ 1.8}                                                 & 9.8 $\pm$ 3.1                                                             & 7.7 $\pm$ 5.5                                                             & 10.9                                                                      \\
\midrule
\rowcolor{color_replay} FDR                                               & 4.3 $\pm$ 3.0                                                             & 10.2 $\pm$ 1.5                                                            & 8.4 $\pm$ 4.9                                                             & 3.5 $\pm$ 0.6                                                             & 18.3 $\pm$ 2.9                                                            & 2.9 $\pm$ 0.2                                                             & 7.9                                                                       \\
\rowcolor{color_replay} ER-ACE                                            & 10.3 $\pm$ 4.5                                                            & 7.1 $\pm$ 2.3                                                             & 13.9 $\pm$ 1.7                                                            & 4.8 $\pm$ 0.6                                                             & 18.2 $\pm$ 2.1                                                            & 3.0 $\pm$ 0.4                                                             & 9.5                                                                       \\
\rowcolor{color_replay} LODE                                              & 8.4 $\pm$ 4.7                                                             & 9.7 $\pm$ 2.6                                                             & \textbf{29.9 $\pm$ 2.7}                                                   & 4.0 $\pm$ 0.4                                                             & 8.0 $\pm$ 5.4                                                             & \dotuline{8.4 $\pm$ 3.4}                                                  & \underline{11.4}                                                          \\
\rowcolor{color_replay} STAR                                              & \textbf{18.4 $\pm$ 1.9}                                                   & 7.4 $\pm$ 1.8                                                             & 7.9 $\pm$ 1.9                                                             & 5.0 $\pm$ 0.2                                                             & \underline{20.9 $\pm$ 0.8}                                                & 4.2 $\pm$ 0.1                                                             & 10.6                                                                      \\
\rowcolor{color_replay} COPE                                              & 10.4 $\pm$ 4.2                                                            & 9.1 $\pm$ 1.5                                                             & 5.9 $\pm$ 13.7                                                            & 5.6 $\pm$ 1.4                                                             & 1.8 $\pm$ 1.4                                                             & 7.0 $\pm$ 1.7                                                             & 6.6                                                                       \\
\rowcolor{color_replay} EFC                                               & 4.1 $\pm$ 2.9                                                             & 10.7 $\pm$ 1.8                                                            & 19.6 $\pm$ 3.9                                                            & 3.1 $\pm$ 0.7                                                             & 11.1 $\pm$ 2.3                                                            & 5.6 $\pm$ 1.4                                                             & 9.0                                                                       \\
\rowcolor{color_replay} SARL                                              & 0.8 $\pm$ 0.5                                                             & 2.5 $\pm$ 0.7                                                             & -0.8 $\pm$ 0.5                                                            & 4.9 $\pm$ 0.7                                                             & \dotuline{18.9 $\pm$ 3.1}                                                 & -1.3 $\pm$ 1.8                                                            & 4.2                                                                       \\
\midrule
\rowcolor{color_tailored} $\bigstar$-CL-VREX                              & 2.9 $\pm$ 1.9                                                             & 8.5 $\pm$ 2.4                                                             & 15.3 $\pm$ 2.3                                                            & 4.6 $\pm$ 0.4                                                             & 13.4 $\pm$ 3.3                                                            & 6.0 $\pm$ 1.7                                                             & 8.4                                                                       \\
\rowcolor{color_tailored} $\bigstar$-CL-Fishr                             & 10.9 $\pm$ 3.9                                                            & 3.4 $\pm$ 1.3                                                             & 16.9 $\pm$ 1.1                                                            & 2.9 $\pm$ 0.5                                                             & 8.0 $\pm$ 0.9                                                             & 3.8 $\pm$ 0.5                                                             & 7.7                                                                       \\
\rowcolor{color_tailored} $\bigstar$-CL-CORAL                             & 10.3 $\pm$ 4.6                                                            & 7.4 $\pm$ 1.8                                                             & \dotuline{22.7 $\pm$ 5.3}                                                 & \dotuline{7.7 $\pm$ 1.1}                                                  & 12.3 $\pm$ 1.7                                                            & 3.5 $\pm$ 1.0                                                             & 10.7                                                                      \\
\rowcolor{color_tailored} $\bigstar$-CL-MMD                               & \dotuline{11.1 $\pm$ 4.3}                                                 & 7.1 $\pm$ 1.3                                                             & 8.0 $\pm$ 1.6                                                             & 6.8 $\pm$ 1.4                                                             & 14.1 $\pm$ 1.9                                                            & 2.2 $\pm$ 0.4                                                             & 8.2                                                                       \\
\rowcolor{color_tailored} $\bigstar$-CL-ANDMask                           & 10.4 $\pm$ 4.6                                                            & 8.4 $\pm$ 0.9                                                             & 8.7 $\pm$ 6.4                                                             & 4.9 $\pm$ 0.8                                                             & 12.7 $\pm$ 3.0                                                            & 5.2 $\pm$ 1.5                                                             & 8.4                                                                       \\
\bottomrule
\end{tabular}}
\end{table*}

\subsection{Runtimes}
\begin{figure}[!htbp]
\centering
    \begin{subfigure}{.33\linewidth}
        \centering
        \includegraphics[width=\linewidth]{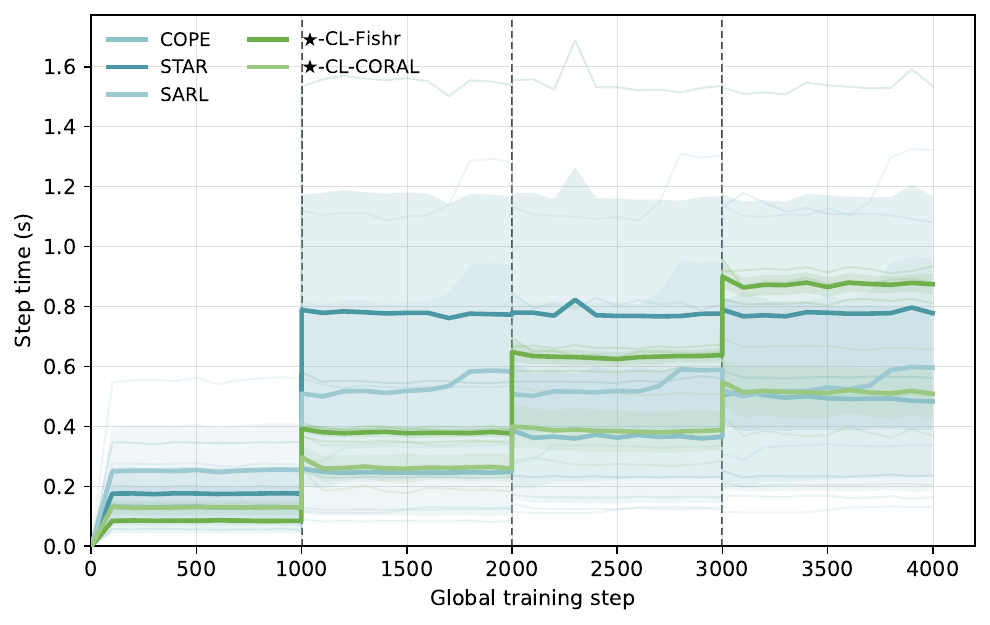}
    \end{subfigure}%
    \begin{subfigure}{.33\linewidth}
        \centering
        \includegraphics[width=\linewidth]{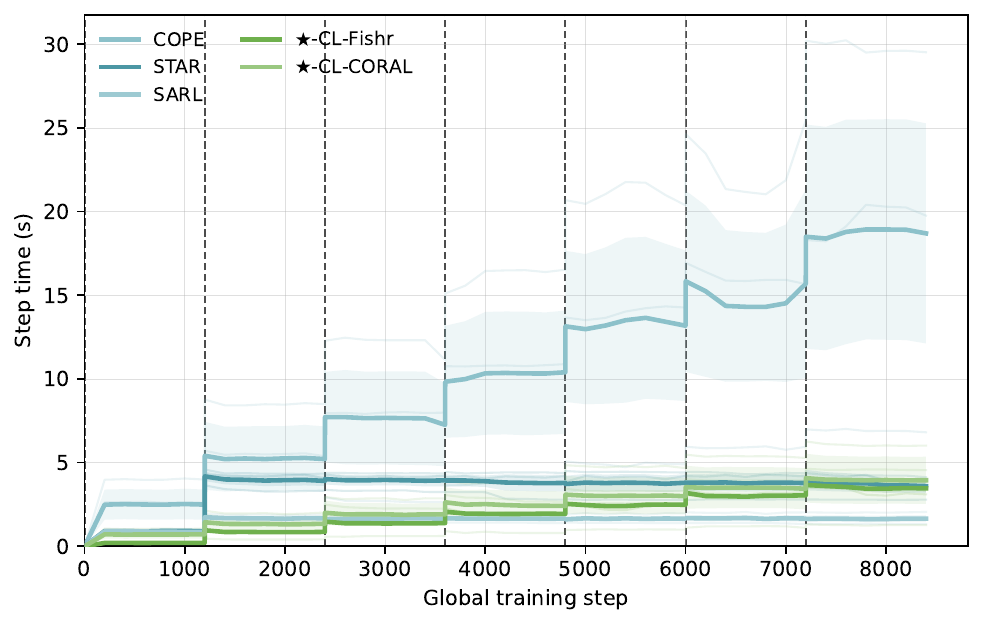}
    \end{subfigure}%
    \begin{subfigure}{.33\linewidth}
        \centering
        \includegraphics[width=\linewidth]{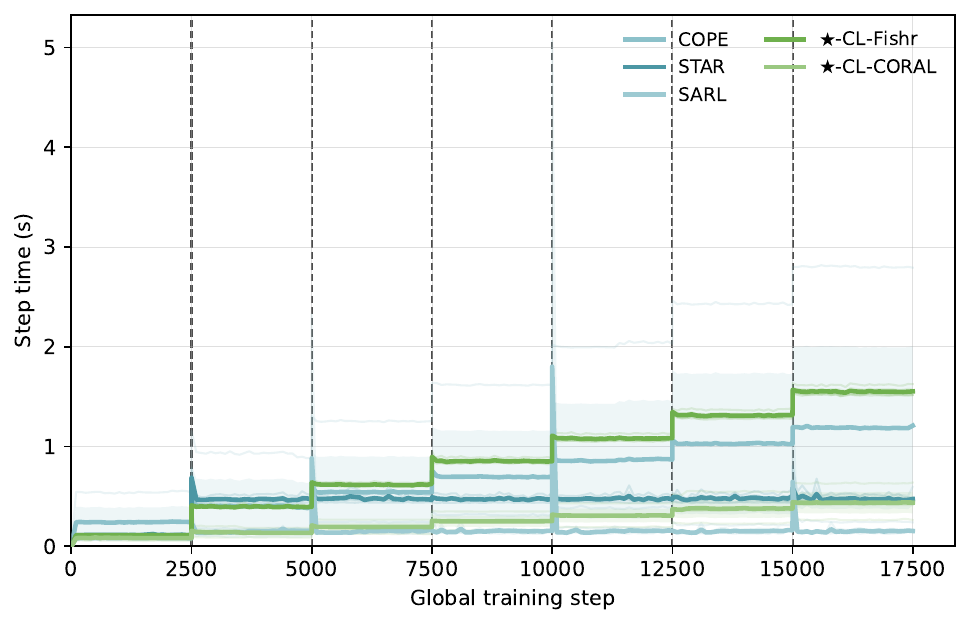}
    \end{subfigure}
    \caption{
    {
    \setlength{\fboxsep}{0pt}
    \textbf{Average per-step runtimes over time.}
    We plot the average per-step runtime for methods COPE, STAR, SARL, $\bigstar$-CL-Fishr and $\bigstar$-CL-CORAL across datasets RotatedMNIST (left), CIFAR10C (middle), and TinyImageNetC (right).
    }}
    \label{fig:runtimes_1}
\end{figure}

\begin{figure}[!htbp]
\centering
    \begin{subfigure}{.45\linewidth}
        \centering
        \includegraphics[width=\linewidth]{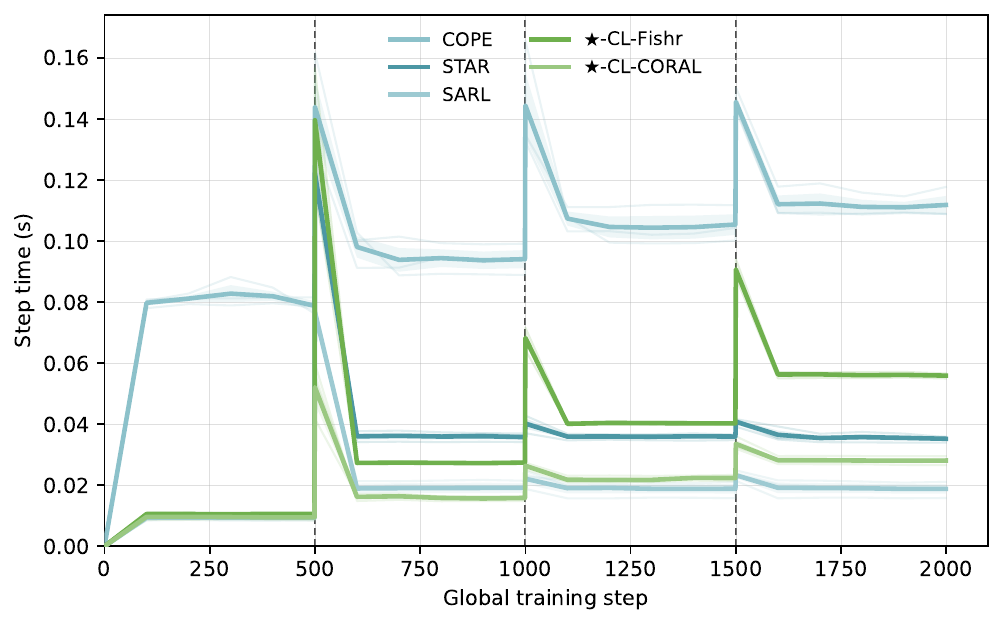}
    \end{subfigure}%
    \begin{subfigure}{.45\linewidth}
        \centering
        \includegraphics[width=\linewidth]{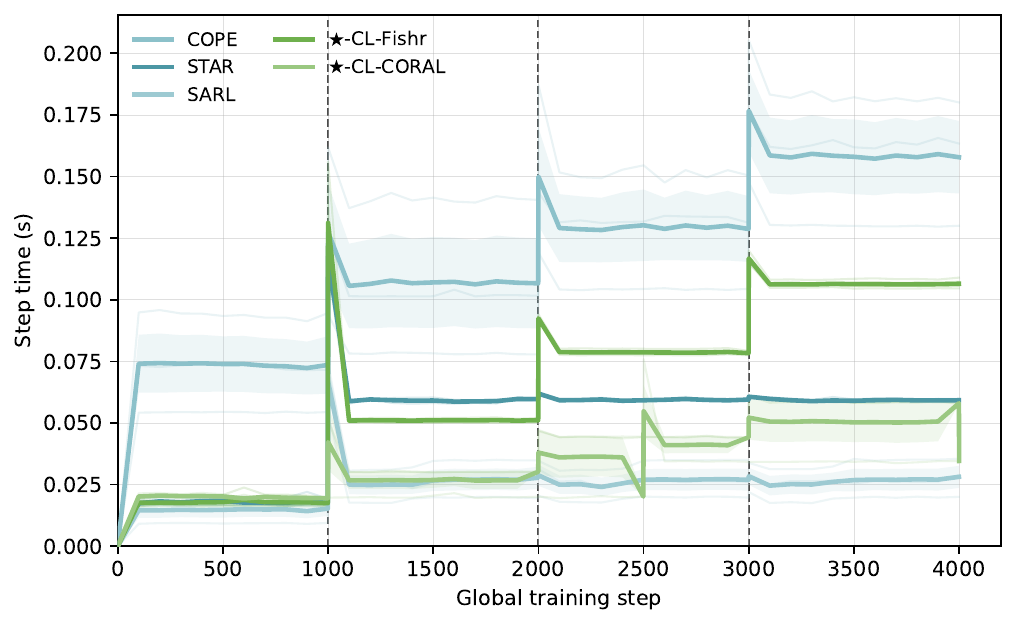}
    \end{subfigure}
    \caption{
    {
    \setlength{\fboxsep}{0pt}
    \textbf{Average per-step runtimes over time.}
        We plot the average per-step runtime for methods COPE, STAR, SARL, $\bigstar$-CL-Fishr and $\bigstar$-CL-CORAL across datasets RotatedMNIST (left), CIFAR10C (middle), and TinyImageNetC (right).}}
    \label{fig:runtimes_2}
\end{figure}

\begin{figure}[!htbp]
\centering
    \begin{subfigure}{.33\linewidth}
        \centering
        \includegraphics[width=\linewidth]{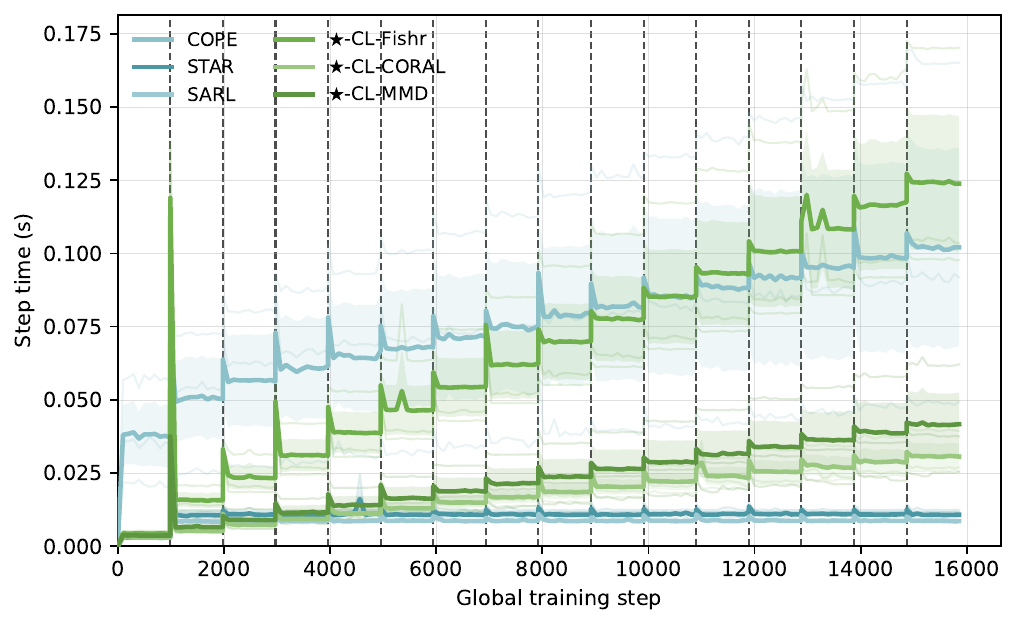}
    \end{subfigure}%
    \begin{subfigure}{.33\linewidth}
        \centering
        \includegraphics[width=\linewidth]{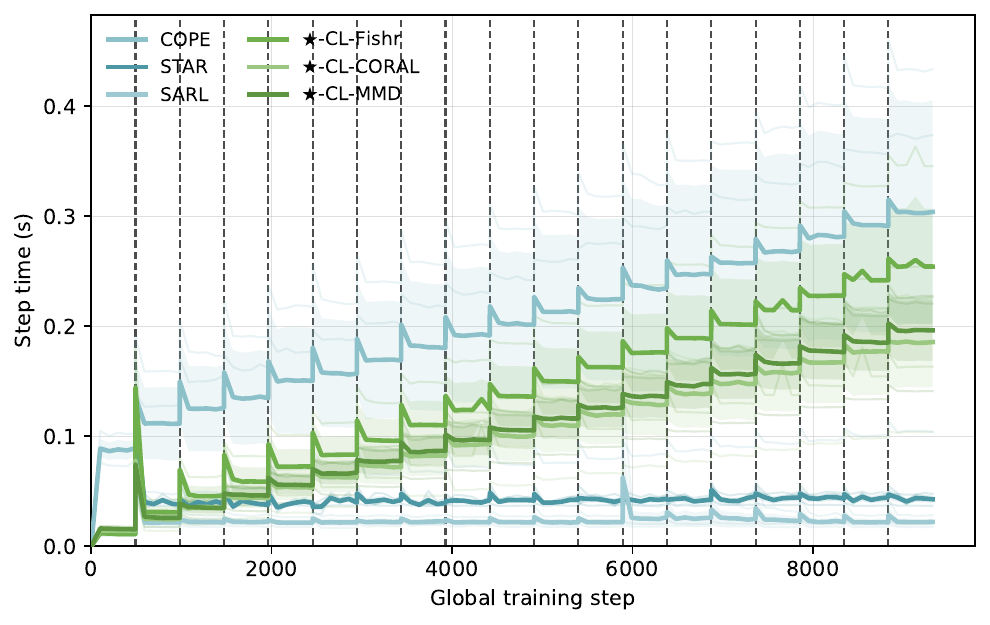}
    \end{subfigure}
    \begin{subfigure}{.33\linewidth}
        \centering
        \includegraphics[width=\linewidth]{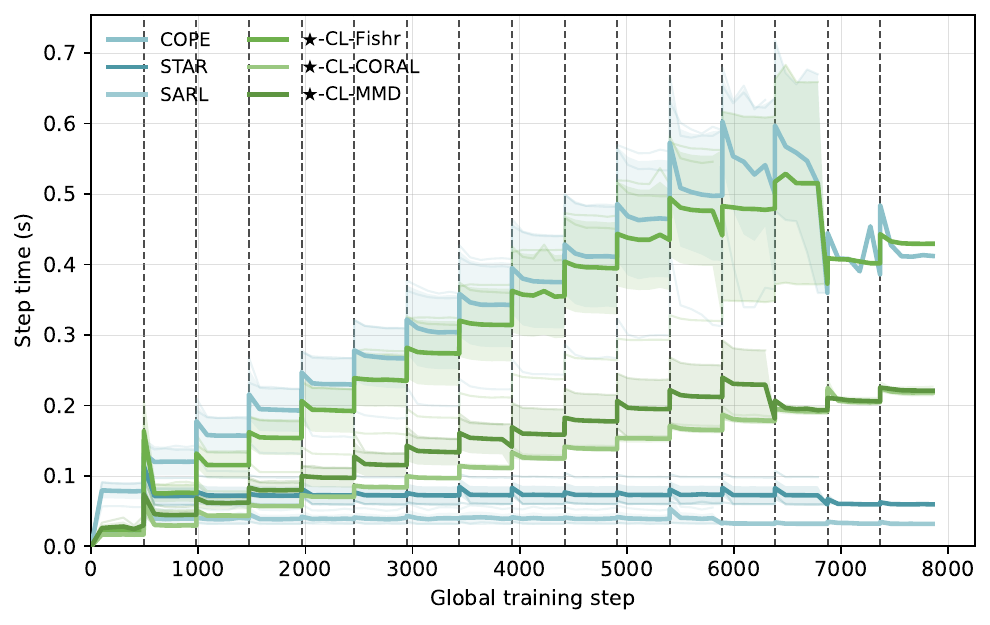}
    \end{subfigure}
    \caption{
    {
    \setlength{\fboxsep}{0pt}
    \textbf{Average per-step runtimes over time (extended datasets).}
        We plot the average per-step runtime for methods COPE, STAR, SARL, $\bigstar$-CL-Fishr and $\bigstar$-CL-CORAL across extended datasets (15 source domains) RotatedMNISTExtended (left), WM811KExtended (middle), and Camelyon17Extended (right).}}
    \label{fig:runtimes_extended}
\end{figure}

We here provide runtimes. We provide the total average runtimes in \cref{tab:runtime_results}, and a comparison of per-step runtimes in \cref{fig:runtimes_1,fig:runtimes_2}. We find that both total and per-step runtimes are comparable to or lower than existing works. To further validate this, we performed additional experiments with extended versions of datasets RotatedMNIST, WM811K, and Camelyon17. For the extended versions, we create 15 source domains by duplicating source domains. (This experiment was solely conducted as a runtime check, not for results-based benchmarking purposes.) The results for the "+Extended" datasets are in \cref{fig:runtimes_extended}. We observe that, also on considerably (+\SI{200}{\percent}) longer domain sequences, the per-step runtimes remain comparable to the baselines.

\begin{table*}[htbp]
\centering
\caption{Runtimes for the methods. Shown: mean$\pm$standard error runtimes in minutes across three trials.}
\label{tab:runtime_results}
\adjustbox{max width=\textwidth}{%
\begin{tabular}{lccccccc}
\toprule
\textbf{Method}                            & \textbf{RotatedMNIST}                      & \textbf{CIFAR10C}                          & \textbf{TinyImageNetC}                     & \textbf{WM811K}                            & \textbf{Covertype}                         & \textbf{Camelyon17}                        & \textbf{Avg}                               \\
\midrule
\rowcolor{color_finetune} Finetune                                        & 53.7 $\pm$ 3.9                                                            & 257.6 $\pm$ 20.2                                                          & 756.9 $\pm$ 8.9                                                           & 1.5 $\pm$ 0.3                                                             & 4.9 $\pm$ 0.1                                                             & \textbf{133.5 $\pm$ 2.3}                                                  & 201.3                                                                     \\
\midrule
\rowcolor{color_optimization} AGEM                                              & 57.2 $\pm$ 4.7                                                            & 406.3 $\pm$ 2.3                                                           & 789.5 $\pm$ 24.2                                                          & \dotuline{1.5 $\pm$ 0.0}                                                  & 4.6 $\pm$ 0.1                                                             & 204.6 $\pm$ 33.7                                                          & 244.0                                                                     \\
\rowcolor{color_optimization} UPGD                                        & \textbf{27.8 $\pm$ 1.2}                                                   & \textbf{14.3 $\pm$ 1.5}                                                   & \dotuline{117.8 $\pm$ 15.2}                                               & 7.8 $\pm$ 0.5                                                             & 0.8 $\pm$ 0.0                                                             & 173.9 $\pm$ 20.9                                                          & \underline{57.1}                                                          \\
\midrule
\rowcolor{color_regularization} EWC                                       & 54.8 $\pm$ 3.3                                                            & 278.5 $\pm$ 25.8                                                          & 1891.5 $\pm$ 760.9                                                        & 2.4 $\pm$ 0.0                                                             & 8.1 $\pm$ 0.1                                                             & 255.0 $\pm$ 89.2                                                          & 415.0                                                                     \\
\rowcolor{color_regularization} SI                                        & 56.2 $\pm$ 5.8                                                            & 309.1 $\pm$ 6.4                                                           & 820.9 $\pm$ 11.4                                                          & 1.7 $\pm$ 0.3                                                             & 5.0 $\pm$ 0.4                                                             & 158.6 $\pm$ 17.8                                                          & 225.2                                                                     \\
\rowcolor{color_regularization} SNR                                       & 44.7 $\pm$ 6.3                                                            & \underline{22.6 $\pm$ 2.3}                                                & \textbf{100.6 $\pm$ 12.8}                                                 & 8.1 $\pm$ 1.3                                                             & \textbf{0.6 $\pm$ 0.0}                                                    & \dotuline{157.9 $\pm$ 18.8}                                               & \textbf{55.8}                                                             \\
\midrule
\rowcolor{color_replay} FDR                                               & 40.2 $\pm$ 2.9                                                            & 256.3 $\pm$ 12.1                                                          & 806.2 $\pm$ 13.0                                                          & 1.6 $\pm$ 0.0                                                             & \dotuline{0.7 $\pm$ 0.0}                                                  & 161.4 $\pm$ 14.6                                                          & 211.1                                                                     \\
\rowcolor{color_replay} ER-ACE                                            & 55.2 $\pm$ 14.2                                                           & 350.5 $\pm$ 21.1                                                          & \underline{115.2 $\pm$ 10.5}                                              & \underline{1.5 $\pm$ 0.0}                                                 & 1.1 $\pm$ 0.0                                                             & 289.8 $\pm$ 77.0                                                          & 135.5                                                                     \\
\rowcolor{color_replay} LODE                                              & \underline{30.6 $\pm$ 1.9}                                                & \dotuline{233.4 $\pm$ 13.3}                                               & 776.7 $\pm$ 17.2                                                          & \textbf{1.4 $\pm$ 0.0}                                                    & \underline{0.6 $\pm$ 0.0}                                                 & 180.7 $\pm$ 17.2                                                          & 203.9                                                                     \\
\rowcolor{color_replay} STAR                                              & 66.7 $\pm$ 17.0                                                           & 627.9 $\pm$ 49.9                                                          & 872.2 $\pm$ 25.0                                                          & 1.9 $\pm$ 0.0                                                             & 0.9 $\pm$ 0.1                                                             & 158.5 $\pm$ 16.2                                                          & 288.0                                                                     \\
\rowcolor{color_replay} COPE                                              & 48.2 $\pm$ 9.3                                                            & 1604.8 $\pm$ 410.3                                                        & 958.4 $\pm$ 100.7                                                         & 4.6 $\pm$ 0.3                                                             & 7.2 $\pm$ 0.2                                                             & 222.2 $\pm$ 50.2                                                          & 474.2                                                                     \\
\rowcolor{color_replay} EFC                                               & \dotuline{39.9 $\pm$ 4.5}                                                 & 301.5 $\pm$ 25.5                                                          & 195.0 $\pm$ 21.5                                                          & 2.5 $\pm$ 0.3                                                             & 0.8 $\pm$ 0.0                                                             & 205.8 $\pm$ 27.4                                                          & \dotuline{124.2}                                                          \\
\rowcolor{color_replay} SARL                                              & 55.8 $\pm$ 15.7                                                           & 367.9 $\pm$ 23.2                                                          & 829.7 $\pm$ 6.4                                                           & 1.7 $\pm$ 0.0                                                             & 0.8 $\pm$ 0.1                                                             & 166.8 $\pm$ 18.7                                                          & 237.1                                                                     \\
\midrule
\rowcolor{color_tailored} $\bigstar$-CL-VREX                              & 77.5 $\pm$ 14.8                                                           & 588.0 $\pm$ 78.3                                                          & 875.3 $\pm$ 12.5                                                          & 1.8 $\pm$ 0.0                                                             & 5.0 $\pm$ 0.2                                                             & 180.6 $\pm$ 15.7                                                          & 288.0                                                                     \\
\rowcolor{color_tailored} $\bigstar$-CL-Fishr                             & 58.3 $\pm$ 1.8                                                            & 423.9 $\pm$ 8.5                                                           & 974.0 $\pm$ 18.0                                                          & 2.2 $\pm$ 0.0                                                             & 5.0 $\pm$ 0.1                                                             & \underline{157.1 $\pm$ 19.2}                                              & 270.1                                                                     \\
\rowcolor{color_tailored} $\bigstar$-CL-CORAL                             & 48.0 $\pm$ 2.3                                                            & 484.3 $\pm$ 95.3                                                          & 794.2 $\pm$ 11.3                                                          & 1.9 $\pm$ 0.3                                                             & 5.4 $\pm$ 0.4                                                             & 318.2 $\pm$ 106.9                                                         & 275.3                                                                     \\
\rowcolor{color_tailored} $\bigstar$-CL-MMD                               & 48.1 $\pm$ 2.9                                                            & 598.3 $\pm$ 9.0                                                           & 875.7 $\pm$ 5.1                                                           & 2.0 $\pm$ 0.3                                                             & 4.7 $\pm$ 0.0                                                             & 196.5 $\pm$ 14.3                                                          & 287.6                                                                     \\
\rowcolor{color_tailored} $\bigstar$-CL-ANDMask                           & 55.3 $\pm$ 4.9                                                            & 294.7 $\pm$ 10.3                                                          & 887.3 $\pm$ 9.5                                                           & 2.2 $\pm$ 0.1                                                             & 4.9 $\pm$ 0.2                                                             & 219.8 $\pm$ 52.2                                                          & 244.0                                                                     \\
\bottomrule
\end{tabular}}
\end{table*}

\newpage
\subsection{Ablation of the invariance alignment}
We here present the results for two experiments: (1) We disable the invariance alignment entirely, by setting $\beta$ to $0$\footnote{\noindent Note that ablating the effect of the invariance penalty itself (i.e., \Cref{eq:replay_penalty_generic}) would simultaneously deactivate the alignment loss, as there would not be any statistics to align. Thus, this would fall back to normal fine-tuning with a replay buffer. The results for this ``non-ablation'' can be found as the ER-ACE method, which essentially uses replay with sequential fine-tuning, in \Cref{tab:main_results}.}. (2) We dynamically recompute the anchor statistics $\priorPrime$ for previous domains at the end of the current domain. The results for the ablations can be found in \cref{tab:abl_no_inv_align,tab:abl_dyn_anchors}. We find that \textit{anchoring to the original (unaltered) representations generally improves the performance.}

\begin{table}[htbp]
\centering
\caption{Results for disabling the sequential invariance alignment. Shown: mean$\pm$standard error target domain accuracy across three independent runs.}
\label{tab:abl_no_inv_align}
\adjustbox{max width=\linewidth}{%
\begin{tabular}{l|lccc}
\toprule
\textbf{Ablation} & \textbf{Method}                            & \textbf{RotatedMNIST}                      & \textbf{Covertype}                         & \textbf{Avg}                               \\
\midrule
 \multirow{5}{*}{W/o alignment} & CL-VREX                                     & 62.4 $\pm$ 1.3                             & 37.1 $\pm$ 0.4                             & 49.7                                       \\
&CL-Fishr                                    & 74.7 $\pm$ 4.5                             & 37.5 $\pm$ 0.8                             & 56.1                                       \\
&CL-CORAL                                    & 69.5 $\pm$ 1.3                             & 37.5 $\pm$ 1.6                             & 53.5                                       \\
&CL-MMD                                      & 66.5 $\pm$ 0.5                             & 37.6 $\pm$ 1.0                             & 52.0                                       \\
&CL-ANDMask                                  & 70.8 $\pm$ 4.1                             & 34.2 $\pm$ 1.0                             & 52.5                                       \\
\midrule
\multirow{5}{*}{With alignment} & CL-VREX                             & 70.2 $\pm$ 5.8                             & 40.8 $\pm$ 1.5                             & 55.5                                       \\
&CL-Fishr                             & 68.6 $\pm$ 0.4                             & 27.0 $\pm$ 7.7                             & 47.8                                       \\
&CL-CORAL                             & 72.8 $\pm$ 2.6                             & 45.2 $\pm$ 3.9                             & 59.0                                       \\
&CL-MMD                                & 70.7 $\pm$ 0.5                             & 37.6 $\pm$ 0.9                             & 54.2                                       \\
&CL-ANDMask                          & 71.4 $\pm$ 2.6                             & 43.7 $\pm$ 2.9                             & 57.6                                       \\

\bottomrule
\end{tabular}}
\end{table}

\begin{table}[htbp]
\begin{center}
\caption{\textbf{Ablating the effect of dynamically recomputing $\priorPrime$}. We recompute $\priorPrime$ for previous domains at the end of the current domain, using the data stored in the replay buffer. For comparison, we give the results for fixing the anchor representations (``Static $\priorPrime$''). We report the mean accuracy (macro F1 for WM811K)$\pm$standard error \emph{target domain performance} across three independent runs.}
\label{tab:abl_dyn_anchors}
\adjustbox{max width=\textwidth}{%
\begin{tabular}{llcccccc}
\toprule
&
\textbf{Method}                                                           & \textbf{RotatedMNIST}                                                     & \textbf{CIFAR10C}                                                         & \textbf{WM811K}                                                           & \textbf{Covertype}                                                        & \textbf{Camelyon17}                                                       & \textbf{Avg}                                                              \\
\midrule
\multirow{4}{*}{Static $\priorPrime$} &
  $\bigstar$-CL-VREX                              & 70.2 $\pm$ 5.8                                                            & 67.2 $\pm$ 1.4                                                            & 84.1 $\pm$ 0.3                                                            & 40.8 $\pm$ 1.5                                                            & 91.5 $\pm$ 0.6                                                            & 70.8                                                                      \\
& $\bigstar$-CL-Fishr                             & 68.6 $\pm$ 0.4                                                            & 64.2 $\pm$ 1.2                                                            & 83.6 $\pm$ 0.5                                                            & 27.0 $\pm$ 7.7                                                            & 89.7 $\pm$ 2.0                                                            & 66.6                                                                      \\
& $\bigstar$-CL-CORAL                             & 72.8 $\pm$ 2.6                                                            & 68.5 $\pm$ 1.7                                                            & 84.8 $\pm$ 0.6                                                            & 45.2 $\pm$ 3.9                                                            & 91.7 $\pm$ 0.4                                                            & 72.6                                                                      \\
& $\bigstar$-CL-MMD                               & 70.7 $\pm$ 0.5                                                            & 69.0 $\pm$ 0.5                                                            & 85.5 $\pm$ 0.4                                                            & 37.6 $\pm$ 0.9                                                            & 90.1 $\pm$ 0.5                                                            & 70.6                                                                      \\
\midrule
\multirow{4}{*}{Dynamic $\priorPrime$} &  $\bigstar$-CL-VREX                              & 71.8 $\pm$ 6.0                                                            & 65.6 $\pm$ 1.3                                                            & 74.0 $\pm$ 1.7                                                            & 36.5 $\pm$ 0.6                                                            & 90.7 $\pm$ 0.3                                                                         & 67.7                                                                         \\
&  $\bigstar$-CL-Fishr                             & 67.9 $\pm$ 1.1                                                            & 22.9 $\pm$ 4.0                                                            & 76.3 $\pm$ 0.7                                                            & 29.9 $\pm$ 6.0                                                            & 89.9 $\pm$ 1.5                                                                        & 57.4                                                                         \\
& $\bigstar$-CL-CORAL                             & 74.7 $\pm$ 1.0                                                            & 66.3 $\pm$ 1.3                                                            & 73.6 $\pm$ 1.5                                                            & 33.9 $\pm$ 0.6                                                            & 88.8 $\pm$ 0.0                                                            & 67.5                                                                      \\
& $\bigstar$-CL-MMD                               & 68.3 $\pm$ 0.4                                                            & 66.9 $\pm$ 0.5                                                            & 71.6 $\pm$ 1.9                                                            & 38.6 $\pm$ 1.3                                                            & 89.4 $\pm$ 1.0                                                            & 67.0                                                                      \\
\bottomrule
\end{tabular}}
\end{center}
\end{table}

\newpage
\subsection{Robustness check: reduced buffer sizes}
We here present the full results for our robustness experiments that reduce the buffer capacity to \SI{50}{\percent} (in \Cref{tab:robustness_50p_buffer}) and \SI{25}{\percent} (in \Cref{tab:robustness_25p_buffer}).

\begin{table*}[h]
\caption{Results for a \SI{50}{\percent} reduced buffer size. Shown: mean$\pm$standard error target domain accuracy across three independent runs. Marked: \textbf{best}, \underline{second}, \dotuline{third}.}
\label{tab:robustness_50p_buffer}
\adjustbox{max width=\linewidth}{%
\begin{tabular}{l|cccc|ccc}
\toprule
& \multicolumn{4}{c|}{Accuracy ($\uparrow$)} & \multicolumn{3}{c}{Rank ($\downarrow$)} \\
\midrule
\textbf{Method}                            & \textbf{RotatedMNIST}                      & \textbf{Covertype}                         & \textbf{CIFAR10C}                          & \textbf{Avg}                               & \textbf{Arith. mean}                       & \textbf{Geom. mean}                        & \textbf{Median}                            \\
\midrule
\rowcolor{color_replay} FDR                                               & 57.8 $\pm$ 0.6                                                            & \underline{65.8 $\pm$ 0.1}                                                & 37.7 $\pm$ 0.9                                                            & 53.8                                                                      & \underline{4.0}                                                           & \dotuline{3.6}                                                            & \dotuline{4.0}                                                            \\
\rowcolor{color_replay} ER-ACE                                            & 56.1 $\pm$ 2.0                                                            & 64.4 $\pm$ 0.8                                                            & \dotuline{38.1 $\pm$ 0.6}                                                 & 52.8                                                                      & 5.7                                                                       & 5.3                                                                       & 7.0                                                                       \\
\rowcolor{color_replay} STAR                                              & 55.6 $\pm$ 2.5                                                            & \dotuline{65.6 $\pm$ 0.9}                                                 & \textbf{39.0 $\pm$ 1.1}                                                   & 53.4                                                                      & \underline{4.0}                                                           & \underline{2.9}                                                           & \underline{3.0}                                                           \\
\midrule
\rowcolor{color_tailored} $\bigstar$-CL-VREX                              & \textbf{68.1 $\pm$ 1.0}                                                   & \textbf{66.2 $\pm$ 1.0}                                                   & \underline{38.4 $\pm$ 1.3}                                                & \textbf{57.5}                                                             & \textbf{1.3}                                                              & \textbf{1.3}                                                              & \textbf{1.0}                                                              \\
\rowcolor{color_tailored} $\bigstar$-CL-Fishr                             & \dotuline{67.5 $\pm$ 0.7}                                                 & 64.4 $\pm$ 0.8                                                            & 29.2 $\pm$ 7.6                                                            & 53.7                                                                      & 5.7                                                                       & 5.2                                                                       & 6.0                                                                       \\
\rowcolor{color_tailored} $\bigstar$-CL-CORAL                             & 66.0 $\pm$ 3.5                                                            & 65.3 $\pm$ 1.2                                                            & 37.3 $\pm$ 0.7                                                            & \underline{56.2}                                                          & \dotuline{4.7}                                                            & 4.6                                                                       & \dotuline{4.0}                                                            \\
\rowcolor{color_tailored} $\bigstar$-CL-MMD                               & 64.9 $\pm$ 7.6                                                            & 65.3 $\pm$ 0.5                                                            & 37.5 $\pm$ 0.7                                                            & \dotuline{55.9}                                                           & 5.0                                                                       & 5.0                                                                       & 5.0                                                                       \\
\rowcolor{color_tailored} $\bigstar$-CL-ANDMask                           & \underline{67.5 $\pm$ 1.2}                                                & 62.0 $\pm$ 0.6                                                            & 36.0 $\pm$ 1.3                                                            & 55.2                                                                      & 5.7                                                                       & 4.8                                                                       & 7.0                                                                       \\
\bottomrule
\end{tabular}}
\end{table*}

\begin{table*}[h]
\caption{Results for a \SI{75}{\percent} reduced buffer size. Shown: mean$\pm$standard error target domain accuracy across three independent runs. Marked: \textbf{best}, \underline{second}, \dotuline{third}.}
\label{tab:robustness_25p_buffer}
\adjustbox{max width=\linewidth}{%
\begin{tabular}{l|cccc|ccc}
\toprule
& \multicolumn{4}{c|}{Accuracy ($\uparrow$)} & \multicolumn{3}{c}{Rank ($\downarrow$)} \\
\textbf{Method}                            & \textbf{RotatedMNIST}                      & \textbf{Covertype}                         & \textbf{CIFAR10C}                          & \textbf{Avg}                               & \textbf{Arith. mean}                       & \textbf{Geom. mean}                        & \textbf{Median}                            \\
\midrule
\rowcolor{color_replay} FDR                                               & 62.5 $\pm$ 2.7                                                            & 64.5 $\pm$ 0.6                                                            & 37.3 $\pm$ 0.5                                                            & \dotuline{54.8}                                                           & 5.3                                                                       & 5.3                                                                       & 5.0                                                                       \\
\rowcolor{color_replay} ER-ACE                                            & 54.0 $\pm$ 2.1                                                            & 63.7 $\pm$ 0.5                                                            & \dotuline{37.6 $\pm$ 0.4}                                                 & 51.8                                                                      & 5.3                                                                       & 5.0                                                                       & 6.0                                                                       \\
\rowcolor{color_replay} STAR                                              & 52.9 $\pm$ 2.6                                                            & \underline{66.6 $\pm$ 0.4}                                                & \underline{37.7 $\pm$ 0.7}                                                & 52.4                                                                      & \dotuline{4.0}                                                            & \underline{3.2}                                                           & \underline{2.0}                                                           \\
\midrule
\rowcolor{color_tailored} $\bigstar$-CL-VREX                              & \textbf{67.1 $\pm$ 8.1}                                                   & \textbf{67.0 $\pm$ 0.6}                                                   & \textbf{41.1 $\pm$ 2.8}                                                   & \textbf{58.4}                                                             & \textbf{1.0}                                                              & \textbf{1.0}                                                              & \textbf{1.0}                                                              \\
\rowcolor{color_tailored} $\bigstar$-CL-Fishr                             & 63.9 $\pm$ 1.9                                                            & 63.3 $\pm$ 1.1                                                            & 28.7 $\pm$ 6.9                                                            & 52.0                                                                      & 6.3                                                                       & 6.1                                                                       & 7.0                                                                       \\
\rowcolor{color_tailored} $\bigstar$-CL-CORAL                             & \underline{65.5 $\pm$ 2.7}                                                & 65.0 $\pm$ 1.6                                                            & 37.5 $\pm$ 2.1                                                            & \underline{56.0}                                                          & \underline{3.3}                                                           & \underline{3.2}                                                           & \dotuline{4.0}                                                            \\
\rowcolor{color_tailored} $\bigstar$-CL-MMD                               & 55.5 $\pm$ 3.0                                                            & \dotuline{65.1 $\pm$ 1.0}                                                 & 37.4 $\pm$ 0.9                                                            & 52.7                                                                      & 4.7                                                                       & \dotuline{4.5}                                                            & 5.0                                                                       \\
\rowcolor{color_tailored} $\bigstar$-CL-ANDMask                           & \dotuline{64.5 $\pm$ 1.4}                                                 & 60.6 $\pm$ 1.0                                                            & 35.8 $\pm$ 1.3                                                            & 53.7                                                                      & 6.0                                                                       & 5.5                                                                       & 7.0                                                                       \\
\bottomrule
\end{tabular}}
\end{table*}

\newpage
\subsection{Source-domain performance}\label{sec:source_performance}
In this section, we discuss how we report the performance on the source domain sequence. Recall that, in continual learning, a model is sequentially trained on a sequence of source (training) domains. Commonly, the models are evaluated w.r.t. their ability to maintain or even retrospectively improve the performance across these training domains. For this evaluation, the CL literature utilizes a hold-out subset of the training distribution. This is related to \emph{in-domain generalization}, which evaluates the performance of a model on an IID set of the training distribution \citep[c.f.][]{hardt2016train}.

In contrast, in our paper we are interested in the post-training performance, that is, the ability of continually trained models to generalize to an unseen target domain. This is relevant in field such as manufacturing or medicine, where storage or privacy concerns disallow retaining or sharing the training data, yet models need to generalize to new products (in manufacturing) or hospitals (in medicine) after deployment. As such, after continual training on the source-domain sequence, we primarily evaluate the performance on an unseen target domain. Due to this setting, our reporting of the source performance is thus slightly different to the CL literature. In the CL literature, the best runs are selected based on validation splits of the training distributions, and metrics are reported for the test splits of the selected runs (ie, in-domain generalization). In our setting, we equally select the best runs based on the validation splits (see \cref{sec:hparam_details}), but differently report metrics for a novel distribution (out-of-domain generalization). As such, no test split of the training distributions is required for our setting, and hence not available

To report the performance on the source domains, we thus \textbf{have three options:} \textbf{Option~1}: select the best runs as used throughout the paper (as described in \Cref{sec:hparam_details}) and report the validation-split source performance. This directly gives the metrics used to select the best runs w.r.t. target domain performance, i.e., the runs reported in the tables throughout this paper. However, this is basically optimization on the test set \citep[cf.][]{doi:10.1126/science.aaa9375,shankar2021image}. \textbf{Option~2}: select the best runs as in Option 1, but report the training-split source performance. \textbf{Option~3}: select the best runs based on the training-splits of the source domains, and report the validation-split performance. However, the selected runs are then not related to the runs reported in this paper

Given these three options, we can directly discard the last one as it essentially studies a distinct setting. From the remaining options, Option 2 is the most sensible one, as it avoids reporting results that are optimized on the test set. Additionally, Option 2 reports the performance on the data that was actually used during continually training on the source domains. The results are in \cref{tab:source_domain_performance}.

\begin{table}[hbtp]
\centering
\caption{Source domain performance; computed as described in \Cref{sec:source_performance}. Shown: mean$\pm$standard error source domain accuracy (\emph{macro F1 for WM811K}) across three independent runs.}
\label{tab:source_domain_performance}
\adjustbox{max width=\textwidth}{%
\begin{tabular}{lccccccc}
\toprule
\textbf{Method}                            & \textbf{RotatedMNIST}                      & \textbf{CIFAR10C}                          & \textbf{TinyImageNetC}                     & \textbf{WM811K}                            & \textbf{Covertype}                         & \textbf{Camelyon17}                        & \textbf{Avg}                               \\
\midrule
\rowcolor{color_finetune} Finetune                                        & 96.5 $\pm$ 0.2                                                            & 84.5 $\pm$ 1.0                                                            & 67.2 $\pm$ 0.9                                                            & 91.6 $\pm$ 0.3                                                            & 64.3 $\pm$ 0.5                                                            & 92.4 $\pm$ 0.7                                                            & 82.7                                                                      \\
\midrule
\rowcolor{color_optimization} AGEM                                              & 97.7 $\pm$ 0.3                                                            & 90.1 $\pm$ 1.2                                                            & 72.3 $\pm$ 0.9                                                            & 91.6 $\pm$ 0.7                                                            & 67.3 $\pm$ 0.3                                                            & 94.7 $\pm$ 0.2                                                            & 85.6                                                                      \\
\rowcolor{color_optimization} UPGD                                        & 85.0 $\pm$ 2.3                                                            & 63.7 $\pm$ 6.4                                                            & 46.8 $\pm$ 8.0                                                            & 75.0 $\pm$ 1.6                                                            & 54.1 $\pm$ 3.5                                                            & 91.4 $\pm$ 0.6                                                            & 69.3                                                                      \\
\midrule
\rowcolor{color_regularization} EWC                                       & 96.7 $\pm$ 0.3                                                            & 86.2 $\pm$ 0.7                                                            & 67.5 $\pm$ 1.5                                                            & 92.2 $\pm$ 0.2                                                            & 64.6 $\pm$ 1.4                                                            & 92.6 $\pm$ 0.2                                                            & 83.3                                                                      \\
\rowcolor{color_regularization} SI                                        & 96.1 $\pm$ 0.4                                                            & 82.4 $\pm$ 0.9                                                            & 65.9 $\pm$ 1.2                                                            & 91.4 $\pm$ 1.7                                                            & 64.5 $\pm$ 0.7                                                            & 92.8 $\pm$ 0.6                                                            & 82.2                                                                      \\
\rowcolor{color_regularization} SNR                                       & 96.5 $\pm$ 0.4                                                            & 82.0 $\pm$ 0.3                                                            & 66.1 $\pm$ 0.6                                                            & 77.7 $\pm$ 1.0                                                            & 62.4 $\pm$ 1.6                                                            & 91.7 $\pm$ 0.3                                                            & 79.4                                                                      \\
\midrule
\rowcolor{color_replay} FDR                                               & 98.9 $\pm$ 0.1                                                            & 88.9 $\pm$ 0.5                                                            & 83.2 $\pm$ 0.3                                                            & 92.2 $\pm$ 0.4                                                            & 80.0 $\pm$ 0.6                                                            & 94.8 $\pm$ 0.0                                                            & 89.7                                                                      \\
\rowcolor{color_replay} ER-ACE                                            & 99.1 $\pm$ 0.0                                                            & 91.1 $\pm$ 1.1                                                            & 87.3 $\pm$ 1.7                                                            & 93.8 $\pm$ 0.4                                                            & 81.5 $\pm$ 0.8                                                            & 95.4 $\pm$ 0.2                                                            & 91.4                                                                      \\
\rowcolor{color_replay} LODE                                              & 96.8 $\pm$ 0.2                                                            & 88.0 $\pm$ 0.5                                                            & 64.1 $\pm$ 3.1                                                            & 86.3 $\pm$ 1.3                                                            & 67.9 $\pm$ 1.9                                                            & 70.1 $\pm$ 7.7                                                            & 78.9                                                                      \\
\rowcolor{color_replay} STAR                                              & 98.8 $\pm$ 0.0                                                            & 93.1 $\pm$ 0.2                                                            & 83.5 $\pm$ 0.1                                                            & 94.4 $\pm$ 0.5                                                            & 80.2 $\pm$ 0.4                                                            & 95.4 $\pm$ 0.0                                                            & 90.9                                                                      \\
\rowcolor{color_replay} COPE                                              & 97.7 $\pm$ 0.1                                                            & 88.6 $\pm$ 1.4                                                            & 45.8 $\pm$ 10.4                                                           & 91.0 $\pm$ 1.1                                                            & 2.0 $\pm$ 1.0                                                             & 94.6 $\pm$ 0.2                                                            & 69.9                                                                      \\
\rowcolor{color_replay} EFC                                               & 98.5 $\pm$ 0.2                                                            & 89.8 $\pm$ 1.2                                                            & 73.9 $\pm$ 4.2                                                            & 93.6 $\pm$ 1.3                                                            & 66.3 $\pm$ 2.3                                                            & 94.7 $\pm$ 0.2                                                            & 86.1                                                                      \\
\rowcolor{color_replay} SARL                                              & 98.8 $\pm$ 0.1                                                            & 84.5 $\pm$ 3.1                                                            & 0.7 $\pm$ 0.1                                                             & 82.1 $\pm$ 1.6                                                            & 75.4 $\pm$ 0.5                                                            & 84.7 $\pm$ 0.6                                                            & 71.0                                                                      \\
\midrule
\rowcolor{color_tailored} $\bigstar$-CL-VREX                              & 98.6 $\pm$ 0.1                                                            & 89.7 $\pm$ 0.5                                                            & 83.5 $\pm$ 0.8                                                            & 94.7 $\pm$ 0.5                                                            & 72.9 $\pm$ 1.1                                                            & 95.3 $\pm$ 0.2                                                            & 89.1                                                                      \\
\rowcolor{color_tailored} $\bigstar$-CL-Fishr                             & 99.3 $\pm$ 0.0                                                            & 81.3 $\pm$ 1.3                                                            & 85.9 $\pm$ 2.6                                                            & 96.4 $\pm$ 0.1                                                            & 63.1 $\pm$ 1.3                                                            & 94.9 $\pm$ 0.1                                                            & 86.8                                                                      \\
\rowcolor{color_tailored} $\bigstar$-CL-CORAL                             & 99.1 $\pm$ 0.0                                                            & 91.2 $\pm$ 1.0                                                            & 85.7 $\pm$ 1.9                                                            & 93.6 $\pm$ 1.0                                                            & 63.2 $\pm$ 0.7                                                            & 95.4 $\pm$ 0.1                                                            & 88.0                                                                      \\
\rowcolor{color_tailored} $\bigstar$-CL-MMD                               & 99.0 $\pm$ 0.1                                                            & 91.1 $\pm$ 0.5                                                            & 83.7 $\pm$ 0.5                                                            & 93.8 $\pm$ 1.0                                                            & 73.5 $\pm$ 0.7                                                            & 95.6 $\pm$ 0.0                                                            & 89.5                                                                      \\
\rowcolor{color_tailored} $\bigstar$-CL-ANDMask                           & 99.0 $\pm$ 0.1                                                            & 88.9 $\pm$ 0.4                                                            & 36.3 $\pm$ 4.1                                                            & 93.8 $\pm$ 0.2                                                            & 70.0 $\pm$ 1.7                                                            & 94.4 $\pm$ 0.2                                                            & 80.4                                                                      \\
\bottomrule
\end{tabular}}
\end{table}

\newpage
\subsection{Robustness check: different target domains}\label{sec:robustness_different_target_domain}
For our main experiments, we fix the target domain per dataset as indicated in \Cref{sec:dataset_details}. For this robustness experiment, we change the target domain. For RotatedMNIST, we select $D^0$; for CIFAR10C, we select $D^{\mathrm{brightness+contrast}}$; and for Covertype we select $D^{\mathrm{Rawah}}$. The new target domains are thus all from the former set of source domains; hence, a new hyperparameter study is required. The results are given in \cref{tab:robustness_different_target_domain}.

\begin{table*}[!htbp]
\centering
\caption{Results for different target domains. Shown: mean$\pm$standard error target domain accuracy across three independent runs. Marked: \textbf{best}, \underline{second}, \dotuline{third}.}
\label{tab:robustness_different_target_domain}
\adjustbox{max width=\textwidth}{%
\begin{tabular}{lcccc}
\toprule
\textbf{Method}                            & \textbf{RotatedMNIST}                      & \textbf{CIFAR10C}                          & \textbf{Covertype}                         & \textbf{Avg}                               \\
\midrule
\rowcolor{color_finetune} Finetune                                        & 54.8 $\pm$ 1.1                                                            & 49.7 $\pm$ 1.6                                                            & \textbf{49.2 $\pm$ 2.8}                                                   & 51.2                                                                      \\
\midrule
\rowcolor{color_optimization} AGEM                                        & 65.5 $\pm$ 2.3                                                            & 49.8 $\pm$ 0.3                                                            & 35.5 $\pm$ 1.4                                                            & 50.3                                                                      \\
\rowcolor{color_optimization} UPGD                                        & 34.2 $\pm$ 1.9                                                            & 39.4 $\pm$ 0.5                                                            & \underline{46.6 $\pm$ 4.1}                                                & 40.1                                                                      \\
\midrule
\rowcolor{color_regularization} EWC                                       & 60.4 $\pm$ 2.4                                                            & 49.6 $\pm$ 1.5                                                            & 40.4 $\pm$ 1.9                                                            & 50.1                                                                      \\
\rowcolor{color_regularization} SI                                        & 53.4 $\pm$ 1.9                                                            & 50.0 $\pm$ 1.0                                                            & 35.4 $\pm$ 1.6                                                            & 46.3                                                                      \\
\midrule
\rowcolor{color_replay} FDR                                               & 86.0 $\pm$ 0.5                                                            & \textbf{52.5 $\pm$ 1.2}                                                   & 39.3 $\pm$ 1.7                                                            & \underline{59.3}                                                          \\
\rowcolor{color_replay} ER-ACE                                            & \dotuline{88.4 $\pm$ 3.3}                                                 & 45.7 $\pm$ 1.5                                                            & 40.2 $\pm$ 3.3                                                            & 58.1                                                                      \\
\rowcolor{color_replay} STAR                                              & 84.6 $\pm$ 0.3                                                            & \underline{51.7 $\pm$ 0.4}                                                & 39.4 $\pm$ 1.1                                                            & 58.6                                                                      \\
\midrule
\rowcolor{color_tailored} $\bigstar$-CL-VREX                              & 85.9 $\pm$ 1.6                                                            & \dotuline{50.4 $\pm$ 1.2}                                                 & 38.3 $\pm$ 1.9                                                            & 58.2                                                                      \\
\rowcolor{color_tailored} $\bigstar$-CL-Fishr                             & \textbf{91.8 $\pm$ 0.4}                                                   & 49.6 $\pm$ 1.8                                                            & 35.7 $\pm$ 3.0                                                            & \dotuline{59.1}                                                           \\
\rowcolor{color_tailored} $\bigstar$-CL-CORAL                             & 87.1 $\pm$ 0.8                                                            & 48.2 $\pm$ 1.0                                                            & 38.5 $\pm$ 1.4                                                            & 57.9                                                                      \\
\rowcolor{color_tailored} $\bigstar$-CL-MMD                               & 85.7 $\pm$ 1.2                                                            & 49.0 $\pm$ 0.9                                                            & \dotuline{42.4 $\pm$ 1.0}                                                 & 59.0                                                                      \\
\rowcolor{color_tailored} $\bigstar$-CL-ANDMask                           & \underline{89.6 $\pm$ 0.6}                                                & 49.6 $\pm$ 0.7                                                            & 40.4 $\pm$ 1.1                                                            & \textbf{59.8}                                                             \\
\bottomrule
\end{tabular}}
\end{table*}

\subsection{Robustness: different compute limit}
For our main experiments, we fix the compute budget, which is the number of training steps per domain. We here set the compute budget to \SI{50}{\percent} of the original value. Result are in \cref{tab:different_compute_budget}.

\begin{table*}[h]
\centering
\caption{Results for a reduced compute budget. Shown: mean$\pm$standard error target domain accuracy across three independent runs. Marked: \textbf{best}, \underline{second}, \dotuline{third}.}
\label{tab:different_compute_budget}
\adjustbox{max width=\textwidth}{%
\begin{tabular}{lcccc}
\toprule
\textbf{Method}                            & \textbf{RotatedMNIST}                      & \textbf{CIFAR10C}                          & \textbf{Covertype}                         & \textbf{Avg}                               \\
\midrule
\rowcolor{color_finetune} Finetune                                        & 39.4 $\pm$ 1.7                                                            & 58.4 $\pm$ 1.5                                                            & 27.9 $\pm$ 5.8                                                            & 41.9                                                                      \\
\midrule
\rowcolor{color_optimization} AGEM                                        & 48.7 $\pm$ 4.7                                                            & 64.1 $\pm$ 0.8                                                            & 36.9 $\pm$ 0.5                                                            & 49.9                                                                      \\
\rowcolor{color_optimization} UPGD                                        & 26.6 $\pm$ 6.3                                                            & 55.0 $\pm$ 5.6                                                            & 18.6 $\pm$ 8.0                                                            & 33.4                                                                      \\
\midrule
\rowcolor{color_regularization} EWC                                       & 33.8 $\pm$ 0.4                                                            & 61.3 $\pm$ 0.8                                                            & 29.8 $\pm$ 5.9                                                            & 41.6                                                                      \\
\rowcolor{color_regularization} SI                                        & 36.7 $\pm$ 0.8                                                            & 60.4 $\pm$ 0.5                                                            & \textbf{41.8 $\pm$ 3.2}                                                   & 46.3                                                                      \\
\midrule
\rowcolor{color_replay} FDR                                               & \dotuline{68.4 $\pm$ 0.9}                                                 & 64.2 $\pm$ 0.8                                                            & \underline{38.9 $\pm$ 0.1}                                                & \dotuline{57.2}                                                           \\
\rowcolor{color_replay} ER-ACE                                            & 63.8 $\pm$ 3.1                                                            & 63.8 $\pm$ 0.9                                                            & 37.3 $\pm$ 0.2                                                            & 55.0                                                                      \\
\rowcolor{color_replay} STAR                                              & 67.2 $\pm$ 0.6                                                            & \underline{65.0 $\pm$ 0.7}                                                & 37.8 $\pm$ 0.6                                                            & 56.7                                                                      \\
\midrule
\rowcolor{color_tailored} $\bigstar$-CL-VREX                              & \underline{69.4 $\pm$ 5.2}                                                & \dotuline{64.5 $\pm$ 0.9}                                                 & \dotuline{37.9 $\pm$ 0.7}                                                 & \underline{57.2}                                                          \\
\rowcolor{color_tailored} $\bigstar$-CL-Fishr                             & 64.0 $\pm$ 2.0                                                            & 60.1 $\pm$ 2.9                                                            & 34.8 $\pm$ 1.4                                                            & 53.0                                                                      \\
\rowcolor{color_tailored} $\bigstar$-CL-CORAL                             & \textbf{73.1 $\pm$ 3.3}                                                   & \textbf{66.0 $\pm$ 1.2}                                                   & 36.4 $\pm$ 1.1                                                            & \textbf{58.5}                                                             \\
\rowcolor{color_tailored} $\bigstar$-CL-MMD                               & 67.4 $\pm$ 2.3                                                            & 64.2 $\pm$ 0.7                                                            & 36.7 $\pm$ 0.4                                                            & 56.1                                                                      \\
\rowcolor{color_tailored} $\bigstar$-CL-ANDMask                           & 66.4 $\pm$ 0.2                                                            & 63.9 $\pm$ 0.5                                                            & 33.9 $\pm$ 2.9                                                            & 54.7                                                                      \\
\bottomrule
\end{tabular}}
\end{table*}

\newpage
\subsection{Visualizations: Hyperparameter sensitivity}
\begin{figure}[!htbp]
\centering
    \begin{subfigure}{.40\linewidth}
        \centering
        \includegraphics[width=\linewidth]{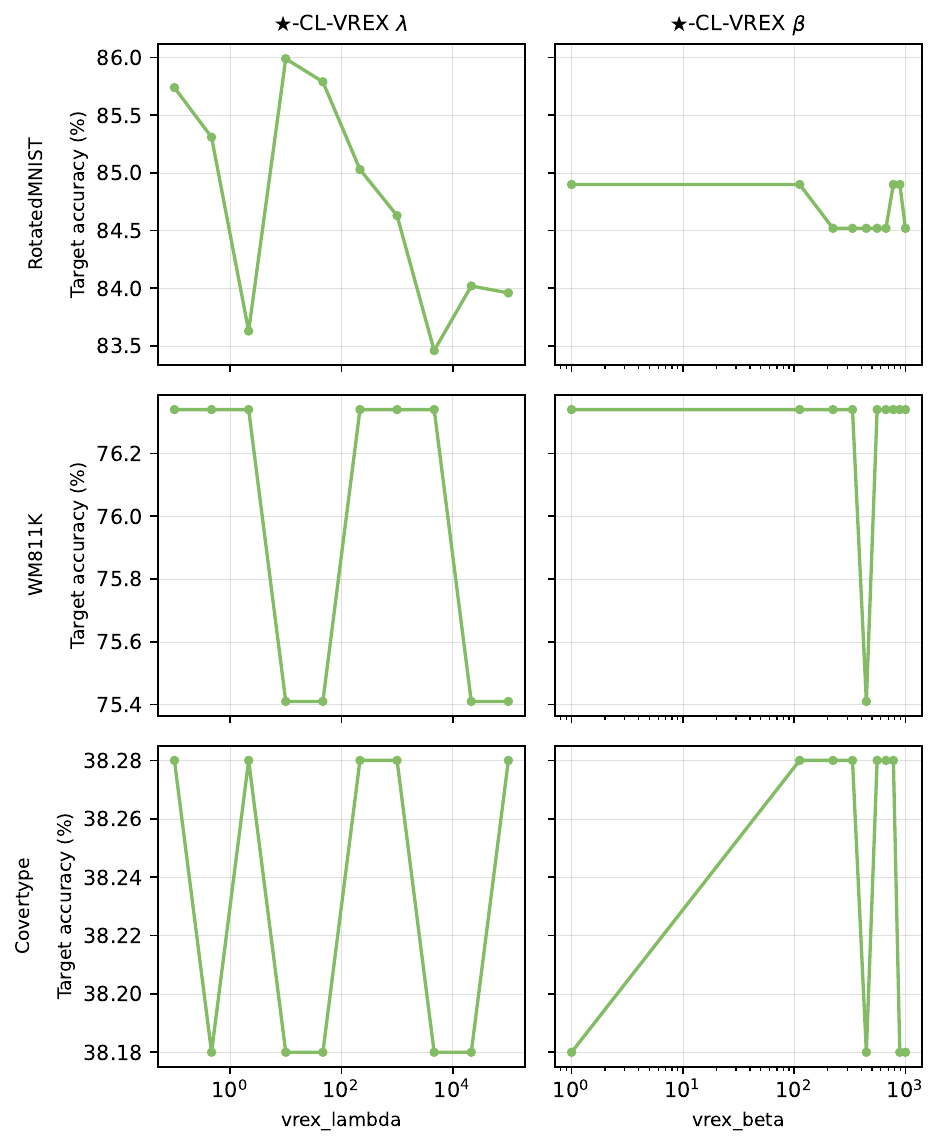}
    \end{subfigure}%
    \begin{subfigure}{.40\linewidth}
        \centering
        \includegraphics[width=\linewidth]{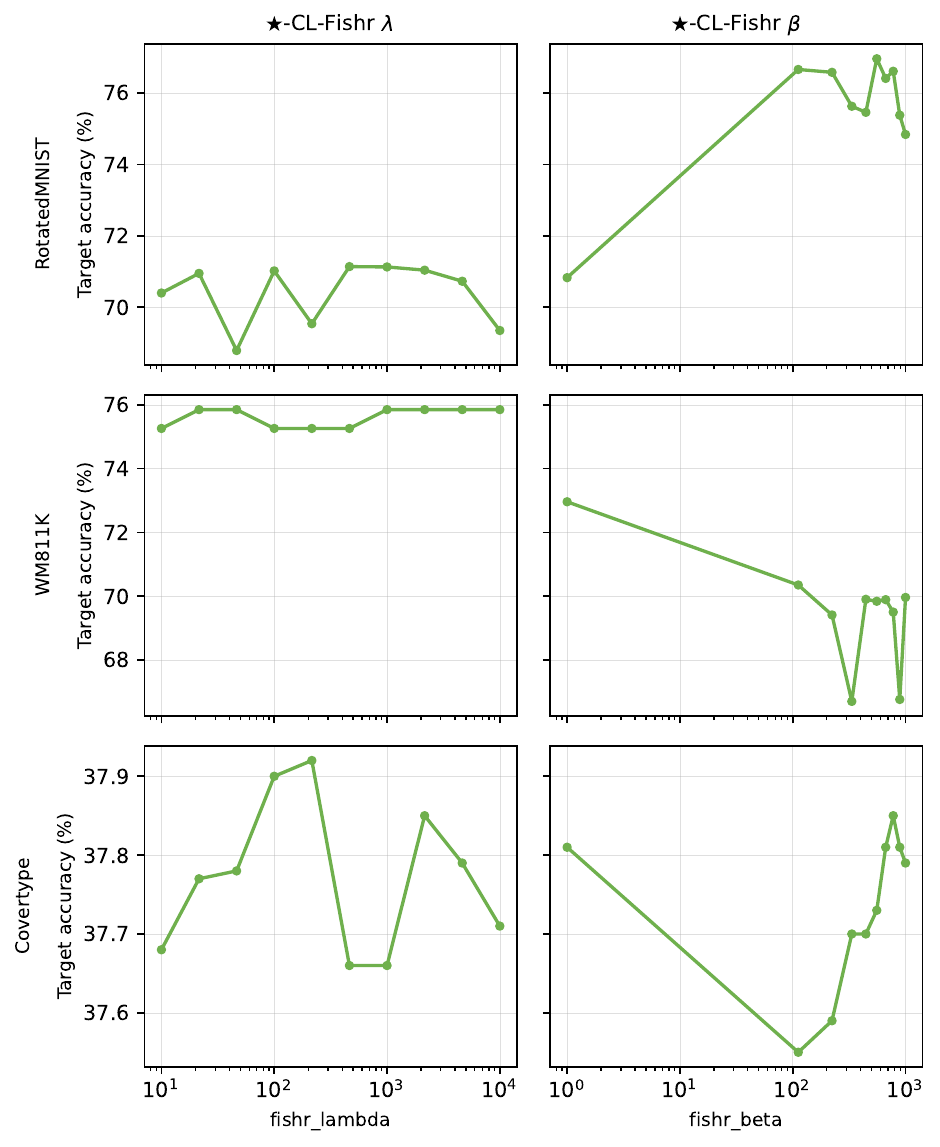}
    \end{subfigure}
    \vspace{-5pt}
    \caption{
    {
    \setlength{\fboxsep}{0pt}
    \textbf{Hyperparameter sensitivity plots (I).}
    We ablate the effect of varying $\lambda$ and $\beta$ parameters for $\bigstar$-CL-VREX (left) and $\bigstar$-CL-Fishr (right).
    }}
    \label{fig:sensitivity_1}
    \vspace{-7pt}
\end{figure}

\begin{figure}[!htbp]
\centering
    \begin{subfigure}{.4\linewidth}
        \centering
        \includegraphics[width=\linewidth]{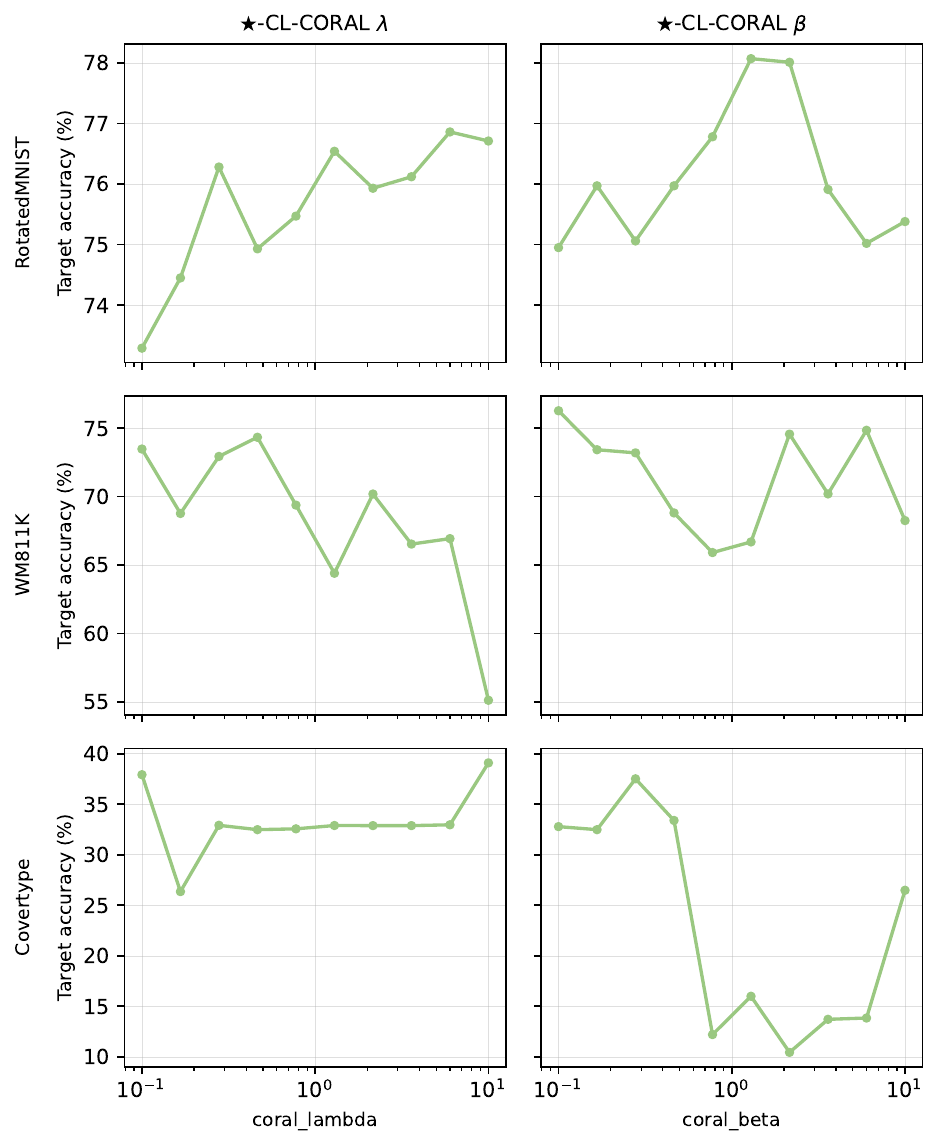}
    \end{subfigure}%
    \begin{subfigure}{.4\linewidth}
        \centering
        \includegraphics[width=\linewidth]{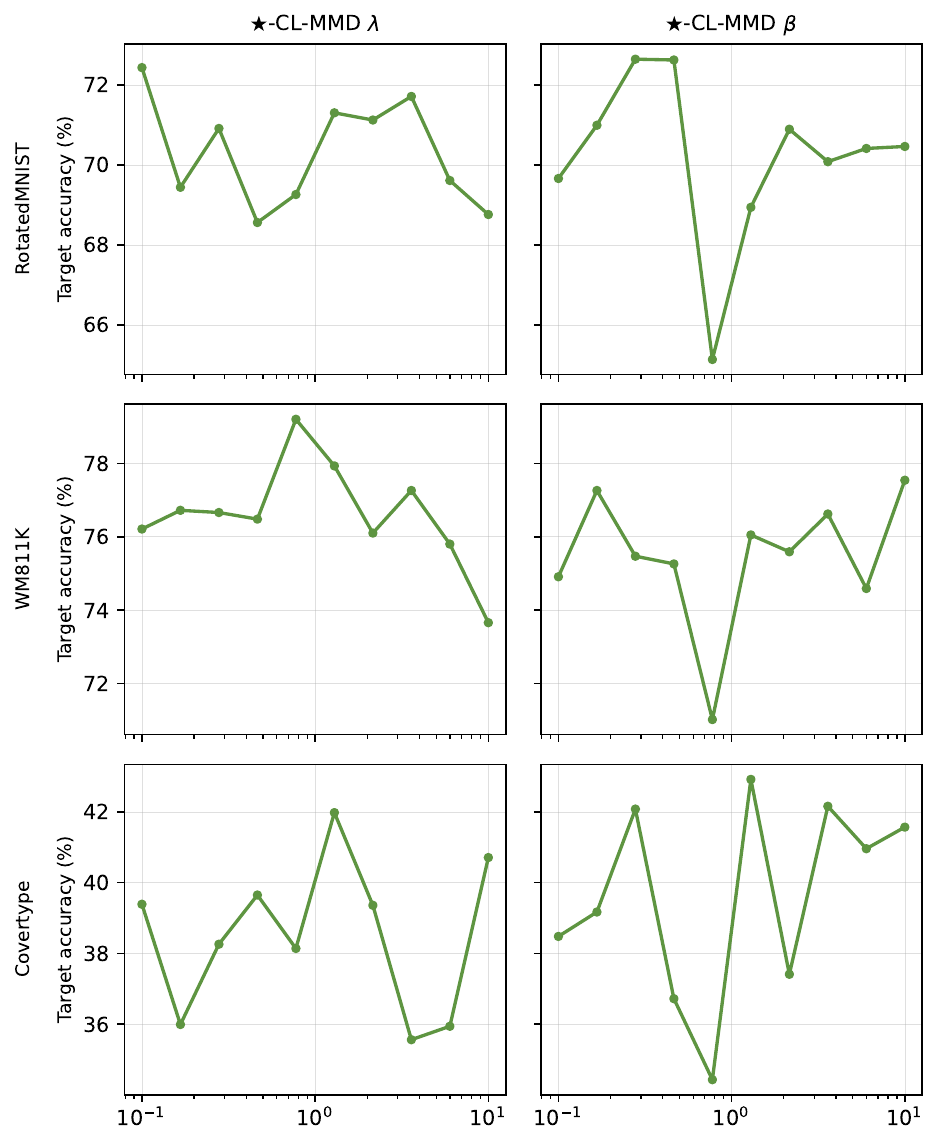}
    \end{subfigure}
    \vspace{-5pt}
    \caption{
    {
    \setlength{\fboxsep}{0pt}
    \textbf{Hyperparameter sensitivity plots (II).}
    We ablate the effect of varying $\lambda$ and $\beta$ parameters for $\bigstar$-CL-CORAL (left) and $\bigstar$-CL-MMD (right).
    }}
    \label{fig:sensitivity_2}
    \vspace{-7pt}
\end{figure}

We here analyze the sensitivity of our proposed methods to changes in the $\lambda$ and $\beta$ parameters. For this, we select the best run of the three runs, and vary $\lambda,\beta$ across 10 different values picked from their value range in \Cref{tab:hyperparameter_details}. We perform this analysis on three different datasets, namely, RotatedMNIST, WM811K, Covertype.
The sensitivity plots are in \cref{fig:sensitivity_1,fig:sensitivity_2}. We find that our proposed $\bigstar$-CL methods are relatively robust, with changes often being less than 2pp.

\newpage
\subsection{Visualizations: feature spaces}
\begin{figure}[!htbp]
\centering
    \begin{subfigure}{.3\linewidth}
        \centering
        \includegraphics[width=\linewidth]{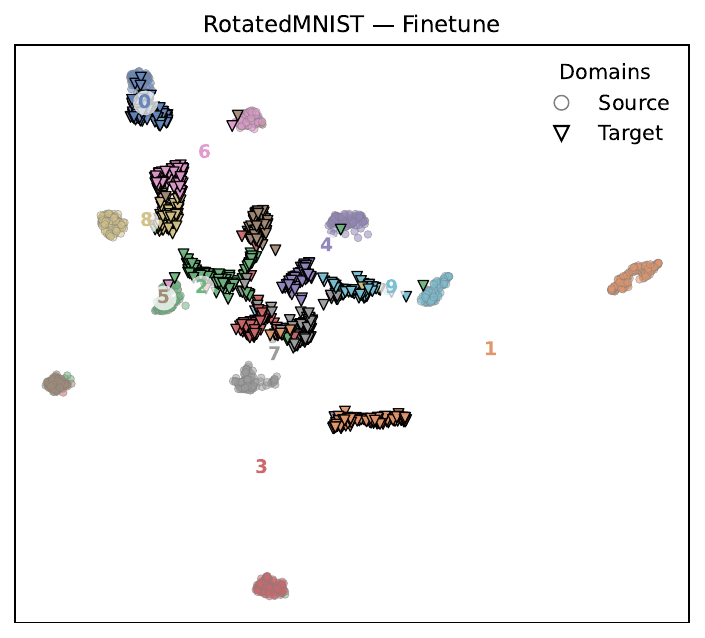}
    \end{subfigure}%
    \begin{subfigure}{.3\linewidth}
        \centering
        \includegraphics[width=\linewidth]{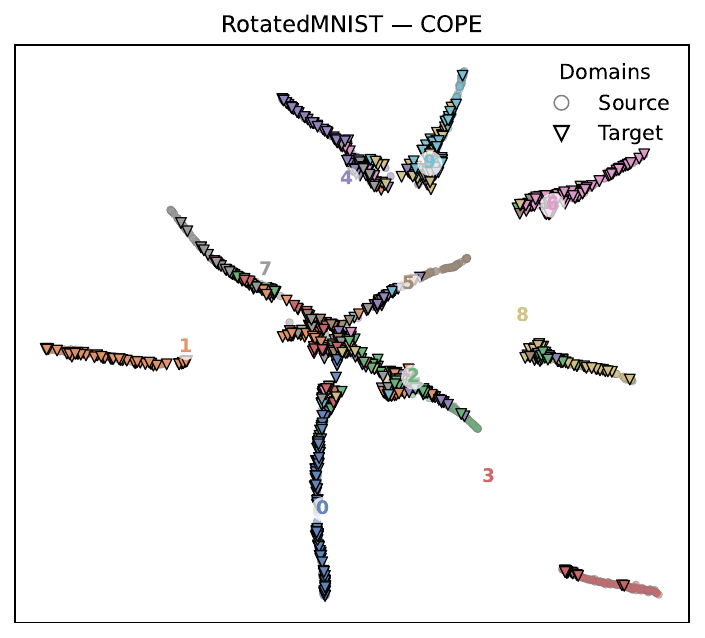}
    \end{subfigure}%
    \begin{subfigure}{.3\linewidth}
        \centering
        \includegraphics[width=\linewidth]{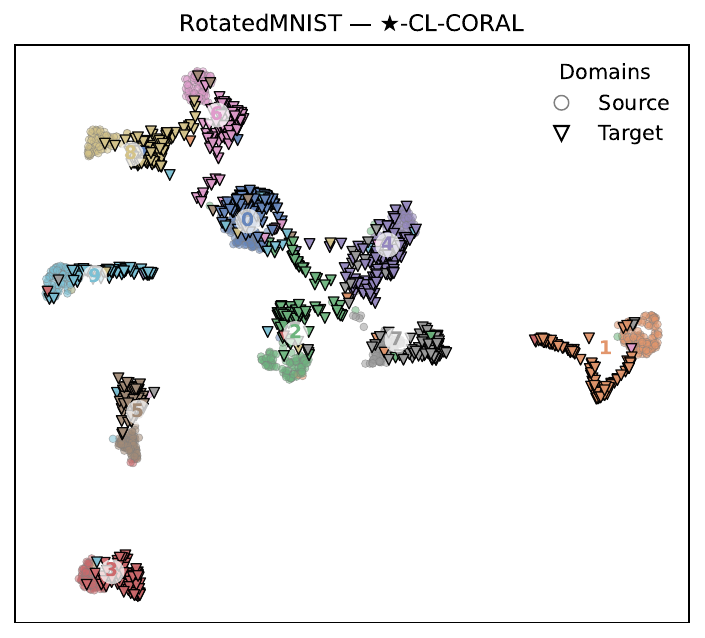}
    \end{subfigure}
    \vspace{-5pt}
    \caption{
    {
    \setlength{\fboxsep}{0pt}
    \textbf{Feature space visualizations (RotatedMNIST).}
    We visualize the feature space of Finetune (left), COPE (middle), and $\bigstar$-CL-CORAL (right) using UMAP \citep{mcinnes2018umap}. Colored numbers denote cluster centroids for the respective class.
    }}
    \label{fig:feature_space_tripple}
    \vspace{-7pt}
\end{figure}

\begin{figure}[!htbp]
\centering
    \begin{subfigure}{.45\linewidth}
        \centering
        \includegraphics[width=\linewidth]{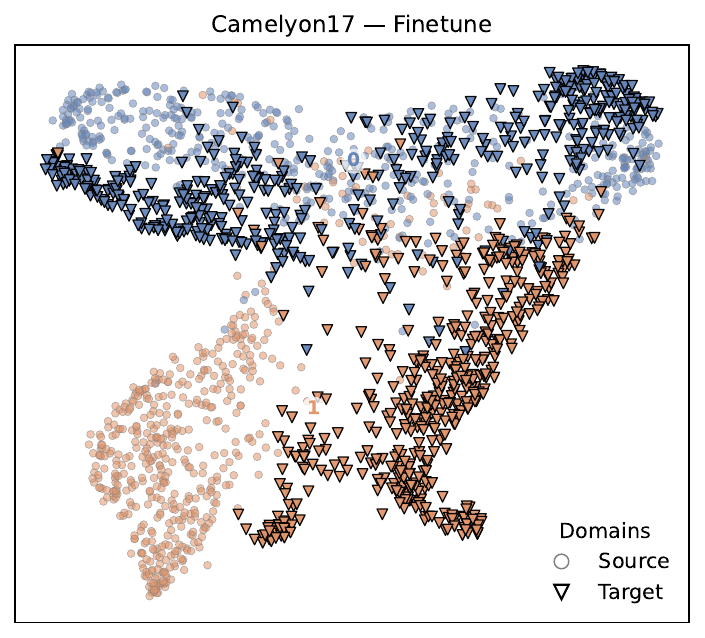}
    \end{subfigure}%
    \begin{subfigure}{.45\linewidth}
        \centering
        \includegraphics[width=\linewidth]{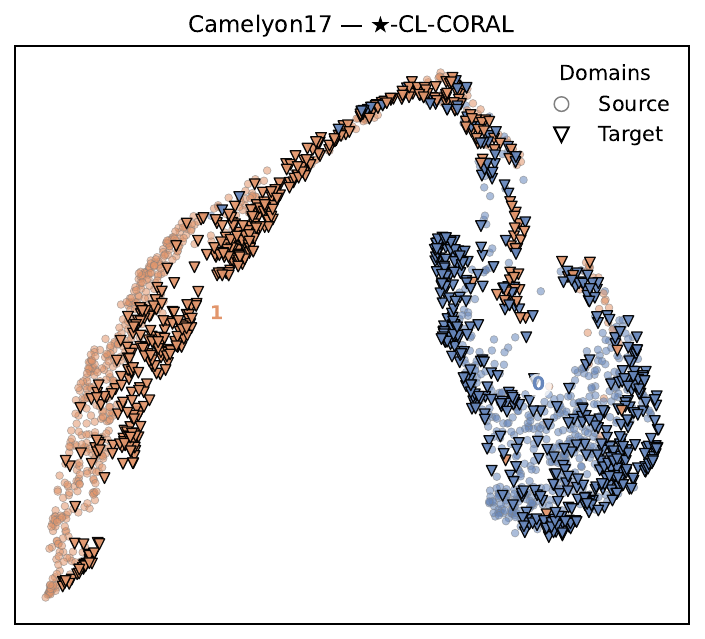}
    \end{subfigure}
    \vspace{-5pt}
    \caption{
    {
    \setlength{\fboxsep}{0pt}
    \textbf{Feature space visualizations (Camelyon17).}
    We visualize the feature space of Finetune (left) and $\bigstar$-CL-CORAL (right) using UMAP \citep{mcinnes2018umap}. Colored numbers denote cluster centroids for the respective class.
    }}
    \label{fig:feature_space_double}
    \vspace{-7pt}
\end{figure}


\end{document}